\documentclass{article}

\usepackage[eandd, final]{neurips_2026}

\usepackage[T1]{fontenc}
\usepackage[utf8]{inputenc}
\usepackage{microtype}
\usepackage{graphicx}
\usepackage{booktabs}
\usepackage{multirow}
\usepackage{amsmath,amssymb}
\usepackage{url}
\usepackage[hidelinks,breaklinks=true]{hyperref}
\usepackage{xcolor}
\usepackage{colortbl}
\usepackage{wrapfig}
\usepackage{float}
\usepackage{placeins}
\usepackage{pgfplots}
\pgfplotsset{compat=1.18}
\usepgfplotslibrary{groupplots}

\usepackage{algorithm}
\usepackage{algpseudocode}
\usepackage{algorithmicx}

\definecolor{globeblue}{HTML}{2A6DB0}

\makeatletter
\newenvironment{breakablealgorithm}
  {%
   \begin{center}
     \refstepcounter{algorithm}%
     \hrule height.8pt depth0pt \kern2pt%
     \renewcommand{\caption}[2][\relax]{%
       {\raggedright\textbf{\ALG@name~\thealgorithm} ##2\par}%
       \ifx\relax##1\relax
         \addcontentsline{loa}{algorithm}{\protect\numberline{\thealgorithm}##2}%
       \else
         \addcontentsline{loa}{algorithm}{\protect\numberline{\thealgorithm}##1}%
       \fi
       \kern2pt\hrule\kern2pt
     }%
  }{%
     \kern2pt\hrule\relax%
   \end{center}
  }
\makeatother

\usepackage{tikz}
\usetikzlibrary{positioning, arrows.meta, shapes.geometric, fit, backgrounds, calc}

\usepackage[breakable]{tcolorbox}
\usepackage{fancyvrb}
\usepackage{enumitem}

\usepackage{fontawesome5}

\tcbset{
  examplebox/.style={
    boxrule=0.4pt,
    colback=white,
    fonttitle=\bfseries\small,
    left=4pt, right=4pt, top=4pt, bottom=4pt,
    before skip=6pt, after skip=6pt,
    breakable
  }
}

\newcommand{\hflogo}{\raisebox{-0.18ex}{\includegraphics[height=1em]{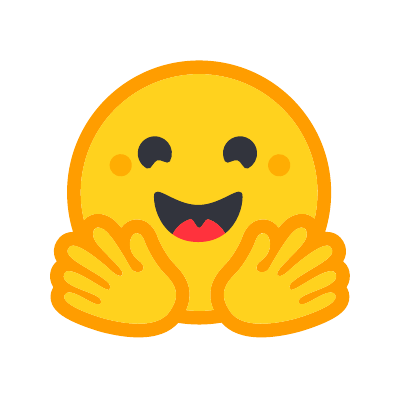}}}

\title{SciR: A Controllable Benchmark for Scientific Reasoning in LLMs}

\author{%
  Pierre Beckmann\textsuperscript{1,2} \quad
  Marco Valentino\textsuperscript{3} \quad
  Andr\'{e} Freitas\textsuperscript{1,4,5}\\[0.4em]
  \textsuperscript{1}Idiap Research Institute, Switzerland\\
  \textsuperscript{2}EPFL, Switzerland\\
  \textsuperscript{3}School of Computer Science, University of Sheffield, UK\\
  \textsuperscript{4}Department of Computer Science, University of Manchester, UK\\
  \textsuperscript{5}National Biomarker Centre, CRUK Manchester Institute, UK\\[0.3em]
  \texttt{pierrebeckmann@gmail.com}
}

\begin{document}
\maketitle

\vspace{-0.4em}
\begin{center}
\hflogo~\textbf{Data:} \href{https://huggingface.co/datasets/sci-reason/scir}{\nolinkurl{https://huggingface.co/datasets/sci-reason/scir}}\\
\faGithub~\textbf{Code:} \href{https://github.com/idiap/scir}{\nolinkurl{https://github.com/idiap/scir}}\\
{\color{globeblue}\faGlobe}~\textbf{Website:} \href{https://scir-explorer.github.io}{\nolinkurl{https://scir-explorer.github.io}}
\end{center}
\vspace{0.4em}

\begin{abstract}
Three paradigmatic forms of inference recur across scientific reasoning: deduction, induction, and causal abduction. Reliably evaluating LLMs on these in scientific settings is currently out of reach: scientific benchmarks built on human annotations are costly and lack mechanistic ground truth, while synthetic logical-reasoning benchmarks do not resemble real scientific documents. We introduce \textbf{SciR}, a benchmark that combines multi-paradigm reasoning with controllable scientific rendering, anchored on three paradigmatic scientific problems. Tasks are generated from formal objects (deduction tree, inductive rule hypothesis, causal graph) to guarantee verifiable answers, then rendered into multi-document scientific discourse via per-track domain-tuned genres. The construction lets us independently vary two difficulty axes: how hard it is to extract the key information needed for inference, and how hard the principled inference itself is. We test six models. Both axes hurt every model, and their effects compound. The rendering even hurts neuro-symbolic pipelines, which hand inference to a verified solver. The two axes yield a per-model extraction-vs-inference profile: for instance, reasoning models like deepseek-r1 mostly surpass non-reasoning instruct models on the inference axis. To our knowledge, SciR is the first multi-paradigm scientific-reasoning benchmark with parametric control on both extraction and inference difficulty.
\end{abstract}

\section{Introduction}
\label{sec:intro}

Three paradigmatic forms of inference recur across scientific reasoning: \emph{deduction} (drawing consequences from established rules), \emph{induction} (generalising rules from observations), and \emph{causal abduction} (inferring causal structure from effects). Yet we lack tools to evaluate LLMs on these inference types in realistic scientific domains, both at the level of the underlying task and at the level of the scientific-document surface.

Existing benchmarks fall short on at least one dimension. Formal reasoning benchmarks \citep{tafjord2021proofwriter,dalvi2021entailmentbank,han2024folio,chen2025justlogic,lin2025zebralogic} provide auditable ground truth and parametric difficulty but no scientific surface. Science-facing benchmarks~\citep{wadden2020scifact,majumder2024discoverybench,acharya2026causcibench,shojaee2025llmsrbench} carry realistic surface but each covers one task family at a time, with answers that reduce to human conclusions and difficulty that cannot be varied as a benchmark parameter. Closing this gap requires three properties at once: \emph{formal control} (auditable ground truth and parametric difficulty), \emph{scientific grounding} (problems that resemble realistic scientific settings), and \emph{surface heterogeneity} (the rendering tests information extraction from realistic documents, not just symbolic inference on clean premises).

We introduce \textbf{SciR}, a benchmark designed around these three properties. Each of three tracks is anchored on a \emph{paradigmatic} scientific problem: the Sachs et al.~\citep{sachs2005causal} signalling network for causal abduction, drug--drug interactions over DrugBank~\citep{knox2024drugbank,dhami2018drugdrug} for induction, and developmental-biology lineage syllogisms for deduction. Tasks are generated as formal objects before being rendered into multi-document scientific discourse via a \emph{domain-tuned scientific rendering scheme}. First, the rendering is domain-specific: each track has its own curated pool of scientific document genres (lab notebooks, database entries, EHR notes, perturbation screens, etc.), so rendered tasks look like the kind of documents that arise in that scientific setting. Second, the scheme guarantees information preservation by construction, so rendering difficulty can be increased without losing solvability. Two orthogonal axes can then be dialled independently: \emph{inference complexity} (parameters of the formal object) and \emph{premise obfuscation} (parameters of the rendering). The two axes also give a diagnostic, decomposing model failure into extraction and inference. This paper provides:
\begin{itemize}[leftmargin=*, nosep]
    \item \textbf{SciR}: a controllable benchmark spanning all three reasoning paradigms across three paradigmatic scientific problems, paired with a domain-tuned \emph{scientific rendering scheme} (cross-validated, information-preserving by construction). The two-axis design (inference complexity and premise obfuscation) yields a per-model profile of \emph{premise-extraction} and \emph{principled-inference} skill.
    \item \textbf{Findings from six models}: (i) both axes hurt every model, and their effects compound (e.g., gpt-4o CoT loses {$9.4$} pts along inference complexity alone, $12.5$ along premise obfuscation alone, and {$25.1$} along both combined); (ii) reasoning models lead instruct models on both axes, with a wider gap on principled inference (deepseek-r1 over gpt-4o by $53$ pts on inference vs $24$ pts on extraction); (iii) the rendering hurts every architecture class, including neuro-symbolic pipelines that hand inference to a verified solver (e.g., {on Hard, gpt-4o's NS accuracy roughly halves, from $88.6$ on NL to $43.4$ on Obf}).
\end{itemize}
Our methodology is general and not tied to a specific domain, providing mechanisms for future exploration in other scientific contexts. Our findings further show that challenges in scientific reasoning come from two distinct axes, underlying inference and the complexity of scientific discourse, providing actionable insights for future model development.

\section{Benchmark design}

SciR is a benchmark for testing the three paradigmatic forms of scientific reasoning under conditions that approximate real scientific work. Each task consists of multiple scientific documents (lab notes, mechanistic statements, drug labels, intervention records, table fragments, background facts), and the solver must produce an answer with verifiable ground truth. The three paradigms each map to a paradigmatic scientific problem: \emph{deduction} over lineage syllogisms in developmental biology, \emph{induction} over drug--drug interactions on DrugBank facts, and \emph{causal abduction} over the Sachs et al.~\citep{sachs2005causal} protein-signalling network. Each task is generated from a latent formal object (deduction tree, inductive rule hypothesis, partially observed causal graph), so the answer is determined by the object itself before rendering. Two axes can be dialled independently: \emph{inference complexity} (parameters of the formal object) and \emph{premise obfuscation} (parameters of the rendering).

\begin{figure}[t]
    \centering
    \includegraphics[width=\textwidth]{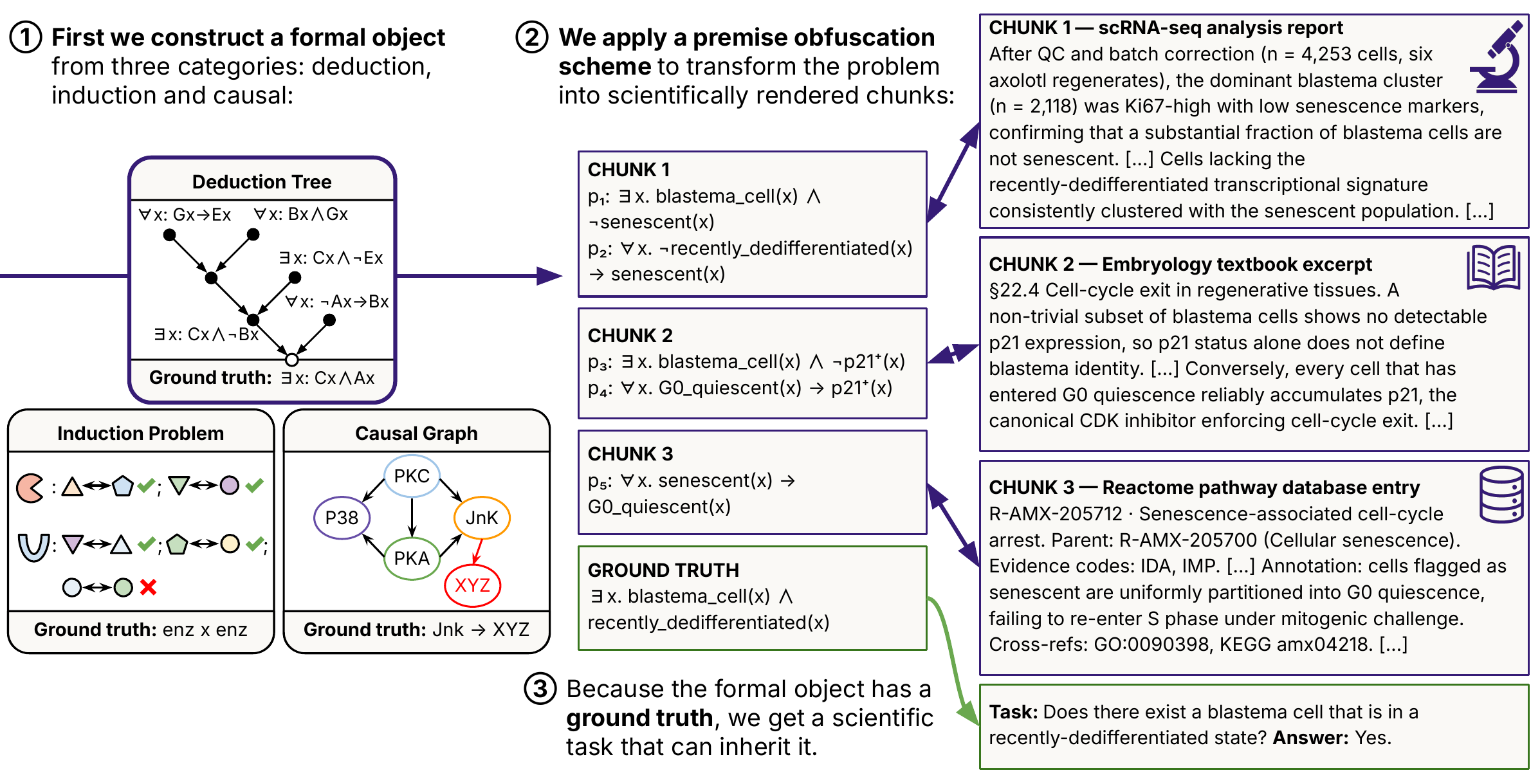}
    \caption{\textbf{Pipeline overview.} A formal object (here, a deduction tree) is rendered into multi-genre scientific chunks (in this example an scRNA-seq report, an embryology textbook excerpt, and a Reactome entry) and concatenated into the final task. The same pipeline applies across all three tracks, with the rendering contract described in \S\ref{sec:obf}.}
    \label{fig:pipeline}
\end{figure}

\paragraph{A lightweight formal view.} Let $\Delta = \{\delta_1, \ldots, \delta_m\}$ be the rendered document collection presented to the solver. Solving decomposes into \emph{premise extraction} $\mathcal{E}$ and \emph{principled inference} $\vdash_f$,
\[
\Delta \xrightarrow{\mathcal{E}} \Gamma \;\to\; z_f \;\vdash_f\; y,
\]
where $\mathcal{E}$ extracts typed evidence $\Gamma$ from heterogeneous scientific documents, $z_f$ is the latent formal or computational object of family $f$, and $\vdash_f$ is the family-appropriate inference relation that yields the answer $y$. Generation runs in the opposite direction (Fig.~\ref{fig:pipeline}),
\[
z_f \xrightarrow{\mathrm{ground}} s_f \xrightarrow{\mathcal{O}} \Delta,
\]
where $s_f$ is the scientifically grounded problem state and $\mathcal{O}$ is the discourse-rendering operator (\S\ref{sec:obf}). The two parametric axes of the benchmark are inference complexity (parameters of $s_f$) and premise obfuscation (parameters of $\mathcal{O}$).

\subsection{Deduction track}

\begin{figure}[ht]
    \centering
    \begin{minipage}[c]{0.5\linewidth}
        \centering
        \includegraphics[width=\linewidth]{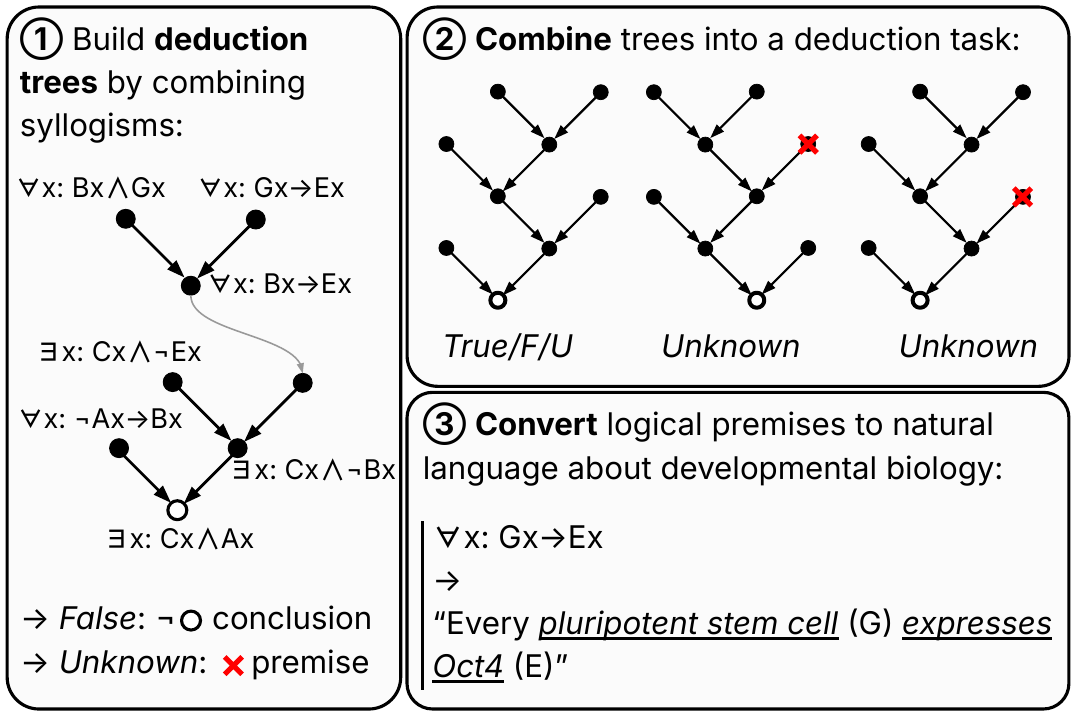}
    \end{minipage}\hfill
    \begin{minipage}[c]{0.46\linewidth}
        \caption{\textbf{Deduction dataset generation.} Trees of syllogisms are built by chained premise replacement; labelling them \textit{True}, \textit{False}, or \textit{Unknown} amounts to keeping, negating the conclusion, or deleting a premise. Each task pairs a base tree with one or more \textit{Unknown} distractor trees; logical premises are then converted to natural language and instantiated with developmental-biology pathway data. The task is to recover the validity of the conclusion from these premises.}
        \label{fig:deduction}
    \end{minipage}
\end{figure}

The deduction dataset asks whether a claim about a cellular or developmental process follows from a set of premises describing relations in that process (answer: \textit{True}, \textit{False}, or \textit{Unknown}). Each task is built from first-order-logic syllogisms (2--3 premises that deductively imply a conclusion) chained into deduction trees and instantiated with developmental-biology terminology (Fig.~\ref{fig:deduction}). Premise extraction $\mathcal{E}$ must produce from $\Delta$ a set of FOL premises $\Gamma_{\text{prem}}$ and the target hypothesis $\Gamma_{\text{hyp}}$; principled inference is FOL entailment $\Gamma_{\text{prem}} \vdash \Gamma_{\text{hyp}}$.

Replacements are accepted only when they introduce no new implication between variables already related in the tree, keeping the logical structure coherent. Predicates are instantiated with entities from one of 20 hand-curated developmental-biology contexts (embryonic cell-state transitions, germ-layer specification, stem-cell differentiation, \ldots). See App.~\ref{app:alg-deduction} for pseudocode and App.~\ref{app:deduction-details} for the full context list and a bigger example tree.

\paragraph{Difficulty knobs.} Two knobs control difficulty: \emph{depth} (number of premise replacements) sets proof length, and \emph{width} (number of distractor trees) adds parallel exploration paths. Depth is purely an inference-complexity knob. Width is primarily inference-complexity but also dilutes task-relevant information.

\subsection{Induction track}

\begin{figure}[ht]
    \centering
    \begin{minipage}[c]{0.5\linewidth}
        \centering
        \includegraphics[width=\linewidth]{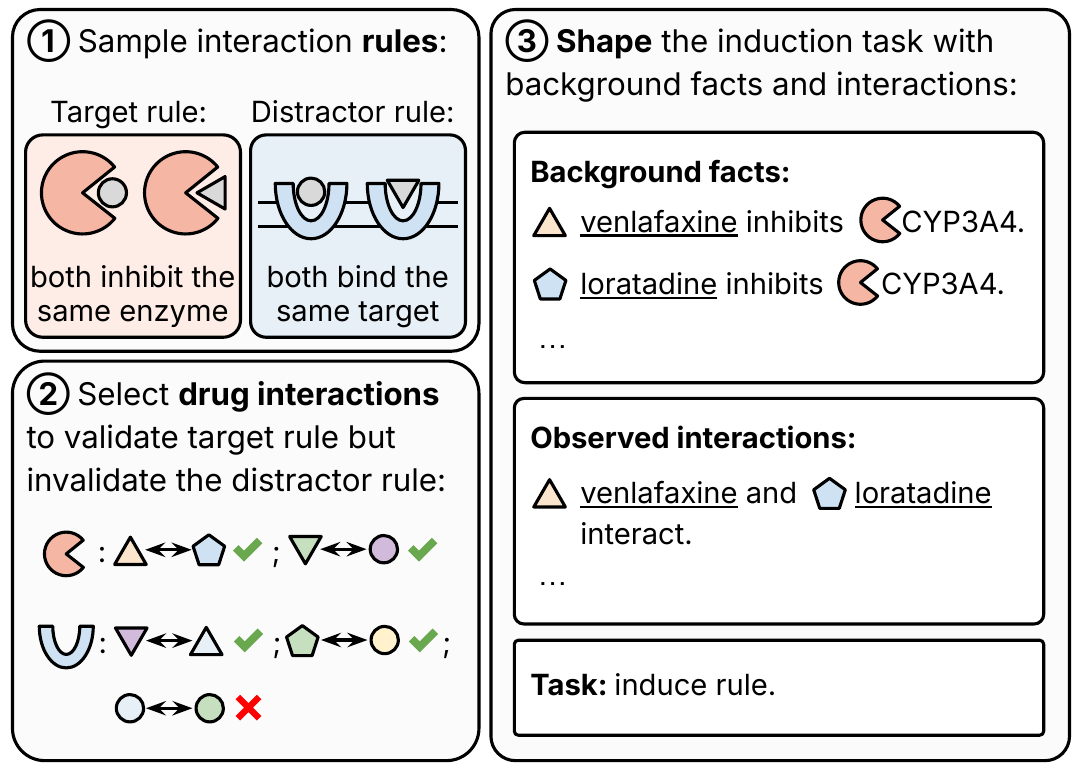}
    \end{minipage}\hfill
    \begin{minipage}[c]{0.46\linewidth}
        \caption{\textbf{Induction dataset generation.} A target rule and one or more distractor rules are sampled from a curated set of drug-interaction patterns (e.g., \emph{both inhibit the same enzyme} vs.\ \emph{both bind the same target}). Drug pairs are then selected so the positives support each rule, while one negative example per distractor invalidates that distractor without falsifying the target. The full task is rendered as background drug--protein facts plus observed interactions. From these, the task is to induce the rule.}
        \label{fig:induction}
    \end{minipage}
\end{figure}

The induction dataset asks which rule explains a set of observed drug--drug interactions, given each drug's enzyme / transporter / target profile (drawn from DrugBank~\citep{knox2024drugbank}). Background facts are restricted to predicates used by the candidate rules, but the resulting fact set is still large: each candidate hypothesis must be checked against every positive and negative example simultaneously, with consistent supporting facts retrieved for each. Premise extraction $\mathcal{E}$ must produce from $\Delta$ drug--protein background facts $\Gamma_{\text{facts}}$ and positive and negative drug-pair observations $\Gamma_{\text{pos}}$ and $\Gamma_{\text{neg}}$; {principled inference selects a rule that fires on at least one example of $\Gamma_{\text{pos}}$ and on none of $\Gamma_{\text{neg}}$ given $\Gamma_{\text{facts}}$; any such rule is a correct answer $y$ (App.~\ref{app:models}).} Fig.~\ref{fig:induction} summarises the pipeline; see App.~\ref{app:alg-induction} for pseudocode.

\paragraph{Difficulty knobs.} Two knobs control task difficulty: the number of distractor rules ($n_{\text{dist}}$) sets how many alternative hypotheses the solver must eliminate, and the number of positive examples per rule ($n_{\text{pos}}$) enlarges the background fact set against which each candidate rule must be checked. Both are primarily inference-complexity. More facts also dilute the task-relevant information.

\subsection{Causal track}

\begin{figure}[ht]
    \centering
    \begin{minipage}[c]{0.5\linewidth}
        \centering
        \includegraphics[width=\linewidth]{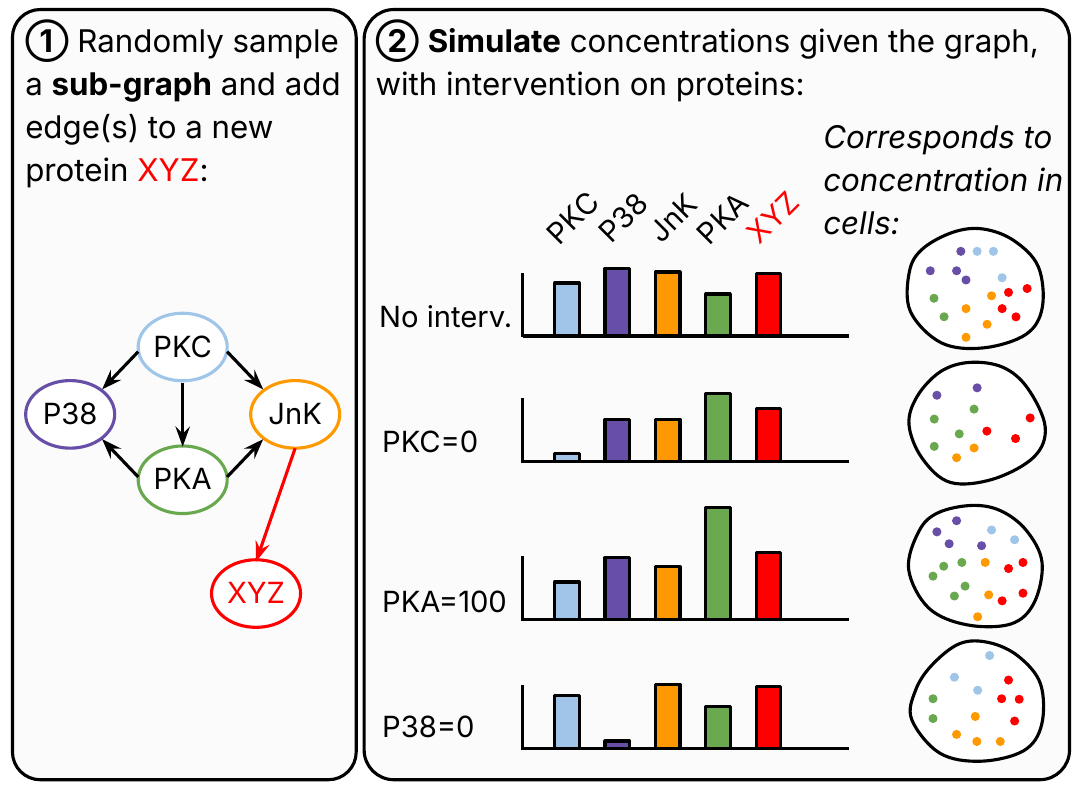}
    \end{minipage}\hfill
    \begin{minipage}[c]{0.46\linewidth}
        \caption{\textbf{Causal dataset generation.} A connected subgraph is sampled from the Sachs protein-signalling network and a fictional protein \textit{XYZ} is added with a random set of edges. Protein concentrations are simulated with a linear Gaussian SCM under observational and per-node do-interventions (inhibition to $\sim 0$, activation to a high value), yielding a table (one row per cell, one column per protein) that matches the format of the original Sachs data. From this table, the task is to recover the edges of \textit{XYZ}.}
        \label{fig:causal}
    \end{minipage}
\end{figure}

The causal dataset asks how a new signalling protein \textit{XYZ} connects to a known sub-network, given observational and interventional concentration data. Tasks follow the protein-signalling setting of \citet{sachs2005causal}; using an invented protein rules out answers that rely on memorised biology. Premise extraction $\mathcal{E}$ must produce from $\Delta$ observational concentration rows $\Gamma_{\text{obs}}$, interventional rows $\Gamma_{\text{int}}$ tagged by the perturbed protein, and the known subgraph edges $\Gamma_{\text{known}}$; principled inference identifies the edges incident to \textit{XYZ} from $\Gamma_{\text{obs}} \cup \Gamma_{\text{int}}$ given $\Gamma_{\text{known}}$, and that edge set is the answer $y$. Fig.~\ref{fig:causal} summarises the pipeline; see App.~\ref{app:alg-causal} for pseudocode.

\paragraph{Difficulty knobs.} Three knobs control task difficulty: the subgraph size sets how much context must be integrated, the number of \textit{XYZ} edges sets the cardinality of the hidden structure, and the number of samples per environment sets the strength of the statistical signal. More samples make inference easier in principle, but enlarge the raw table and thereby make extraction harder.

\subsection{Scientific rendering scheme}
\label{sec:obf}

\begin{figure}[ht]
    \centering
    \begin{minipage}[c]{0.5\linewidth}
        \centering
        \includegraphics[width=\linewidth]{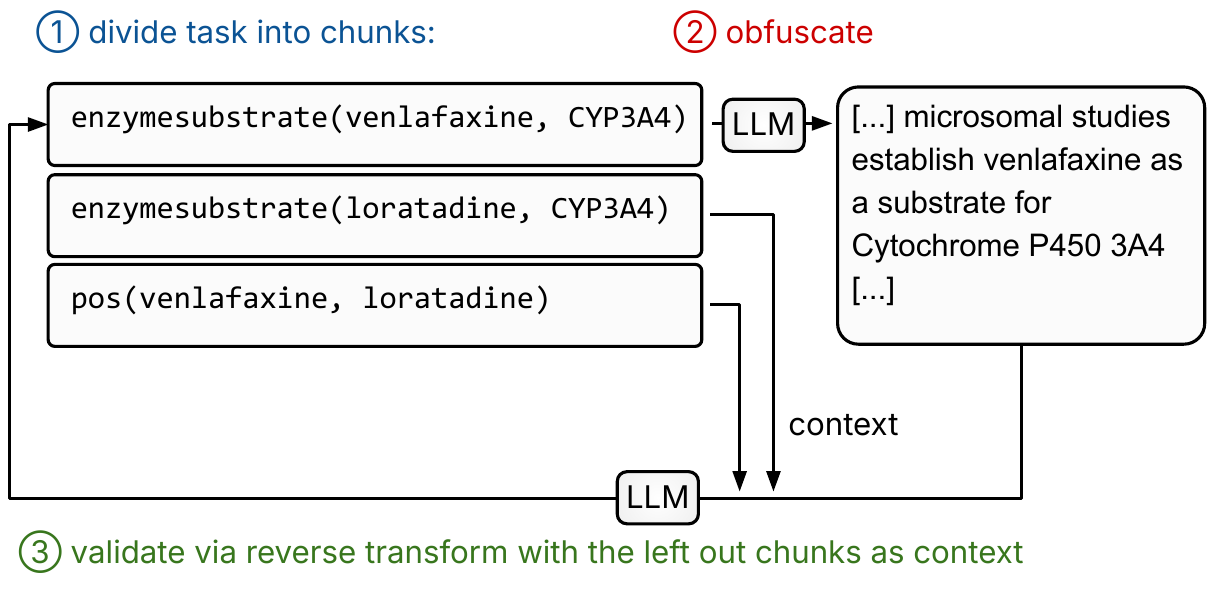}
    \end{minipage}\hfill
    \begin{minipage}[c]{0.46\linewidth}
        \caption{\textbf{Scientific rendering scheme.} The task is split into chunks, each rewritten by an LLM into a scientifically realistic document format. Each rewrite is accepted only if a second LLM can invert it, given the remaining chunks as context, back to the original structured chunk, so information preservation follows from the scheme.}
        \label{fig:obfuscate}
    \end{minipage}
\end{figure}

The scheme serves two purposes. \textbf{Scientific instantiation}: it renders the formal object into the shape of documents that genuinely arise in the target scientific domain (pharmacology paperwork for drug interactions, proteomics reports for signalling, developmental-biology genres for lineage logic). \textbf{Premise obfuscation}: it hides the task's premises in surface noise (distractor content, multi-genre fragmentation, narrative paraphrase) so the solver has to actually extract them. Both effects come from a single transform, accepted only when a cross-validated \emph{invertibility contract} shows information is preserved (Fig.~\ref{fig:obfuscate}).

\textbf{The contract.} A task is split into chunks $c_1, \ldots, c_N$. For each chunk $c_i$, an LLM $\mathcal{T}$ (structured $\to$ scientific discourse) proposes a rendering $\mathcal{T}(c_i)$; a second LLM $\mathcal{T}^{-1}$ tries to invert it \emph{given the other chunks as context}, and the rendering is kept only when $\mathcal{T}^{-1}\bigl(\mathcal{T}(c_i)\,\big|\,c_{\neq i}\bigr) = c_i$. The contextual chunks let the inverse recover entity names and cross-chunk references that an isolated chunk would lose. At solve time, however, the solver sees all chunks at once with no such anchoring, so the difficulty comes from putting the chunks together. Obfuscation strength scales with the chunk count $N$, the size of the per-track style pool, and the capability of the rendering LLM.

\textbf{Per-track domain instantiation.} Each task is split into $N$ chunks rewritten into genres from a curated pool of eight document types per track, each shaped by track-specific transform constraints. {Genres are chosen so that evidence appears where it would in practice, e.g.\ FDA drug labels, DrugBank entries, and EHR discharge summaries for induction, and wet-lab notebooks, LC-MS/MS reports, and perturbation screens for causal discovery.} End-to-end task examples are in App.~\ref{app:examples}; the rendering pseudocode in App.~\ref{app:alg-obfuscate}; the full style pools and per-track constraints in App.~\ref{app:styles}. The pipeline is released as code rather than a fixed test set, so harder tiers can be regenerated as models saturate (also our main practical defence against contamination).

\section{Evaluation and results}
\label{sec:eval}

\subsection{Setup}

Each track runs at Easy and Hard tiers (parametric knobs: {depth and width} for deduction, distractor-rule and positive-example counts for induction, subgraph size and \textit{XYZ}-edge count for causal). Concrete settings, answer formats, and chance baselines are in App.~\ref{app:models}. The scientific rendering uses $N{=}2$ chunks per task{ ($3$ for induction)} from per-track style pools (App.~\ref{app:styles}), with o3-mini playing both renderer and inverse roles. The construction keeps only chunks that o3-mini itself can invert. We therefore include deepseek-r1 as a second, unrelated reasoning baseline, so the reasoning-vs-instruct comparison does not rest on the renderer alone.

We evaluate three architectures per task. \textbf{CoT} prompts a base LLM to reason step by step and emit an answer. \textbf{Neuro-symbolic (NS)} pairs an LLM formaliser with a per-domain symbolic backend (Prover9~\citep{mccune2005prover9} for deduction, Popper~\citep{cropper2021popper} for induction, GIES~\citep{hauser2012gies} for causal): the LLM converts the rendered task into formal input and the solver returns a verified answer. \textbf{SymbCoT$^{*}$} is a single-pass variant of SymbCoT~\citep{xu2024symbcot}: same formalisation as NS, but a second LLM call answers from the formalised text instead of running a solver. We evaluate six base models (\texttt{gpt-4o}, \texttt{o3-mini}, \texttt{deepseek-r1}, \texttt{llama-3.3-70b}, \texttt{qwen3-30b}, \texttt{olmo-3.1-32b}) on OpenRouter at temperature~0 with $n{=}200$ tasks per cell. Decoding and retry budgets are in App.~\ref{app:models}.

\subsection{Main results}

\begin{figure}[t]
\centering
\begin{tikzpicture}
\begin{groupplot}[
  group style={group size=4 by 1, horizontal sep=0.45cm, ylabels at=edge left, yticklabels at=edge left},
  width=4.0cm, height=4.0cm,
  ymin=0, ymax=100,
  ytick={0,25,50,75,100},
  ymajorgrids,
  ylabel={Chance-norm.\ accuracy (\%)},
  ylabel style={font=\scriptsize},
  tick label style={font=\tiny},
  title style={font=\small, yshift=-3pt},
  xtick=data,
  x tick label style={rotate=35, anchor=east, font=\tiny, yshift=-3pt, xshift=4pt},
  symbolic x coords={gpt-4o, o3-mini, deepseek, llama, qwen, olmo},
  enlarge x limits=0.10,
]

\nextgroupplot[title={NL $\cdot$ Easy},
  legend style={font=\tiny, draw=none, fill=white, fill opacity=0.85, text opacity=1, inner sep=2pt, row sep=-1pt},
  legend pos=south west]
\addplot[only marks, mark=*, mark size=2pt, color=red!75!black] coordinates {
  (gpt-4o, 42.6) (o3-mini, 71.0) (deepseek, 95.6) (llama, 24.2) (qwen, 71.5) (olmo, 33.5)
};
\addlegendentry{CoT}
\addplot[only marks, mark=square*, mark size=2pt, color=green!50!black] coordinates {
  (gpt-4o, 92.9) (o3-mini, 95.6) (deepseek, 93.7) (llama, 86.9) (qwen, 76.7) (olmo, 65.0)
};
\addlegendentry{{NS}}
\draw[->, >=stealth, thick, green!50!black] (axis cs:gpt-4o, 42.6) -- (axis cs:gpt-4o, 92.9);
\draw[->, >=stealth, thick, green!50!black] (axis cs:o3-mini, 71.0) -- (axis cs:o3-mini, 95.6);
\draw[->, >=stealth, thick, red!75!black] (axis cs:deepseek, 95.6) -- (axis cs:deepseek, 93.7);
\draw[->, >=stealth, thick, green!50!black] (axis cs:llama, 24.2) -- (axis cs:llama, 86.9);
\draw[->, >=stealth, thick, green!50!black] (axis cs:qwen, 71.5) -- (axis cs:qwen, 76.7);
\draw[->, >=stealth, thick, green!50!black] (axis cs:olmo, 33.5) -- (axis cs:olmo, 65.0);

\nextgroupplot[title={NL $\cdot$ Hard}]
\addplot[only marks, mark=*, mark size=2pt, color=red!75!black] coordinates {
  (gpt-4o, 33.2) (o3-mini, 52.3) (deepseek, 89.5) (llama, 19.2) (qwen, 51.2) (olmo, 19.3)
};
\addplot[only marks, mark=square*, mark size=2pt, color=green!50!black] coordinates {
  (gpt-4o, 88.6) (o3-mini, 87.7) (deepseek, 82.8) (llama, 82.0) (qwen, 65.6) (olmo, 43.5)
};
\draw[->, >=stealth, thick, green!50!black] (axis cs:gpt-4o, 33.2) -- (axis cs:gpt-4o, 88.6);
\draw[->, >=stealth, thick, green!50!black] (axis cs:o3-mini, 52.3) -- (axis cs:o3-mini, 87.7);
\draw[->, >=stealth, thick, red!75!black] (axis cs:deepseek, 89.5) -- (axis cs:deepseek, 82.8);
\draw[->, >=stealth, thick, green!50!black] (axis cs:llama, 19.2) -- (axis cs:llama, 82.0);
\draw[->, >=stealth, thick, green!50!black] (axis cs:qwen, 51.2) -- (axis cs:qwen, 65.6);
\draw[->, >=stealth, thick, green!50!black] (axis cs:olmo, 19.3) -- (axis cs:olmo, 43.5);

\nextgroupplot[title={Obf $\cdot$ Easy}]
\addplot[only marks, mark=*, mark size=2pt, color=red!75!black] coordinates {
  (gpt-4o, 30.1) (o3-mini, 43.8) (deepseek, 59.2) (llama, 14.2) (qwen, 38.6) (olmo, 13.8)
};
\addplot[only marks, mark=square*, mark size=2pt, color=green!50!black] coordinates {
  (gpt-4o, 58.9) (o3-mini, 77.1) (deepseek, 82.6) (llama, 52.2) (qwen, 56.7) (olmo, 25.9)
};
\draw[->, >=stealth, thick, green!50!black] (axis cs:gpt-4o, 30.1) -- (axis cs:gpt-4o, 58.9);
\draw[->, >=stealth, thick, green!50!black] (axis cs:o3-mini, 43.8) -- (axis cs:o3-mini, 77.1);
\draw[->, >=stealth, thick, green!50!black] (axis cs:deepseek, 59.2) -- (axis cs:deepseek, 82.6);
\draw[->, >=stealth, thick, green!50!black] (axis cs:llama, 14.2) -- (axis cs:llama, 52.2);
\draw[->, >=stealth, thick, green!50!black] (axis cs:qwen, 38.6) -- (axis cs:qwen, 56.7);
\draw[->, >=stealth, thick, green!50!black] (axis cs:olmo, 13.8) -- (axis cs:olmo, 25.9);

\nextgroupplot[title={Obf $\cdot$ Hard}]
\addplot[only marks, mark=*, mark size=2pt, color=red!75!black] coordinates {
  (gpt-4o, 17.5) (o3-mini, 35.4) (deepseek, 47.6) (llama, 11.5) (qwen, 22.8) (olmo, 9.0)
};
\addplot[only marks, mark=square*, mark size=2pt, color=green!50!black] coordinates {
  (gpt-4o, 43.4) (o3-mini, 65.5) (deepseek, 67.3) (llama, 46.1) (qwen, 47.7) (olmo, 14.1)
};
\draw[->, >=stealth, thick, green!50!black] (axis cs:gpt-4o, 17.5) -- (axis cs:gpt-4o, 43.4);
\draw[->, >=stealth, thick, green!50!black] (axis cs:o3-mini, 35.4) -- (axis cs:o3-mini, 65.5);
\draw[->, >=stealth, thick, green!50!black] (axis cs:deepseek, 47.6) -- (axis cs:deepseek, 67.3);
\draw[->, >=stealth, thick, green!50!black] (axis cs:llama, 11.5) -- (axis cs:llama, 46.1);
\draw[->, >=stealth, thick, green!50!black] (axis cs:qwen, 22.8) -- (axis cs:qwen, 47.7);
\draw[->, >=stealth, thick, green!50!black] (axis cs:olmo, 9.0) -- (axis cs:olmo, 14.1);

\end{groupplot}
\end{tikzpicture}
\caption{\textbf{Both inference complexity and rendering obfuscation lower accuracy, and their effects compound.} Each panel is one (tier, rendering) cell averaged across the three tracks. Circles show CoT, squares show neuro-symbolic, and arrows trace each model's lift from one to the other. Reading left to right (NL Easy through Obf Hard), accuracy falls on both methods for nearly every model. The drop along the rendering axis applies even to the neuro-symbolic pipeline, which hands inference to a verified solver.}
\label{fig:cot-vs-ns}
\end{figure}
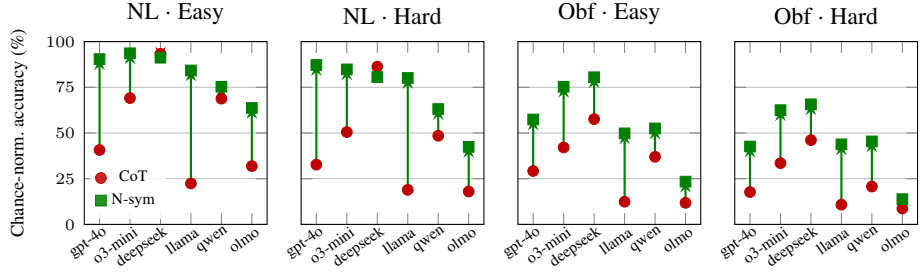

Table~\ref{tab:results} reports raw accuracy across all cells. From here on, all summary statistics, figures, and per-model probe scores use \emph{chance-normalised accuracy} $(acc - chance)/(1 - chance)${, clipped at $0$}. Fig.~\ref{fig:cot-vs-ns} provides a chance-normalised summary of CoT vs.\ NS along the two axes.

\begin{table}[ht]
\centering
\scriptsize
\setlength{\tabcolsep}{3.5pt}
\setlength{\arrayrulewidth}{0.5pt}
\setlength{\extrarowheight}{1pt}
\renewcommand{\arraystretch}{1.15}
\arrayrulecolor{black}
\newcommand{\tr}[3]{#1\,$\stackrel{\scriptscriptstyle #2}{\to}$\,#3}
\newcommand{\trL}[3]{\textbf{#1}\,$\stackrel{\scriptscriptstyle #2}{\to}$\,#3}
\newcommand{\trR}[3]{#1\,$\stackrel{\scriptscriptstyle #2}{\to}$\,\textbf{#3}}
\newcommand{\trLR}[3]{\textbf{#1}\,$\stackrel{\scriptscriptstyle #2}{\to}$\,\textbf{#3}}
\newcommand{\bc}[1]{\textcolor{red!55!black}{#1}}
 \caption{\textbf{Evaluating six models on SciR across all tracks, tiers, and rendering levels.} Each cell reads NL\,$\to$\,Obf with the obfuscation gap (Obf $-$ NL) above the arrow ($n{=}200$, temperature 0). The top row gives the random-chance baseline per column; \bc{below-chance values} are in muted red. \textbf{Bold} marks the per-column maximum (NL and Obf separately). \textbf{CoT}: chain-of-thought from the {task text}. \textbf{{NS}}: LLM formaliser + symbolic solver (Prover9 / Popper / GIES). \textbf{SymbCoT$^{*}$}: same formalisation, but a second LLM call answers from it.}
\label{tab:results}
\begin{tabular}{l l | c c | c c | c c}
\toprule
 & & \multicolumn{2}{c|}{Deduction} & \multicolumn{2}{c|}{Induction} & \multicolumn{2}{c}{Causal} \\
\cmidrule(lr){3-4}\cmidrule(lr){5-6}\cmidrule(lr){7-8}
Model & Config. & Easy & Hard & Easy & Hard & Easy & Hard \\
 & & {\tiny NL$\to$Obf} & {\tiny NL$\to$Obf} & {\tiny NL$\to$Obf} & {\tiny NL$\to$Obf} & {\tiny NL$\to$Obf} & {\tiny NL$\to$Obf} \\
\midrule
\rowcolor{black!8}
\multicolumn{2}{l|}{\emph{Random chance}} & \multicolumn{2}{c|}{33.3} & 3.2 & 3.5 & 10.0 & 5.0 \\
\multirow{3}{*}{gpt-4o}         & CoT             & \tr{59.0}{-15.0}{44.0} & \tr{54.0}{-20.5}{33.5} & \tr{46.0}{-9.5}{36.5} & \tr{28.0}{-2.5}{25.5} & \tr{50.5}{-4.5}{46.0} & \tr{46.0}{-13.0}{33.0} \\
                                & {NS}         & \tr{99.0}{-37.0}{62.0} & \tr{97.0}{-54.0}{43.0} & \tr{82.5}{-30.5}{52.0} & \tr{73.5}{-33.0}{40.5} & \tr{98.5}{-13.5}{85.0} & \trL{98.0}{-19.5}{78.5} \\
                                & SymbCoT$^{*}$   & \tr{60.0}{-4.0}{56.0} & \tr{54.5}{-14.0}{40.5} & \tr{10.5}{+8.5}{19.0} & \tr{4.0}{+6.5}{10.5} & \tr{45.0}{-6.5}{38.5} & \tr{37.0}{-1.0}{36.0} \\
\rowcolor{black!5}  & CoT             & \tr{69.0}{-37.0}{\bc{32.0}} & \tr{52.5}{-29.0}{\bc{23.5}} & \tr{63.0}{-16.5}{46.5} & \tr{45.5}{-6.0}{39.5} & \tr{98.0}{-10.0}{88.0} & \tr{85.5}{-15.0}{70.5} \\
\rowcolor{black!5} \multirow{-2}{*}{o3-mini}\hspace{-2pt} & {NS}         & \trL{100.0}{-20.5}{79.5} & \trL{99.0}{-32.0}{67.0} & \tr{90.0}{-23.0}{67.0} & \tr{74.0}{-23.0}{51.0} & \tr{97.5}{-1.0}{96.5} & \trR{92.0}{+5.0}{97.0} \\
\rowcolor{black!5}  & SymbCoT$^{*}$   & \tr{71.0}{-13.5}{57.5} & \tr{63.0}{-13.5}{49.5} & \tr{17.0}{+6.0}{23.0} & \tr{17.0}{+4.5}{21.5} & \tr{98.5}{-5.0}{93.5} & \tr{80.0}{-1.5}{78.5} \\
\multirow{3}{*}{deepseek-r1}    & CoT             & \tr{97.5}{-35.0}{62.5} & \tr{93.0}{-36.5}{56.5} & \tr{91.5}{-55.5}{36.0} & \trL{91.5}{-65.5}{26.0} & \trLR{99.5}{+0.5}{100.0} & \tr{88.5}{-3.0}{85.5} \\
                                & {NS}         & \trR{98.5}{-14.5}{84.0} & \trR{98.0}{-24.5}{73.5} & \trLR{93.0}{-9.5}{83.5} & \trR{84.5}{-15.5}{69.0} & \tr{91.5}{-1.5}{90.0} & \tr{69.0}{+6.0}{75.0} \\
                                & SymbCoT$^{*}$   & \tr{93.5}{-16.0}{77.5} & \tr{89.5}{-18.0}{71.5} & \tr{4.5}{+8.0}{12.5} & \tr{4.0}{+5.0}{9.0} & \tr{93.0}{-3.5}{89.5} & \tr{67.0}{+0.0}{67.0} \\
\rowcolor{black!5}  & CoT             & \tr{43.5}{-15.5}{\bc{28.0}} & \tr{42.5}{-12.5}{\bc{30.0}} & \tr{26.5}{-3.5}{23.0} & \tr{17.5}{+1.0}{18.5} & \tr{40.0}{-10.0}{30.0} & \tr{33.0}{-10.0}{23.0} \\
\rowcolor{black!5} \multirow{-2}{*}{\shortstack[l]{llama-\\3.3-70b}} & {NS}         & \tr{90.5}{-37.5}{53.0} & \tr{88.5}{-40.5}{48.0} & \tr{80.0}{-35.0}{45.0} & \tr{67.5}{-28.5}{39.0} & \tr{96.0}{-10.5}{85.5} & \tr{97.0}{-16.5}{80.5} \\
\rowcolor{black!5}  & SymbCoT$^{*}$   & \tr{49.0}{-8.5}{40.5} & \tr{43.0}{-6.5}{36.5} & \tr{\bc{1.0}}{+9.5}{10.5} & \tr{\bc{0.0}}{+6.5}{6.5} & \tr{26.0}{+4.5}{30.5} & \tr{28.0}{+1.5}{29.5} \\
\multirow{3}{*}{qwen3-30b}      & CoT             & \tr{90.0}{-43.0}{47.0} & \tr{80.0}{-41.0}{39.0} & \tr{58.0}{-32.5}{25.5} & \tr{44.0}{-28.0}{16.0} & \tr{75.5}{-0.5}{75.0} & \tr{44.5}{+5.0}{49.5} \\
                                & {NS}         & \tr{88.0}{-22.0}{66.0} & \tr{77.5}{-19.0}{58.5} & \tr{53.5}{-4.0}{49.5} & \tr{41.5}{-3.0}{38.5} & \tr{96.5}{-20.5}{76.0} & \tr{91.5}{-21.0}{70.5} \\
                                & SymbCoT$^{*}$   & \tr{83.5}{-10.0}{73.5} & \tr{71.5}{-9.0}{62.5} & \tr{27.5}{-2.5}{25.0} & \tr{17.0}{+6.5}{23.5} & \tr{49.0}{-8.0}{41.0} & \tr{29.0}{+0.0}{29.0} \\
\rowcolor{black!5}  & CoT             & \tr{71.5}{-38.5}{\bc{33.0}} & \tr{49.5}{-19.5}{\bc{30.0}} & \tr{45.0}{-13.0}{32.0} & \tr{36.0}{-12.5}{23.5} & \tr{\bc{5.5}}{+15.0}{20.5} & \tr{\bc{4.0}}{+7.0}{11.0} \\
\rowcolor{black!5} \multirow{-2}{*}{\shortstack[l]{olmo-\\3.1-32b}} & {NS}         & \tr{91.5}{-48.5}{43.0} & \tr{82.0}{-43.5}{38.5} & \tr{30.5}{+0.0}{30.5} & \tr{19.5}{-2.0}{17.5} & \tr{81.5}{-40.0}{41.5} & \tr{44.0}{-20.0}{24.0} \\
\rowcolor{black!5}  & SymbCoT$^{*}$   & \tr{75.0}{-39.0}{36.0} & \tr{52.0}{-22.0}{\bc{30.0}} & \tr{14.5}{+7.5}{22.0} & \tr{10.0}{+7.5}{17.5} & \tr{\bc{8.0}}{+0.0}{\bc{8.0}} & \tr{\bc{2.5}}{+0.0}{\bc{2.5}} \\
\bottomrule
\end{tabular}
\end{table}

\subsection{How models split between inference and extraction}

\paragraph{Both axes hurt every model, and their effects compound.} {Averaged over tracks (Fig.~\ref{fig:cot-vs-ns}), increasing inference complexity hurts CoT accuracy for every model, and so does premise obfuscation.} Combined, the cost is larger than either alone. Fig.~\ref{fig:cot-vs-ns} traces the typical trajectory: gpt-4o moves $42.6\to33.2\to30.1\to17.5$ across (NL Easy, NL Hard, Obf Easy, Obf Hard), and every other model follows the same downward pattern, with Hard\,$\cdot$\,Obf consistently the lowest cell.

\paragraph{The same trends apply to neuro-symbolic.} Even though every task has a verifiable formal object underneath and the symbolic solver is itself sound, NS scores still drop along both axes{, most sharply along obfuscation} (Table~\ref{tab:results}, Fig.~\ref{fig:cot-vs-ns}): gpt-4o's NS averages $92.9\to88.6\to58.9\to43.4$, and deepseek-r1 NS falls $26.4$\,pts from NL Easy to Obf Hard. {Premise obfuscation is therefore a real difficulty, not an artefact of LLM-only solving. Reasoning models also lead NS in the obfuscated cells, suggesting that formalisation benefits from reasoning ability. SciR is thus a testbed for hybrid methods that strengthen the formalisation step (e.g., multi-pass extraction, self-consistency over formalisations).} {Note that the solver cannot be the bottleneck: every retained task's gold formal input is solved by its solver during construction, so NS errors come from formalising the documents.}

\paragraph{Most of NS's lift comes from the symbolic solver.} SymbCoT$^{*}$ (sharing NS's formalisation but answering via a second LLM call) trails NS in nearly every cell of Table~\ref{tab:results}. It is especially bad on induction, where it even trails CoT {(e.g., gpt-4o on NL Easy: $10.5$ vs $46.0$ for CoT)}, most plausibly because the Popper-style Prolog output is unfamiliar input to the LLM.

\begin{figure}[t]
\centering
\begin{tikzpicture}
\begin{groupplot}[
  group style={group size=4 by 2, horizontal sep=0.4cm, vertical sep=0.55cm,
               ylabels at=edge left, yticklabels at=edge left,
               xlabels at=edge bottom, xticklabels at=edge bottom},
  width=3.7cm, height=3.7cm,
  xmin=0, xmax=100, ymin=0, ymax=100,
  xtick={0,25,50,75,100}, ytick={0,25,50,75,100},
  axis equal image,
  grid=both, grid style={dotted, gray!30},
  xlabel={Extraction (Obf$\cdot$NS, \%)},
  ylabel={Inference (NL$\cdot$CoT, \%)},
  xlabel style={font=\scriptsize}, ylabel style={font=\scriptsize},
  tick label style={font=\tiny},
  title style={font=\scriptsize, yshift=-3pt},
]
\nextgroupplot[title={Easy $\cdot$ Deduction}]
\addplot[gray!55, dashed, no marks, forget plot] coordinates {(0,0) (100,100)};
\addplot[only marks, mark=*, mark size=2.72pt, color=blue!50!cyan, fill=blue!50!cyan, fill opacity=0.55] coordinates {(43.0, 38.5)};
\addplot[only marks, mark=*, mark size=1.20pt, color=blue!70!black, fill=blue!70!black, fill opacity=0.55] coordinates {(29.5, 15.2)};
\addplot[only marks, mark=*, mark size=3.08pt, color=teal!70!black, fill=teal!70!black, fill opacity=0.55] coordinates {(49.0, 85.0)};
\addplot[only marks, mark=*, mark size=1.20pt, color=cyan!45!gray, fill=cyan!45!gray, fill opacity=0.55] coordinates {(14.5, 57.2)};
\addplot[only marks, mark=square*, mark size=1.20pt, color=orange!80!black, fill=orange!80!black, fill opacity=0.55] coordinates {(69.2, 53.5)};
\addplot[only marks, mark=square*, mark size=4.50pt, color=red!70!black, fill=red!70!black, fill opacity=0.55] coordinates {(76.0, 96.2)};
\nextgroupplot[title={Easy $\cdot$ Induction}]
\addplot[gray!55, dashed, no marks, forget plot] coordinates {(0,0) (100,100)};
\addplot[only marks, mark=*, mark size=3.99pt, color=blue!50!cyan, fill=blue!50!cyan, fill opacity=0.55] coordinates {(50.4, 44.2)};
\addplot[only marks, mark=*, mark size=3.08pt, color=blue!70!black, fill=blue!70!black, fill opacity=0.55] coordinates {(43.2, 24.1)};
\addplot[only marks, mark=*, mark size=3.27pt, color=teal!70!black, fill=teal!70!black, fill opacity=0.55] coordinates {(47.8, 56.6)};
\addplot[only marks, mark=*, mark size=3.71pt, color=cyan!45!gray, fill=cyan!45!gray, fill opacity=0.55] coordinates {(28.2, 43.2)};
\addplot[only marks, mark=square*, mark size=4.55pt, color=orange!80!black, fill=orange!80!black, fill opacity=0.55] coordinates {(65.9, 61.8)};
\addplot[only marks, mark=square*, mark size=3.96pt, color=red!70!black, fill=red!70!black, fill opacity=0.55] coordinates {(83.0, 91.2)};
\nextgroupplot[title={Easy $\cdot$ Causal}]
\addplot[gray!55, dashed, no marks, forget plot] coordinates {(0,0) (100,100)};
\addplot[only marks, mark=*, mark size=4.30pt, color=blue!50!cyan, fill=blue!50!cyan, fill opacity=0.55] coordinates {(83.3, 45.0)};
\addplot[only marks, mark=*, mark size=3.21pt, color=blue!70!black, fill=blue!70!black, fill opacity=0.55] coordinates {(83.9, 33.3)};
\addplot[only marks, mark=*, mark size=5.78pt, color=teal!70!black, fill=teal!70!black, fill opacity=0.55] coordinates {(73.3, 72.8)};
\addplot[only marks, mark=*, mark size=2.32pt, color=cyan!45!gray, fill=cyan!45!gray, fill opacity=0.55] coordinates {(35.0, 0.0)};
\addplot[only marks, mark=square*, mark size=6.33pt, color=orange!80!black, fill=orange!80!black, fill opacity=0.55] coordinates {(96.1, 97.8)};
\addplot[only marks, mark=square*, mark size=6.80pt, color=red!70!black, fill=red!70!black, fill opacity=0.55] coordinates {(88.9, 99.4)};
\nextgroupplot[title={Easy $\cdot$ Average}]
\addplot[gray!55, dashed, no marks, forget plot] coordinates {(0,0) (100,100)};
\addplot[only marks, mark=*, mark size=3.73pt, color=blue!50!cyan, fill=blue!50!cyan, fill opacity=0.55] coordinates {(58.9, 42.6)};
\addplot[only marks, mark=*, mark size=2.57pt, color=blue!70!black, fill=blue!70!black, fill opacity=0.55] coordinates {(52.2, 24.2)};
\addplot[only marks, mark=*, mark size=4.22pt, color=teal!70!black, fill=teal!70!black, fill opacity=0.55] coordinates {(56.7, 71.5)};
\addplot[only marks, mark=*, mark size=2.53pt, color=cyan!45!gray, fill=cyan!45!gray, fill opacity=0.55] coordinates {(25.9, 33.5)};
\addplot[only marks, mark=square*, mark size=4.50pt, color=orange!80!black, fill=orange!80!black, fill opacity=0.55] coordinates {(77.1, 71.0)};
\addplot[only marks, mark=square*, mark size=5.23pt, color=red!70!black, fill=red!70!black, fill opacity=0.55] coordinates {(82.6, 95.6)};
\nextgroupplot[title={Hard $\cdot$ Deduction}]
\addplot[gray!55, dashed, no marks, forget plot] coordinates {(0,0) (100,100)};
\addplot[only marks, mark=*, mark size=1.20pt, color=blue!50!cyan, fill=blue!50!cyan, fill opacity=0.55] coordinates {(14.5, 31.0)};
\addplot[only marks, mark=*, mark size=1.20pt, color=blue!70!black, fill=blue!70!black, fill opacity=0.55] coordinates {(22.0, 13.7)};
\addplot[only marks, mark=*, mark size=1.98pt, color=teal!70!black, fill=teal!70!black, fill opacity=0.55] coordinates {(37.7, 70.0)};
\addplot[only marks, mark=*, mark size=1.20pt, color=cyan!45!gray, fill=cyan!45!gray, fill opacity=0.55] coordinates {(7.8, 24.2)};
\addplot[only marks, mark=square*, mark size=1.20pt, color=orange!80!black, fill=orange!80!black, fill opacity=0.55] coordinates {(50.5, 28.8)};
\addplot[only marks, mark=square*, mark size=4.01pt, color=red!70!black, fill=red!70!black, fill opacity=0.55] coordinates {(60.2, 89.5)};
\nextgroupplot[title={Hard $\cdot$ Induction}]
\addplot[gray!55, dashed, no marks, forget plot] coordinates {(0,0) (100,100)};
\addplot[only marks, mark=*, mark size=3.25pt, color=blue!50!cyan, fill=blue!50!cyan, fill opacity=0.55] coordinates {(38.3, 25.4)};
\addplot[only marks, mark=*, mark size=2.68pt, color=blue!70!black, fill=blue!70!black, fill opacity=0.55] coordinates {(36.8, 14.5)};
\addplot[only marks, mark=*, mark size=2.45pt, color=teal!70!black, fill=teal!70!black, fill opacity=0.55] coordinates {(36.3, 42.0)};
\addplot[only marks, mark=*, mark size=3.10pt, color=cyan!45!gray, fill=cyan!45!gray, fill opacity=0.55] coordinates {(14.5, 33.7)};
\addplot[only marks, mark=square*, mark size=4.15pt, color=orange!80!black, fill=orange!80!black, fill opacity=0.55] coordinates {(49.2, 43.5)};
\addplot[only marks, mark=square*, mark size=3.28pt, color=red!70!black, fill=red!70!black, fill opacity=0.55] coordinates {(67.9, 91.2)};
\nextgroupplot[title={Hard $\cdot$ Causal}]
\addplot[gray!55, dashed, no marks, forget plot] coordinates {(0,0) (100,100)};
\addplot[only marks, mark=*, mark size=3.69pt, color=blue!50!cyan, fill=blue!50!cyan, fill opacity=0.55] coordinates {(77.4, 43.2)};
\addplot[only marks, mark=*, mark size=2.96pt, color=blue!70!black, fill=blue!70!black, fill opacity=0.55] coordinates {(79.5, 29.5)};
\addplot[only marks, mark=*, mark size=4.65pt, color=teal!70!black, fill=teal!70!black, fill opacity=0.55] coordinates {(68.9, 41.6)};
\addplot[only marks, mark=*, mark size=1.71pt, color=cyan!45!gray, fill=cyan!45!gray, fill opacity=0.55] coordinates {(20.0, 0.0)};
\addplot[only marks, mark=square*, mark size=5.65pt, color=orange!80!black, fill=orange!80!black, fill opacity=0.55] coordinates {(96.8, 84.7)};
\addplot[only marks, mark=square*, mark size=6.26pt, color=red!70!black, fill=red!70!black, fill opacity=0.55] coordinates {(73.7, 87.9)};
\nextgroupplot[title={Hard $\cdot$ Average}]
\addplot[gray!55, dashed, no marks, forget plot] coordinates {(0,0) (100,100)};
\addplot[only marks, mark=*, mark size=2.84pt, color=blue!50!cyan, fill=blue!50!cyan, fill opacity=0.55] coordinates {(43.4, 33.2)};
\addplot[only marks, mark=*, mark size=2.31pt, color=blue!70!black, fill=blue!70!black, fill opacity=0.55] coordinates {(46.1, 19.2)};
\addplot[only marks, mark=*, mark size=3.24pt, color=teal!70!black, fill=teal!70!black, fill opacity=0.55] coordinates {(47.7, 51.2)};
\addplot[only marks, mark=*, mark size=2.04pt, color=cyan!45!gray, fill=cyan!45!gray, fill opacity=0.55] coordinates {(14.1, 19.3)};
\addplot[only marks, mark=square*, mark size=4.05pt, color=orange!80!black, fill=orange!80!black, fill opacity=0.55] coordinates {(65.5, 52.3)};
\addplot[only marks, mark=square*, mark size=4.69pt, color=red!70!black, fill=red!70!black, fill opacity=0.55] coordinates {(67.3, 89.5)};
\end{groupplot}
\node[anchor=west, inner sep=0pt, overlay] (legend) at ($(group c4r2.east)+(0.4cm,0)$) {%
\begin{tikzpicture}[every node/.style={font=\tiny, inner sep=1pt}]
\matrix[column sep=4pt, row sep=2pt]{
  \node[fill=blue!50!cyan,  draw=blue!50!cyan,  circle, minimum size=4pt, inner sep=0pt] {}; & \node {gpt-4o}; \\
  \node[fill=blue!70!black, draw=blue!70!black, circle, minimum size=4pt, inner sep=0pt] {}; & \node {llama-3.3-70b}; \\
  \node[fill=teal!70!black, draw=teal!70!black, circle, minimum size=4pt, inner sep=0pt] {}; & \node {qwen3-30b}; \\
  \node[fill=cyan!45!gray,  draw=cyan!45!gray,  circle, minimum size=4pt, inner sep=0pt] {}; & \node {olmo-3.1-32b}; \\
  \node[fill=orange!80!black, draw=orange!80!black, rectangle, minimum size=4pt, inner sep=0pt] {}; & \node {o3-mini}; \\
  \node[fill=red!70!black,  draw=red!70!black,  rectangle, minimum size=4pt, inner sep=0pt] {}; & \node {deepseek-r1}; \\
};
\end{tikzpicture}
};
\end{tikzpicture}
\caption{\textbf{Reasoning models pull ahead more on principled inference than on premise extraction.} Each point is one model in one (tier, track) cell. The $x$-coordinate is how well the model extracts premises from obfuscated documents (Obf\,$\cdot$\,NS); the $y$-coordinate is how well it reasons over clean ones (NL\,$\cdot$\,CoT); marker area scales with joint accuracy (Obf\,$\cdot$\,CoT). Points \emph{above} the diagonal are inference-leaning (the model reasons better than it extracts); points \emph{below} are extraction-leaning. Deepseek-r1 (large red squares) sits clearly above the diagonal in most cells: it is the strongest model overall and the most inference-leaning. The other reasoning model leans the same way, while most instruct models sit near or below the diagonal.}
\label{fig:probes-scatter}
\end{figure}
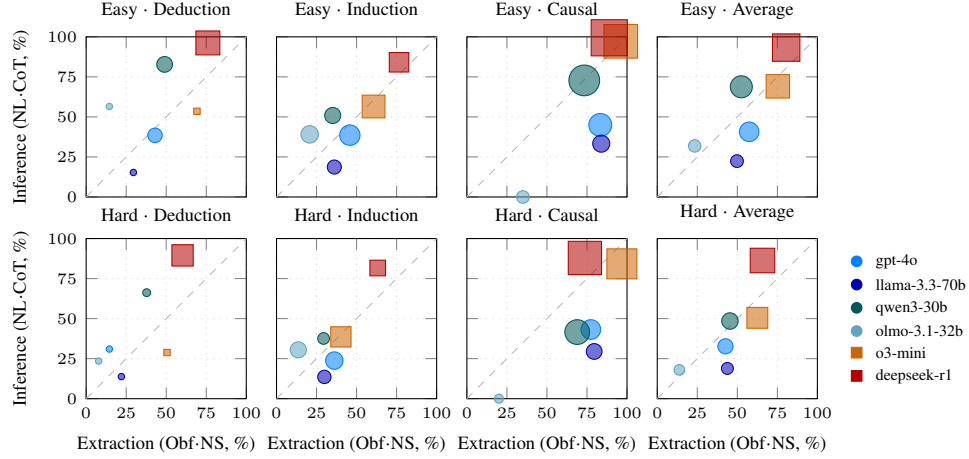

\paragraph{Three diagnostic settings give us an inference--extraction profile per model.} We read each task through three settings (Fig.~\ref{fig:probes-scatter}): \textbf{NL\,$\cdot$\,CoT} for \emph{principled inference} (premises in clean natural language, extraction is trivial), \textbf{Obf\,$\cdot$\,NS} for \emph{premise extraction} (the solver handles inference, so the score reflects finding and formalising the relevant premises in messy text, which still requires reasoning), and \textbf{Obf\,$\cdot$\,CoT} for both jointly. Underlying values are in App.~\ref{app:probes-data}.

\paragraph{Reasoning models lead on both axes, but pull further ahead on principled inference.} Reasoning models score higher than instruct models on both axes, but the gap is wider on inference: on Easy averages, deepseek-r1 leads gpt-4o by $53$\,pts on inference (NL\,$\cdot$\,CoT) versus $24$\,pts on extraction (Obf\,$\cdot$\,NS). Visually, reasoning models sit on or above the inference\,$=$\,extraction diagonal of Fig.~\ref{fig:probes-scatter}, while instruct models tend to be extraction-leaning on average, with llama-3.3-70b at the extreme. Qwen3-30b is a partial exception among instruct models: its inference scores reach reasoning-model territory, possibly reflecting its reasoning-heavy training. The pattern is cleanest for deepseek-r1. O3-mini likely shows it too, but its extraction score is plausibly inflated because it served as the obfuscation renderer (only chunks it could reverse-transform were kept). Every model scores lower on the joint task (Obf\,$\cdot$\,CoT) than on NL\,$\cdot$\,CoT or Obf\,$\cdot$\,NS alone, and the reasoning--instruct gap widens further there.

\paragraph{Comparatively, deduction is the harder track on extraction; causal on inference.} Comparing how models scatter across the three track columns of Fig.~\ref{fig:probes-scatter}, deduction points lean above the diagonal (extraction is harder than inference), causal points lean below it (inference is harder), and induction points sit near the diagonal.

\subsection{Scaling inference complexity}
\label{sec:difficulty}

\begin{figure}[ht]
\centering
\begin{tikzpicture}
\begin{groupplot}[
  group style={group size=3 by 1, horizontal sep=1.4cm},
  width=4.8cm, height=4cm,
  ymin=0, ymax=105, ytick={0,25,50,75,100},
  ymajorgrids,
  ylabel={Accuracy (\%)}, ylabel style={font=\scriptsize},
  xlabel style={font=\scriptsize, yshift=2pt},
  tick label style={font=\tiny},
  title style={font=\small, yshift=-3pt},
  enlarge x limits=0.12,
]
\nextgroupplot[title={Deduction}, xlabel={depth $\times$ distractors}, symbolic x coords={3$\times$0,4$\times$1,5$\times$2,6$\times$3,7$\times$4}, xtick=data,
  legend pos=south west,
  legend style={font=\tiny, draw=none, fill=white, fill opacity=0.85, text opacity=1, inner sep=2pt, row sep=-1pt}]
\addplot+[mark=*, color=red!70!black, thick] coordinates {(3$\times$0,88.0) (4$\times$1,69.0) (5$\times$2,52.5) (6$\times$3,46.0) (7$\times$4,20.0)};
\addlegendentry{CoT}
\addplot+[mark=square*, color=green!50!black, thick] coordinates {(3$\times$0,100.0) (4$\times$1,100.0) (5$\times$2,99.0) (6$\times$3,98.0) (7$\times$4,96.0)};
\addlegendentry{{NS}}

\nextgroupplot[title={Induction}, xlabel={distractor rules ($n_{\text{pos}}{=}2$)}, symbolic x coords={1,2,3,4,5}, xtick=data]
\addplot+[mark=*, color=red!70!black, thick] coordinates {(1,76.0) (2,63.0) (3,45.5) (4,40.0) (5,36.0)};
\addplot+[mark=square*, color=green!50!black, thick] coordinates {(1,94.0) (2,90.0) (3,74.0) (4,74.0) (5,56.0)};

\nextgroupplot[title={Causal}, xlabel={nodes $\times$ \emph{XYZ} edges}, symbolic x coords={4$\times$1,5$\times$1,6$\times$2,7$\times$3,8$\times$4}, xtick=data]
\addplot+[mark=*, color=red!70!black, thick] coordinates {(4$\times$1,98.0) (5$\times$1,98.0) (6$\times$2,85.5) (7$\times$3,64.0) (8$\times$4,62.0)};
\addplot+[mark=square*, color=green!50!black, thick] coordinates {(4$\times$1,100.0) (5$\times$1,97.5) (6$\times$2,92.0) (7$\times$3,82.0) (8$\times$4,78.0)};
\end{groupplot}
\end{tikzpicture}
\caption{\textbf{The inference-complexity axis can be dialled smoothly past where current benchmarks saturate.} Each panel scans one track's inference-complexity knob over five increasing tiers (deduction: depth $\times$ distractor trees; induction: distractor rules; causal: nodes $\times$ \emph{XYZ} edges), holding the natural-language form fixed. o3-mini CoT accuracy falls steadily on every track across the range, showing that the underlying formal generator gives us room to push tasks well past the easy regime. The neuro-symbolic curve is included for reference{; it falls less than CoT on deduction and causal, and by a similar amount on induction}.}
\label{fig:scaling}
\end{figure}
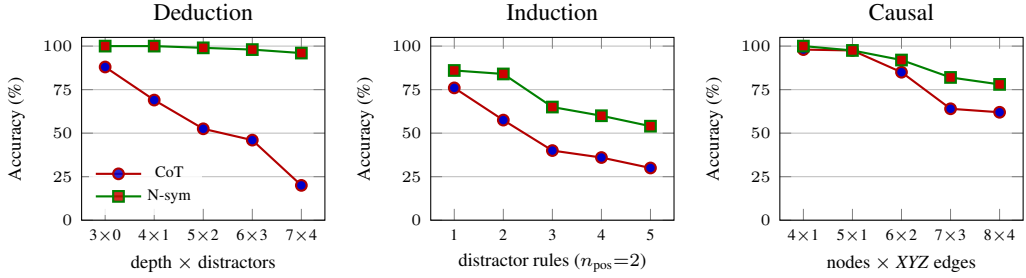

\paragraph{The pipeline produces tasks past where existing benchmarks saturate.} Existing logical-reasoning benchmarks~\citep{tafjord2021proofwriter,saparov2023prontoqa,han2024folio,liu2023logiqa2,lin2025zebralogic} have largely saturated for frontier models. Fig.~\ref{fig:scaling} shows the effect of dialling up the difficulty knob on the natural-language form alone, before the obfuscation axis is engaged: across five tiers per domain, o3-mini's CoT accuracy degrades sharply, while neuro-symbolic stays at ceiling for deduction, holds up for causal, {and stays above CoT on induction but falls by a similar amount.} This test uses one model (o3-mini), NL only, and $n{=}50$ per cell. {The second and third tiers of each track are the Easy and Hard cells of the main evaluation ($n{=}200$).} {A current frontier reasoning model, GPT-5.4, nearly saturates the Hard deduction and causal tiers evaluated here (49/50 and 50/50 on rendered tasks) but not induction (25/50), and a harder causal setting on the full Sachs graph {(clean NL, open answer)} brings it back to 30/50 (App.~\ref{app:frontier}).}

\subsection{Scientific rendering analysis}
\label{sec:obf-analysis}

\begin{table}[t]
\centering
\footnotesize
\setlength{\tabcolsep}{4pt}
\caption{{\textbf{Rendering makes the input several times longer, but the formaliser extracts no more premises.} Hard tier, averages per task. Premises are FOL clauses, ILP facts, or data rows; \emph{gold} is the number actually in the task.}}
\label{tab:obf-compact}
\begin{tabular}{@{}l c c c c c@{}}
\toprule
 & Premises & Input chars & \multicolumn{3}{c}{Premises extracted by the NS formaliser, NL$\to$Obf} \\
\cmidrule(lr){4-6}
Track & (gold) & NL$\to$Obf & deepseek-r1 & gpt-4o & olmo-3.1-32b \\
\midrule
Deduction & 32.8  & 3.2k$\to$9.4k & 38.6$\to$39.6   & 33.9$\to$39.2  & 65.7$\to$49.9 \\
Induction & 142.9 & 8.2k$\to$23k  & 138.6$\to$128.9 & 122.3$\to$48.5 & 152.2$\to$122.3 \\
Causal    & 64.0  & 4.1k$\to$17k  & 47.0$\to$48.0   & 64.0$\to$50.7  & 42.2$\to$48.2 \\
\bottomrule
\end{tabular}
\end{table}

\paragraph{Obfuscation triples to quintuples input length.} Across the six cells, the obfuscated form is on average $3$--$5\times$ longer than the natural-language form. The largest blow-up is on causal {($4.1$k$\to$$17$k characters on Hard)}, where compact data tables get unrolled into narrative reports. {The formaliser does not extract more premises from the longer input: its premise counts rise by at most 16\% and often drop (Table~\ref{tab:obf-compact}).}

\paragraph{Models miss real premises more than they absorb distractors.} {The size of the formaliser's premise list inside NS (FOL clauses, ILP facts, or data rows) shows whether models miss real premises or absorb spurious ones.} The dominant effect is the former: gpt-4o keeps only {under half} of its induction facts on Obf {($48.5$ of $142.9$ gold facts on Hard, versus $122.3$ on NL)}, and olmo-3.1-32b loses content across most cells. {Deduction is the exception: distractor sentences are mis-formalised into spurious clauses, so all three formalisers exceed the gold count on Obf (gpt-4o: $39.2$ clauses against $32.8$).} Across models, deepseek-r1 is the most stable formaliser and gpt-4o the least.

\paragraph{A qualitative look at the scientific rendering.} Looking at the appendix examples, three transformations stand out across the three tracks. \emph{Premises are buried in narrative}: a logical premise such as \texttt{forall x: Shh\_signal(x) -> FoxA2(x)} (App.~\ref{app:deduction-ex}) reappears mid-document in a stem-cell differentiation protocol as ``whenever a cell is treated with the Sonic Hedgehog (Shh) signaling factor, the cell robustly activates the transcription factor FoxA2'', with quantifier, antecedent, and consequent intact but embedded between media-prep details (``a blend of poly-D-lysine and laminin'', ``2\% B27''). \emph{A single task is split across genres}: an induction task (App.~\ref{app:induction-ex}) is delivered as a clinical case report, an in-vitro DDI study, and a PubMed abstract, where one document reports a drug--enzyme fact via clinical narrative ($n$, $p$-value) and another via $K_i$ values from a microsomal assay. \emph{Structured fields are paraphrased and reordered}: the causal task's data table is delivered as a supplementary-data deposit accompanying a paper (App.~\ref{app:causal-ex}), with column order randomised, known causal edges paraphrased into prose (\texttt{pip3 -> plc}, \texttt{pip3 -> akt} become ``pip3 plays a role in the modulation of plc and akt''), and proteins renamed by descriptor (\texttt{plc} $\to$ ``the membrane-localized phospholipase C'').

{
\subsection{Failure analysis from CoT traces}
\label{sec:cot-traces}

\begin{wraptable}{r}{0.43\linewidth}
\vspace{-1.2em}
\centering
\small
\setlength{\tabcolsep}{4pt}
\caption{Reasoning (R) vs extraction (E) errors in failed Hard\,$\cdot$\,Obf CoT traces.}
\label{tab:cot-traces}
\vspace{0.4em}
\begin{tabular}{@{}lcc@{}}
\toprule
Track & gpt-4o R/E & deepseek-r1 R/E \\
\midrule
Deduction & 18/12 & 11/19 \\
Induction & 25/5 & 4/26 \\
Causal & 17/13 & 29/0 \\
\midrule
Total & 60/30 & 44/45 \\
\bottomrule
\end{tabular}
\vspace{-1em}
\end{wraptable}
A CoT trace shows how the model reached its answer, so a failure can often be traced either to what the model read from the documents or to how it reasoned over it. We therefore sort failed traces into two categories: \emph{extraction} errors (E: a premise missed or misread, or distractor wording taken as a premise) and \emph{reasoning} errors (R: an invalid inference from correctly read evidence). For gpt-4o and deepseek-r1, an LLM judge labelled 30 failed Hard\,$\cdot$\,Obf traces per track by comparing each with the task's formal object; we then manually verified every label. App.~\ref{app:cot-traces} catalogues the specific failures we found, including recurring logical fallacies in deduction and, in induction, a tendency to prefer a familiar drug-interaction rule over the data. Induction failures are often silent, so there we count a failure as extraction if the model solves the same task in clean NL.

\paragraph{gpt-4o mostly fails by reasoning, deepseek-r1 mostly by extraction.} Most gpt-4o failures are reasoning errors, whereas most deepseek-r1 failures on deduction and induction are extraction errors (Table~\ref{tab:cot-traces}). The exception is causal discovery, where every deepseek-r1 failure is a reasoning error. This confirms the profiles of Fig.~\ref{fig:probes-scatter}: gpt-4o is extraction-leaning and deepseek-r1 inference-leaning, so each fails mostly on its weaker skill.

}

\section{Limitations}
\label{sec:limitations}

\paragraph{Real scientific corpora of documents rarely have a clean latent formal object.} SciR brings a central challenge of scientific reasoning (premise extraction from heterogeneous documents) into a controlled benchmark with verifiable answers. However, a sample of documents from a real scientific domain rarely exhibits the kind of clean underlying object SciR builds tasks from. An underlying signal exists because the data reflects the external world, but it is typically weaker. SciR is therefore only a proxy for science in the wild. The neuro-symbolic advantage we observe is therefore promising, but cannot be generalised with certainty.

\paragraph{Scope.} Each reasoning mode is instantiated by one paradigmatic case (chained syllogisms; Popper-style rule discovery; causal graphs over Sachs-style tables) and rendered onto one scientific surface (developmental biology; drug interactions; protein signalling). The single surface per mode is what makes the rendering scientifically realistic, but we do not test how the same reasoning generalises across alternative surfaces. {We also do not compare against an equally complex non-scientific rendering, so SciR does not show that scientific rendering is uniquely hard.}

\paragraph{Single-run evaluation.} The full evaluation cost roughly \$1{,}500 in API calls. Multiple seeds per cell are not feasible at our scale, so we report point estimates rather than error bars. Temperature is set to $0$ (deterministic decoding for non-reasoning models{; o3-mini and deepseek-r1 still sample internally}). The gaps we report (typically tens of percentage points) are large relative to per-cell sampling noise at $n=200$. Run-to-run variance is not measured.

{
\section{Related work}

\paragraph{Controlled reasoning benchmarks.} ProofWriter~\citep{tafjord2021proofwriter}, JustLogic~\citep{chen2025justlogic}, and SLR~\citep{helff2025slr} generate problems from formal objects, and FOLIO~\citep{han2024folio} annotates hand-written ones in first-order logic, giving exact answers and often a difficulty knob; SylloBio-NLI~\citep{wysocka2025syllobio} and CLadder~\citep{jin2023cladder} add biomedical and causal content. SciR keeps this formal control but renders the premises into heterogeneous scientific documents, making their extraction a second, tunable source of difficulty.

\paragraph{Scientific benchmarks.} GPQA~\citep{rein2024gpqa} and ScienceQA~\citep{lu2022scienceqa,saikh2022scienceqa} ask authentic scientific questions, SciFact~\citep{wadden2020scifact} verifies claims against the literature, M3SciQA~\citep{li2024m3sciqa} spreads questions over several documents, and CauSciBench~\citep{acharya2026causcibench}, LLM-SRBench~\citep{shojaee2025llmsrbench}, and DiscoveryBench~\citep{majumder2024discoverybench} target causal inference, equation discovery, and open-ended discovery. Their answers rest on human conclusions or a single paradigm, with fixed difficulty; SciR covers three paradigms with exact answers and tunable difficulty.

\paragraph{Reasoning robustness to presentation.} MuSR~\citep{sprague2024musr} renders reasoning trees into narratives, BABILong~\citep{kuratov2024babilong} hides reasoning facts in long filler text, and GSM-IC~\citep{shi2023gsmic} and GSM-$\infty$~\citep{zhou2025gsminfty} show that irrelevant context degrades reasoning, the latter with separate knobs for complexity and noise. These works use synthetic narratives or distractors; SciR renders the premises into realistic scientific documents across three paradigms. App.~\ref{app:related-extended} compares SciR with 25 prior benchmarks.
}

\section{Conclusion}

We introduced \textbf{SciR}, a benchmark spanning the three paradigmatic forms of scientific reasoning over three canonical biology setups: the Sachs signalling network, drug--drug interactions over DrugBank, and developmental-biology lineage syllogisms. Tasks are generated from formal objects and rendered into multi-document scientific discourse via a domain-tuned scheme. A cross-validated invertibility contract lets us push rendering difficulty without losing answer verifiability. Two parametric axes (inference complexity and premise obfuscation) decompose model failure into premise extraction and principled inference. Across six models, both axes hurt every model; reasoning models lead instruct models with a wider gap on principled inference, and the rendering even hurts neuro-symbolic pipelines. We release generators and evaluation code so SciR can be regenerated as models improve.

\section*{Acknowledgments}

This work was funded by the Swiss National Science Foundation (M-Rational project).

\bibliographystyle{plainnat}
\bibliography{custom}

\begin{thebibliography}{63}
\providecommand{\natexlab}[1]{#1}
\providecommand{\url}[1]{\texttt{#1}}
\expandafter\ifx\csname urlstyle\endcsname\relax
  \providecommand{\doi}[1]{doi: #1}\else
  \providecommand{\doi}{doi: \begingroup \urlstyle{rm}\Url}\fi

\bibitem[Acharya et~al.(2026)Acharya, Zhang, Kim, Shrestha, Sun, Cobben,
  Mordig, Emmerson, Haghighat, Danisman, Chen, Jose, Muresanu, Cui, Liu, Qi,
  Pandey, Huang, Sch{\"o}lkopf, and Jin]{acharya2026causcibench}
Sawal Acharya, Terry~Jingchen Zhang, Andrew Kim, Rahul~Babu Shrestha, Xianlin
  Sun, Pepijn Cobben, Maximilian Mordig, Jacob~T. Emmerson, Anahita Haghighat,
  Furkan Danisman, Yuen Chen, Clijo Jose, Andrei~Ioan Muresanu, Justin Cui,
  Jiarui Liu, Yahang Qi, Punya~Syon Pandey, Yinya Huang, Bernhard
  Sch{\"o}lkopf, and Zhijing Jin.
\newblock {CauSciBench}: Can {LLMs} automate causal inference in real-world
  scientific research?
\newblock In \emph{Proc.\ ICML}, 2026.

\bibitem[Arakelyan et~al.(2025)Arakelyan, Minervini, Lewis, Verga, and
  Augenstein]{arakelyan2025flare}
Erik Arakelyan, Pasquale Minervini, Patrick Lewis, Pat Verga, and Isabelle
  Augenstein.
\newblock {FLARE}: Faithful logic-aided reasoning and exploration.
\newblock In \emph{Proc.\ EMNLP}, 2025.
\newblock URL \url{https://aclanthology.org/2025.emnlp-main.1193/}.

\bibitem[Baumg{\"a}rtner et~al.(2025)Baumg{\"a}rtner, Briscoe, and
  Gurevych]{baumgartner2025peerqa}
Tim Baumg{\"a}rtner, Ted Briscoe, and Iryna Gurevych.
\newblock {PeerQA}: A scientific question answering dataset from peer reviews.
\newblock In \emph{Proc.\ NAACL}, 2025.
\newblock URL \url{https://aclanthology.org/2025.naacl-long.22/}.

\bibitem[Chen et~al.(2025)Chen, Zhang, and Tao]{chen2025justlogic}
Michael~K. Chen, Xikun Zhang, and Dacheng Tao.
\newblock {JustLogic}: A comprehensive benchmark for evaluating deductive
  reasoning in large language models.
\newblock \emph{arXiv preprint arXiv:2501.14851}, 2025.

\bibitem[Chen et~al.(2024)Chen, Chi, Wang, and Zhou]{chen2024premise}
Xinyun Chen, Ryan~A. Chi, Xuezhi Wang, and Denny Zhou.
\newblock Premise order matters in reasoning with large language models.
\newblock In \emph{Proc.\ ICML}, 2024.

\bibitem[Cropper and Morel(2021)]{cropper2021popper}
Andrew Cropper and Rolf Morel.
\newblock Learning programs by learning from failures.
\newblock \emph{Machine Learning}, 110:\penalty0 801--856, 2021.

\bibitem[Dalvi et~al.(2021)Dalvi, Jansen, Tafjord, Xie, Smith, Pipatanangkura,
  and Clark]{dalvi2021entailmentbank}
Bhavana Dalvi, Peter Jansen, Oyvind Tafjord, Zhengnan Xie, Hannah Smith,
  Leighanna Pipatanangkura, and Peter Clark.
\newblock Explaining answers with entailment trees.
\newblock In \emph{Proc.\ EMNLP}, 2021.

\bibitem[Dasigi et~al.(2021)Dasigi, Lo, Beltagy, Cohan, Smith, and
  Gardner]{dasigi2021qasper}
Pradeep Dasigi, Kyle Lo, Iz~Beltagy, Arman Cohan, Noah~A. Smith, and Matt
  Gardner.
\newblock A dataset of information-seeking questions and answers anchored in
  research papers.
\newblock In \emph{Proc.\ NAACL-HLT}, 2021.
\newblock URL \url{https://aclanthology.org/2021.naacl-main.365/}.

\bibitem[Dhami et~al.(2018)Dhami, Kunapuli, Das, Page, and
  Natarajan]{dhami2018drugdrug}
Devendra~Singh Dhami, Gautam Kunapuli, Mayukh Das, David Page, and Sriraam
  Natarajan.
\newblock Drug-drug interaction discovery: kernel learning from heterogeneous
  similarities.
\newblock \emph{Smart Health}, 9:\penalty0 88--100, 2018.

\bibitem[Gallifant et~al.(2024)Gallifant, Chen, Moreira, Munch, Gao, Pond,
  Celi, Aerts, Hartvigsen, and Bitterman]{gallifant2024rabbits}
Jack Gallifant, Shan Chen, Pedro Jos{\'e}~Ferreira Moreira, Nikolaj Munch,
  Mingye Gao, Jackson Pond, Leo~Anthony Celi, Hugo Aerts, Thomas Hartvigsen,
  and Danielle Bitterman.
\newblock Language models are surprisingly fragile to drug names in biomedical
  benchmarks.
\newblock In \emph{Findings of EMNLP}, 2024.

\bibitem[Han et~al.(2024)]{han2024folio}
Simeng Han et~al.
\newblock {FOLIO}: Natural language reasoning with first-order logic.
\newblock In \emph{Proc.\ EMNLP}, 2024.

\bibitem[Hauser and B{\"u}hlmann(2012)]{hauser2012gies}
Alain Hauser and Peter B{\"u}hlmann.
\newblock Characterization and greedy learning of interventional {Markov}
  equivalence classes of directed acyclic graphs.
\newblock \emph{Journal of Machine Learning Research}, 13:\penalty0 2409--2464,
  2012.

\bibitem[He et~al.(2026)He, Huang, Sachan, and Jin]{he2026doverifier}
Paul He, Yinya Huang, Mrinmaya Sachan, and Zhijing Jin.
\newblock Uncovering hidden correctness in {LLM} causal reasoning via symbolic
  verification.
\newblock In \emph{Proc.\ EACL}, 2026.

\bibitem[Helff et~al.(2026{\natexlab{a}})Helff, Delfosse, Steinmann, H{\"a}rle,
  Shindo, Schramowski, Stammer, Kersting, and Friedrich]{helff2026gaming}
Lukas Helff, Quentin Delfosse, David Steinmann, Ruben H{\"a}rle, Hikaru Shindo,
  Patrick Schramowski, Wolfgang Stammer, Kristian Kersting, and Felix
  Friedrich.
\newblock {LLMs} gaming verifiers: {RLVR} can lead to reward hacking.
\newblock \emph{arXiv preprint arXiv:2604.15149}, 2026{\natexlab{a}}.

\bibitem[Helff et~al.(2026{\natexlab{b}})Helff, Omar, Friedrich, W{\"u}st,
  Shindo, Mitchell, Woydt, Schramowski, Stammer, and Kersting]{helff2025slr}
Lukas Helff, Ahmad Omar, Felix Friedrich, Antonia W{\"u}st, Hikaru Shindo,
  Rupert Mitchell, Tim Woydt, Patrick Schramowski, Wolfgang Stammer, and
  Kristian Kersting.
\newblock {SLR}: Automated synthesis for scalable logical reasoning.
\newblock In \emph{Proc.\ ACL}, 2026{\natexlab{b}}.

\bibitem[Jin et~al.(2023)Jin, Chen, Leeb, Gresele, Kamal, Lyu, Blin,
  Gonzalez~Adauto, Kleiman-Weiner, Sachan, and Sch{\"o}lkopf]{jin2023cladder}
Zhijing Jin, Yuen Chen, Felix Leeb, Luigi Gresele, Ojasv Kamal, Zhiheng Lyu,
  Kevin Blin, Fernando Gonzalez~Adauto, Max Kleiman-Weiner, Mrinmaya Sachan,
  and Bernhard Sch{\"o}lkopf.
\newblock {CLadder}: Assessing causal reasoning in language models.
\newblock In \emph{Advances in Neural Information Processing Systems
  (NeurIPS)}, 2023.
\newblock URL
  \url{https://proceedings.neurips.cc/paper_files/paper/2023/hash/631bb9434d718ea309af82566347d607-Abstract-Conference.html}.

\bibitem[Jin et~al.(2024)Jin, Liu, Lyu, Poff, Sachan, Mihalcea, Diab, and
  Sch{\"o}lkopf]{jin2024corr2cause}
Zhijing Jin, Jiarui Liu, Zhiheng Lyu, Spencer Poff, Mrinmaya Sachan, Rada
  Mihalcea, Mona Diab, and Bernhard Sch{\"o}lkopf.
\newblock Can large language models infer causation from correlation?
\newblock In \emph{Proc.\ ICLR}, 2024.
\newblock URL \url{https://openreview.net/forum?id=vqIH0ObdqL}.

\bibitem[Khandekar et~al.(2024)Khandekar, Jin, Xiong,
  et~al.]{khandekar2024medcalc}
Nikhil Khandekar, Qiao Jin, Guangzhi Xiong, et~al.
\newblock {MedCalc-Bench}: Evaluating large language models for medical
  calculations.
\newblock In \emph{NeurIPS Datasets and Benchmarks Track}, 2024.

\bibitem[K{\i}c{\i}man et~al.(2024)K{\i}c{\i}man, Ness, Sharma, and
  Tan]{kiciman2024causal}
Emre K{\i}c{\i}man, Robert Ness, Amit Sharma, and Chenhao Tan.
\newblock Causal reasoning and large language models: Opening a new frontier
  for causality.
\newblock \emph{Transactions on Machine Learning Research (TMLR)}, 2024.
\newblock URL \url{https://openreview.net/forum?id=mqoxLkX210}.

\bibitem[Knox et~al.(2024)]{knox2024drugbank}
Craig Knox et~al.
\newblock {DrugBank} 6.0: The {DrugBank} knowledgebase for 2024.
\newblock \emph{Nucleic Acids Research}, 2024.

\bibitem[Kuratov et~al.(2024)Kuratov, Bulatov, Anokhin, Rodkin, Sorokin,
  Sorokin, and Burtsev]{kuratov2024babilong}
Yuri Kuratov, Aydar Bulatov, Petr Anokhin, Ivan Rodkin, Dmitry Sorokin, Artyom
  Sorokin, and Mikhail Burtsev.
\newblock {BABILong}: Testing the limits of {LLMs} with long context
  reasoning-in-a-haystack.
\newblock In \emph{NeurIPS Datasets and Benchmarks Track}, 2024.

\bibitem[Levy et~al.(2024)Levy, Jacoby, and Goldberg]{levy2024sametask}
Mosh Levy, Alon Jacoby, and Yoav Goldberg.
\newblock Same task, more tokens: the impact of input length on the reasoning
  performance of large language models.
\newblock In \emph{Proc.\ ACL}, 2024.

\bibitem[Li et~al.(2024)Li, Shangguan, Zhao, Li, Liu, and Cohan]{li2024m3sciqa}
Chuhan Li, Ziyao Shangguan, Yilun Zhao, Deyuan Li, Yixin Liu, and Arman Cohan.
\newblock {M3SciQA}: A multi-modal multi-document scientific {QA} benchmark for
  evaluating foundation models.
\newblock In \emph{Findings of EMNLP}, 2024.

\bibitem[Lin et~al.(2025{\natexlab{a}})]{lin2025zebralogic}
Bill~Yuchen Lin et~al.
\newblock {ZebraLogic}: On the scaling limits of {LLMs} for logical reasoning.
\newblock In \emph{Proc.\ ICML}, 2025{\natexlab{a}}.

\bibitem[Lin et~al.(2025{\natexlab{b}})Lin, Yuan, Zhou, Wang, Wang, Kong, Shi,
  Li, Yang, Liu, and Sun]{lin2025sirbench}
Brian~S. Lin, Jiaxin Yuan, Zihan Zhou, Shouli Wang, Shuo Wang, Cunliang Kong,
  Qi~Shi, Yuxuan Li, Liner Yang, Zhiyuan Liu, and Maosong Sun.
\newblock On {LLM}-based scientific inductive reasoning beyond equations.
\newblock In \emph{Proc.\ EMNLP}, 2025{\natexlab{b}}.
\newblock URL \url{https://aclanthology.org/2025.emnlp-main.476/}.

\bibitem[Liu et~al.(2023)Liu, Liu, Cui, Teng, Duan, Zhou, and
  Zhang]{liu2023logiqa2}
Hanmeng Liu, Jian Liu, Leyang Cui, Zhiyang Teng, Nan Duan, Ming Zhou, and Yue
  Zhang.
\newblock {LogiQA} 2.0 --- an improved dataset for logical reasoning in natural
  language understanding.
\newblock \emph{IEEE/ACM Transactions on Audio, Speech, and Language
  Processing}, 2023.

\bibitem[Lu et~al.(2024)Lu, Lu, Lange, Foerster, Clune, and
  Ha]{lu2024aiscientist}
Chris Lu, Cong Lu, Robert~Tjarko Lange, Jakob Foerster, Jeff Clune, and David
  Ha.
\newblock The {AI} scientist: Towards fully automated open-ended scientific
  discovery.
\newblock \emph{arXiv preprint arXiv:2408.06292}, 2024.
\newblock URL \url{https://arxiv.org/abs/2408.06292}.

\bibitem[Lu et~al.(2022)Lu, Mishra, Xia, Qiu, Chang, Zhu, Tafjord, Clark, and
  Kalyan]{lu2022scienceqa}
Pan Lu, Swaroop Mishra, Tony Xia, Liang Qiu, Kai-Wei Chang, Song-Chun Zhu,
  Oyvind Tafjord, Peter Clark, and Ashwin Kalyan.
\newblock Learn to explain: Multimodal reasoning via thought chains for science
  question answering.
\newblock In \emph{Advances in Neural Information Processing Systems
  (NeurIPS)}, 2022.
\newblock URL
  \url{https://proceedings.neurips.cc/paper_files/paper/2022/hash/11332b6b6cf4485b84afadb1352d3a9a-Abstract-Conference.html}.

\bibitem[Lu et~al.(2023)Lu, Pan, Liu, Nakov, and Kan]{lu2023scitab}
Xinyuan Lu, Liangming Pan, Qian Liu, Preslav Nakov, and Min-Yen Kan.
\newblock {SCITAB}: A challenging benchmark for compositional reasoning and
  claim verification on scientific tables.
\newblock In \emph{Proc.\ EMNLP}, 2023.
\newblock URL \url{https://aclanthology.org/2023.emnlp-main.483/}.

\bibitem[Ma et~al.(2025)Ma, Du, Wang, Zhang, Wen, Qu, Yang, Liu, Liu, Yue,
  Huang, and Zhang]{ma2025korbench}
Kaijing Ma, Xinrun Du, Yunran Wang, Haoran Zhang, Zhoufutu Wen, Xingwei Qu,
  Jian Yang, Jiaheng Liu, Minghao Liu, Xiang Yue, Wenhao Huang, and Ge~Zhang.
\newblock {KOR-Bench}: Benchmarking language models on knowledge-orthogonal
  reasoning tasks.
\newblock In \emph{International Conference on Learning Representations
  (ICLR)}, 2025.

\bibitem[Ma et~al.(2024)Ma, Gou, Hao, Xu, Wang, Pan, Yang, Cao, Sun, Awadalla,
  and Chen]{ma2024sciagent}
Yubo Ma, Zhibin Gou, Junheng Hao, Ruochen Xu, Shuohang Wang, Liangming Pan,
  Yujiu Yang, Yixin Cao, Aixin Sun, Hany Awadalla, and Weizhu Chen.
\newblock {SciAgent}: Tool-augmented language models for scientific reasoning.
\newblock In \emph{Proc.\ EMNLP}, 2024.
\newblock URL \url{https://aclanthology.org/2024.emnlp-main.880/}.

\bibitem[Majumder et~al.(2025)Majumder, Surana, Agarwal, Dalvi~Mishra, Meena,
  Prakhar, Vora, Khot, Sabharwal, and Clark]{majumder2024discoverybench}
Bodhisattwa~Prasad Majumder, Harshit Surana, Dhruv Agarwal, Bhavana
  Dalvi~Mishra, Abhijeetsingh Meena, Aryan Prakhar, Tirth Vora, Tushar Khot,
  Ashish Sabharwal, and Peter Clark.
\newblock {DiscoveryBench}: Towards data-driven discovery with large language
  models.
\newblock In \emph{Proc.\ ICLR}, 2025.

\bibitem[McCune(2005--2010)]{mccune2005prover9}
William McCune.
\newblock {Prover9} and {Mace4}, 2005--2010.
\newblock \url{https://www.cs.unm.edu/~mccune/prover9/}.

\bibitem[Mirzadeh et~al.(2025)Mirzadeh, Alizadeh, Shahrokhi, Tuzel, Bengio, and
  Farajtabar]{mirzadeh2025gsmsymbolic}
Iman Mirzadeh, Keivan Alizadeh, Hooman Shahrokhi, Oncel Tuzel, Samy Bengio, and
  Mehrdad Farajtabar.
\newblock {GSM-Symbolic}: Understanding the limitations of mathematical
  reasoning in large language models.
\newblock In \emph{Proc.\ ICLR}, 2025.

\bibitem[Modarressi et~al.(2025)Modarressi, Deilamsalehy, Dernoncourt, Bui,
  Rossi, Yoon, and Sch{\"u}tze]{modarressi2025nolima}
Ali Modarressi, Hanieh Deilamsalehy, Franck Dernoncourt, Trung Bui, Ryan~A.
  Rossi, Seunghyun Yoon, and Hinrich Sch{\"u}tze.
\newblock {NoLiMa}: Long-context evaluation beyond literal matching.
\newblock In \emph{Proc.\ ICML}, 2025.

\bibitem[Morishita et~al.(2023)Morishita, Morio, Yamaguchi, and
  Sogawa]{morishita2023fld}
Terufumi Morishita, Gaku Morio, Atsuki Yamaguchi, and Yasuhiro Sogawa.
\newblock Learning deductive reasoning from synthetic corpus based on formal
  logic.
\newblock In \emph{Proc.\ ICML}, 2023.

\bibitem[Olausson et~al.(2023)Olausson, Gu, Lipkin, Zhang, Solar-Lezama,
  Tenenbaum, and Levy]{olausson2023linc}
Theo~X. Olausson, Alex Gu, Ben Lipkin, Cedegao~E. Zhang, Armando Solar-Lezama,
  Joshua~B. Tenenbaum, and Roger Levy.
\newblock {LINC}: A neurosymbolic approach for logical reasoning by combining
  language models with first-order logic provers.
\newblock In \emph{Proc.\ EMNLP}, 2023.
\newblock URL \url{https://aclanthology.org/2023.emnlp-main.313/}.

\bibitem[Pan et~al.(2023)Pan, Albalak, Wang, and Wang]{pan2023logiclm}
Liangming Pan, Alon Albalak, Xinyi Wang, and William~Yang Wang.
\newblock Logic-{LM}: Empowering large language models with symbolic solvers
  for faithful logical reasoning.
\newblock In \emph{Findings of the ACL: EMNLP 2023}, 2023.
\newblock URL \url{https://aclanthology.org/2023.findings-emnlp.248/}.

\bibitem[Qiu et~al.(2024)Qiu, Jiang, Lu, Sclar, Pyatkin, Bhagavatula, Wang,
  Kim, Choi, Dziri, and Ren]{qiu2024phenomenal}
Linlu Qiu, Liwei Jiang, Ximing Lu, Melanie Sclar, Valentina Pyatkin, Chandra
  Bhagavatula, Bailin Wang, Yoon Kim, Yejin Choi, Nouha Dziri, and Xiang Ren.
\newblock Phenomenal yet puzzling: Testing inductive reasoning capabilities of
  language models with hypothesis refinement.
\newblock In \emph{Proc.\ ICLR}, 2024.
\newblock URL \url{https://openreview.net/forum?id=bNt7oajl2a}.

\bibitem[Ranaldi et~al.(2025)Ranaldi, Valentino, and
  Freitas]{ranaldi2025quasar}
Leonardo Ranaldi, Marco Valentino, and Andre Freitas.
\newblock Improving chain-of-thought reasoning via quasi-symbolic abstractions.
\newblock In \emph{Proc.\ ACL}, 2025.
\newblock URL \url{https://aclanthology.org/2025.acl-long.843/}.

\bibitem[Rein et~al.(2024)Rein, Hou, Stickland, Petty, Pang, Dirani, Michael,
  and Bowman]{rein2024gpqa}
David Rein, Betty~Li Hou, Asa~Cooper Stickland, Jackson Petty, Richard~Yuanzhe
  Pang, Julien Dirani, Julian Michael, and Samuel~R. Bowman.
\newblock {GPQA}: A graduate-level {Google}-proof {Q\&A} benchmark.
\newblock In \emph{Conference on Language Modeling (COLM)}, 2024.

\bibitem[Sachs et~al.(2005)Sachs, Perez, Pe'er, Lauffenburger, and
  Nolan]{sachs2005causal}
Karen Sachs, Omar Perez, Dana Pe'er, Douglas~A. Lauffenburger, and Garry~P.
  Nolan.
\newblock Causal protein-signaling networks derived from multiparameter
  single-cell data.
\newblock \emph{Science}, 308\penalty0 (5721):\penalty0 523--529, 2005.
\newblock \doi{10.1126/science.1105809}.

\bibitem[Saikh et~al.(2022)Saikh, Ghosal, Mittal, Ekbal, and
  Bhattacharyya]{saikh2022scienceqa}
Tanik Saikh, Tirthankar Ghosal, Amish Mittal, Asif Ekbal, and Pushpak
  Bhattacharyya.
\newblock {ScienceQA}: A novel resource for question answering on scholarly
  articles.
\newblock \emph{International Journal on Digital Libraries}, 23\penalty0
  (3):\penalty0 289--301, 2022.

\bibitem[Saparov and He(2023)]{saparov2023prontoqa}
Abulhair Saparov and He~He.
\newblock Language models are greedy reasoners: A systematic formal analysis of
  chain-of-thought.
\newblock In \emph{Proc.\ ICLR}, 2023.

\bibitem[Shi et~al.(2023)Shi, Chen, Misra, Scales, Dohan, Chi, Sch{\"a}rli, and
  Zhou]{shi2023gsmic}
Freda Shi, Xinyun Chen, Kanishka Misra, Nathan Scales, David Dohan, Ed~Chi,
  Nathanael Sch{\"a}rli, and Denny Zhou.
\newblock Large language models can be easily distracted by irrelevant context.
\newblock In \emph{Proc.\ ICML}, 2023.

\bibitem[Shojaee et~al.(2025)Shojaee, Nguyen, Meidani, Barati~Farimani, Doan,
  and Reddy]{shojaee2025llmsrbench}
Parshin Shojaee, Ngoc-Hieu Nguyen, Kazem Meidani, Amir Barati~Farimani, Khoa~D.
  Doan, and Chandan~K. Reddy.
\newblock {LLM-SRBench}: A new benchmark for scientific equation discovery with
  large language models.
\newblock In \emph{Proc. ICML}, 2025.

\bibitem[Sinha et~al.(2019)Sinha, Sodhani, Dong, Pineau, and
  Hamilton]{sinha2019clutrr}
Koustuv Sinha, Shagun Sodhani, Jin Dong, Joelle Pineau, and William~L.
  Hamilton.
\newblock {CLUTRR}: A diagnostic benchmark for inductive reasoning from text.
\newblock In \emph{Proc.\ EMNLP-IJCNLP}, 2019.

\bibitem[Sprague et~al.(2024)Sprague, Ye, Bostrom, Chaudhuri, and
  Durrett]{sprague2024musr}
Zayne Sprague, Xi~Ye, Kaj Bostrom, Swarat Chaudhuri, and Greg Durrett.
\newblock {MuSR}: Testing the limits of chain-of-thought with multistep soft
  reasoning.
\newblock In \emph{Proc.\ ICLR}, 2024.

\bibitem[Sun et~al.(2024)Sun, Han, Zhao, Ma, Shen, Chen, Chen, and
  Yu]{sun2024scieval}
Liangtai Sun, Yang Han, Zihan Zhao, Da~Ma, Zhennan Shen, Baocai Chen, Lu~Chen,
  and Kai Yu.
\newblock {SciEval}: A multi-level large language model evaluation benchmark
  for scientific research.
\newblock In \emph{Proc.\ AAAI}, 2024.
\newblock URL \url{https://ojs.aaai.org/index.php/AAAI/article/view/29872}.

\bibitem[Tafjord et~al.(2021)Tafjord, Dalvi, and Clark]{tafjord2021proofwriter}
Oyvind Tafjord, Bhavana Dalvi, and Peter Clark.
\newblock {ProofWriter}: Generating implications, proofs, and abductive
  statements over natural language.
\newblock In \emph{Findings of ACL}, 2021.

\bibitem[Wadden et~al.(2020)Wadden, Lin, Lo, Wang, van Zuylen, Cohan, and
  Hajishirzi]{wadden2020scifact}
David Wadden, Shanchuan Lin, Kyle Lo, Lucy~Lu Wang, Madeleine van Zuylen, Arman
  Cohan, and Hannaneh Hajishirzi.
\newblock Fact or fiction: Verifying scientific claims.
\newblock In \emph{Proc.\ EMNLP}, 2020.
\newblock URL \url{https://aclanthology.org/2020.emnlp-main.609/}.

\bibitem[Wang et~al.(2024)Wang, Zelikman, Poesia, Pu, Haber, and
  Goodman]{wang2024hypothesis}
Ruocheng Wang, Eric Zelikman, Gabriel Poesia, Yewen Pu, Nick Haber, and Noah~D.
  Goodman.
\newblock Hypothesis search: Inductive reasoning with language models.
\newblock In \emph{Proc.\ ICLR}, 2024.
\newblock URL \url{https://openreview.net/forum?id=G7UtIGQmjm}.

\bibitem[Wei et~al.(2025)Wei, Wu, Wan, Suresh, Tan, Zhou, Koyejo, Wang, and
  Aiken]{wei2025satbench}
Anjiang Wei, Yuheng Wu, Yingjia Wan, Tarun Suresh, Huanmi Tan, Zhanke Zhou,
  Sanmi Koyejo, Ke~Wang, and Alex Aiken.
\newblock {SATBench}: Benchmarking {LLMs}' logical reasoning via automated
  puzzle generation from {SAT} formulas.
\newblock In \emph{Proc.\ EMNLP}, 2025.

\bibitem[Welbl et~al.(2017)Welbl, Liu, and Gardner]{welbl2017sciq}
Johannes Welbl, Nelson~F. Liu, and Matt Gardner.
\newblock Crowdsourcing multiple choice science questions.
\newblock In \emph{Proceedings of the 3rd Workshop on Noisy User-generated
  Text}, 2017.

\bibitem[Weston et~al.(2016)Weston, Bordes, Chopra, Rush, van Merri{\"e}nboer,
  Joulin, and Mikolov]{weston2015babi}
Jason Weston, Antoine Bordes, Sumit Chopra, Alexander~M. Rush, Bart van
  Merri{\"e}nboer, Armand Joulin, and Tomas Mikolov.
\newblock Towards {AI}-complete question answering: A set of prerequisite toy
  tasks.
\newblock In \emph{Proc.\ ICLR}, 2016.

\bibitem[Wysocka et~al.(2025)Wysocka, Carvalho, Wysocki, Valentino, and
  Freitas]{wysocka2025syllobio}
Magdalena Wysocka, Danilo Carvalho, Oskar Wysocki, Marco Valentino, and Andre
  Freitas.
\newblock {SylloBio-NLI}: Evaluating large language models on biomedical
  syllogistic reasoning.
\newblock In \emph{Proc.\ NAACL}, 2025.
\newblock URL \url{https://aclanthology.org/2025.naacl-long.371/}.

\bibitem[Xu et~al.(2024)Xu, Fei, Pan, Liu, Lee, and Hsu]{xu2024symbcot}
Jundong Xu, Hao Fei, Liangming Pan, Qian Liu, Mong-Li Lee, and Wynne Hsu.
\newblock Faithful logical reasoning via symbolic chain-of-thought.
\newblock In \emph{Proc. ACL}, 2024.

\bibitem[Xu et~al.(2026)Xu, Beckmann, Valentino, and Freitas]{xu2026adaptive}
Lei Xu, Pierre Beckmann, Marco Valentino, and Andre Freitas.
\newblock Adaptive {LLM}-symbolic reasoning via dynamic logical solver
  composition.
\newblock In \emph{Proc.\ EACL}, 2026.
\newblock URL \url{https://aclanthology.org/2026.eacl-long.54/}.

\bibitem[Yang et~al.(2018)Yang, Qi, Zhang, Bengio, Cohen, Salakhutdinov, and
  Manning]{yang2018hotpotqa}
Zhilin Yang, Peng Qi, Saizheng Zhang, Yoshua Bengio, William Cohen, Ruslan
  Salakhutdinov, and Christopher~D. Manning.
\newblock {HotpotQA}: A dataset for diverse, explainable multi-hop question
  answering.
\newblock In \emph{Proc. EMNLP}, 2018.

\bibitem[Young et~al.(2022)Young, Bao, Bensemann, and
  Witbrock]{young2022abductionrules}
Nathan Young, Qiming Bao, Joshua Bensemann, and Michael Witbrock.
\newblock {AbductionRules}: Training transformers to explain unexpected inputs.
\newblock In \emph{Findings of the Association for Computational Linguistics:
  ACL}, 2022.

\bibitem[Zhong et~al.(2022)Zhong, Wang, Tang, Xu, Guo, Wang, Yin, Zhou, and
  Duan]{zhong2022arlsat}
Wanjun Zhong, Siyuan Wang, Duyu Tang, Zenan Xu, Daya Guo, Jiahai Wang, Jian
  Yin, Ming Zhou, and Nan Duan.
\newblock {AR-LSAT}: Investigating analytical reasoning of text.
\newblock In \emph{Findings of NAACL}, 2022.

\bibitem[Zhou et~al.(2025{\natexlab{a}})Zhou, Liu, Chen, Tian, and
  Chen]{zhou2025gsminfty}
Yang Zhou, Hongyi Liu, Zhuoming Chen, Yuandong Tian, and Beidi Chen.
\newblock {GSM}-$\infty$: How do your {LLMs} behave over infinitely increasing
  reasoning complexity and context length?
\newblock In \emph{Proc.\ ICML}, 2025{\natexlab{a}}.

\bibitem[Zhou et~al.(2025{\natexlab{b}})Zhou, Ye, Ling, Han, Huang, Zhuang,
  Liang, Guo, Guo, Wang, and Zhang]{zhou2025finelogic}
Yujun Zhou, Jiayi Ye, Zipeng Ling, Yufei Han, Yue Huang, Haomin Zhuang, Zhenwen
  Liang, Kehan Guo, Taicheng Guo, Xiangqi Wang, and Xiangliang Zhang.
\newblock Dissecting logical reasoning in {LLMs}: A fine-grained evaluation and
  supervision study.
\newblock In \emph{Findings of the Association for Computational Linguistics:
  EMNLP}, 2025{\natexlab{b}}.

\end{thebibliography}

\newpage
\appendix
\raggedbottom

\FloatBarrier
\section{End-to-end task examples}
\label{app:examples}

{Prompts are abridged. For instance, the full induction prompt also states the answer criterion (a valid rule covers at least one observed interacting pair and no non-interacting pair); full prompts are in the released code.}

\FloatBarrier
\subsection{Deduction}
\label{app:deduction-ex}


\begin{tcolorbox}[examplebox, title=Deductive Reasoning -- Structured Format, colframe=blue!50!black]
\scriptsize
\textbf{Prompt:} You are given a set of premises expressed in first-order logic. Determine whether the hypothesis follows from the premises.

\medskip
\textbf{Premises:}
\begin{Verbatim}[fontsize=\tiny]
forall x: epiblast_transition_signal(x) -> epiblast_thickening_process(x)
forall x: primitive_streak_formation(x) -> (mesoderm_specification(x) or epiblast_transition_signal(x))
forall x: cell_adhesion_loss(x) -> not epithelial_to_mesenchymal_transition_inhibition(x)
forall x: not epithelial_to_mesenchymal_transition_inhibition(x) -> mesoderm_committed_state(x)
forall x: mesoderm_specification(x) -> epiblast_thickening_process(x)
forall x: blastula_stage_signaling(x) -> (cell_adhesion_loss(x) or endoderm_lineage_specification(x))
forall x: epiblast_state(x) -> (primitive_streak_formation(x) or blastula_stage_signaling(x))
forall x: primitive_streak_formation(x) -> migration_induction_signal(x)
forall x: (epiblast_thickening_process(x) and migration_induction_signal(x)) -> mesoderm_committed_state(x)
forall x: endoderm_lineage_specification(x) -> not epithelial_to_mesenchymal_transition_inhibition(x)
\end{Verbatim}

\textbf{Hypothesis:}
\begin{Verbatim}[fontsize=\tiny]
forall x: epiblast_state(x) -> mesoderm_committed_state(x)
\end{Verbatim}

\textbf{Options:} valid \quad|\quad invalid \quad|\quad unknown

\smallskip
\textbf{Answer:} valid
\end{tcolorbox}


\begin{tcolorbox}[examplebox, title=Deductive Reasoning -- Natural Language Format, colframe=green!50!black]
\scriptsize
\textbf{Prompt:} You are given a set of premises about embryonic cell states. Determine whether the hypothesis follows from the premises.

\medskip
\textbf{Document 1:}

\medskip
\tiny
Every member of epiblast transition signal is a member of epiblast thickening process.\par
Every member of primitive streak formation is either a member of mesoderm specification or a member of epiblast transition signal, or both.\par
Whatever is a member of Cell Adhesion Loss, is not a member of epithelial to mesenchymal transition inhibition.\par
Whatever is not a member of epithelial to mesenchymal transition inhibition, is a member of mesoderm committed state.\par
Every member of mesoderm specification is a member of epiblast thickening process.

\medskip
\scriptsize
\textbf{Document 2:}

\medskip
\tiny
Every member of blastula stage signaling is either a member of Cell Adhesion Loss or a member of endoderm lineage specification, or both.\par
Every member of epiblast state is either a member of primitive streak formation or a member of blastula stage signaling, or both.\par
Every member of primitive streak formation is a member of migration induction signal.\par
Every member of epiblast thickening process that is a member of migration induction signal is also a member of mesoderm committed state.\par
Whatever is a member of endoderm lineage specification, is not a member of epithelial to mesenchymal transition inhibition.

\medskip
\scriptsize
\textbf{Hypothesis:} Every member of epiblast state is a member of mesoderm committed state.

\smallskip
\textbf{Options:} valid \quad|\quad invalid \quad|\quad unknown

\smallskip
\textbf{Answer:} valid
\end{tcolorbox}


\begin{tcolorbox}[examplebox, title=Deductive Reasoning -- Obfuscated Format (Document 1 of 2: RNA-Seq Differential Expression Annotation), colframe=red!50!black]
\tiny

Differential RNA-Seq Analysis of Embryonic Cell State Transitions\par
\noindent\rule{\linewidth}{0.4pt}\par
Dr.\ Maria Gonzalez, Dr.\ Ethan Park, and the Developmental Genomics Group

\medskip
\textbf{Abstract:}
In our comprehensive RNA-Seq analysis of embryonic cell populations isolated from murine models, we have consistently observed robust patterns of gene expression that align with known developmental transitions. The cell samples were cultured in DMEM supplemented with 10\% fetal bovine serum at 37\textdegree C and processed using the Illumina TruSeq RNA library preparation protocol. QC metrics, including an average RNA Integrity Number (RIN) of 9.2 determined via Agilent 2100 Bioanalyzer, confirmed the high quality of our RNA samples.

It is very likely that for every cell exhibiting the signal indicative of an epiblast transition, there is a concurrent activation of the epiblast thickening process. This robust observation aligns with the changes in the RNA expression profiles obtained under both standard and stress-induced culture conditions.

Our findings also indicate that for every cell presenting characteristics of primitive streak formation, the expression data reveal that either mesoderm specification is present or the epiblast transition signal is detected. This disjunctive outcome was consistently validated across multiple replicate experiments, including conditions with varied incubation times (ranging from 12 to 24 hours).

In addition, experimental observations under conditions of altered extracellular matrix composition (including treatment with 5\,$\mu$g/mL fibronectin) reveal that for every cell where there is a significant loss in cell adhesion, there is an unequivocal absence of inhibition of the epithelial-to-mesenchymal transition. It should be noted that the technical replicates were processed using a standardized buffer (PBS with 0.1\% BSA) to minimize variance in cell adhesion markers.

Furthermore, our differential expression analysis robustly supports that for every cell in which inhibition of the epithelial-to-mesenchymal transition is not present, these cells are clearly committed to a mesoderm state. This conclusion was consistently observed using both DESeq2 and edgeR analyses, with a fold change cutoff of 2 and an adjusted $p$-value threshold of 0.05.

Finally, evidence strongly suggests that for every cell undergoing mesoderm specification, there is also a consistent involvement in the epiblast thickening process. This correlation was observed under multiple experimental conditions, including varied serum concentrations (ranging from 5\% to 15\%) and consistent across both bulk RNA-seq and single-cell RNA-seq datasets.

\medskip
Additional technical notes:
\begin{itemize}[nosep, leftmargin=2em]
\item The sequencing runs were performed on an Illumina NovaSeq 6000 system with 150\,bp paired-end reads.
\item All samples were subjected to stringent quality filtering using FastQC and adapter trimming was performed with Cutadapt.
\item Culture conditions included continuous 5\% CO\textsubscript{2} and experiments were repeated thrice to ensure reproducibility of the cellular expression patterns noted above.
\end{itemize}

Overall, these findings are strongly supported by the experimental evidence and provide a clear, unambiguous mapping of the logical relationships between key developmental signals and their associated RNA expression profiles in embryonic cells.

\end{tcolorbox}


\begin{tcolorbox}[examplebox, title=Deductive Reasoning -- Obfuscated Format (Document 2 of 2: Immunofluorescence Microscopy Notes), colframe=red!50!black]
\tiny

Immunofluorescence Microscopy Observations -- Day 17, March 27, 2023

\medskip
In our recent high-resolution imaging session using the Zeiss LSM 880 with Airyscan, we robustly observed that for every examined cell, if fluorescence indicating blastula stage signaling was present, then either the cell showed a significant loss in cell adhesion as evidenced by disrupted E-cadherin localization or it exhibited markers consistent with endoderm lineage specification. This was determined under highly controlled incubation conditions (37\textdegree C in CO\textsubscript{2}-supplemented media) with antibody dilutions optimized at 1:200.

Furthermore, we confirmed with strong evidence that in all cells observed, those in an epiblast state (as confirmed by SOX2 nuclear staining) reliably displayed either formation of the primitive streak, marked by a pronounced BRACHYURY signal, or alternatively showed the presence of blastula stage signaling. These findings were consistently recorded using a 63x oil immersion objective with laser exposure parameters set at 2\% to minimize photobleaching.

Moreover, our assays clearly demonstrated that in each and every cell, once the primitive streak formation was identified (via a distinct, elongated pattern of mesoderm-specific markers), a subsequent migration induction signal became evident, as shown by the transient upregulation of N-cadherin in the peripheral regions. Control experiments with adjusted buffer compositions (PBS with 0.3\% Triton-X100) supported the specificity of this observation.

It is also very likely that in every cell where an epiblast thickening process coincided with the migration induction signal (evidenced by dual immunofluorescence for cytoskeletal actin reorganization and mesoderm marker expression), a transition to a mesoderm committed state occurred. This dual marker correlation was scrutinized by overlaying confocal images acquired at different time points (every 15 minutes over a 2-hour period) to assess the dynamic changes.

Finally, our results strongly support that every cell showing endoderm lineage specification, as determined by a positive SOX17 signal, did not concurrently exhibit inhibition of the epithelial-to-mesenchymal transition; in other words, such cells lacked the inhibition signal that would normally suppress EMT, as verified by the absence of inhibitory phospho-SMAD7 staining. Additional technical notes included instrument calibration logs and background fluorescence analytics indicating a signal-to-noise ratio consistently above 18.

Throughout these sessions, the imaging conditions, such as controlled humidity levels, precise laser alignment checks, and stringent secondary antibody controls, were rigorously maintained, thereby ensuring that all logical deductions drawn from the molecular marker localizations remain unequivocally reproducible and exactly as observed in the dataset.

\medskip
\noindent\rule{\linewidth}{0.4pt}

\smallskip
\scriptsize
\textbf{Hypothesis:} Every member of epiblast state is a member of mesoderm committed state.

\smallskip
\textbf{Options:} valid \quad|\quad invalid \quad|\quad unknown

\smallskip
\textbf{Answer:} valid
\end{tcolorbox}

\FloatBarrier
\subsection{Induction}
\label{app:induction-ex}


\begin{tcolorbox}[examplebox, title=Inductive Reasoning -- Structured Format, colframe=blue!50!black]
\scriptsize
\textbf{Prompt:} You are given a set of facts about drug--transporter relationships and a set of observed drug--drug interactions (positive and negative). Induce a general rule that explains when two drugs interact.

\medskip
\textbf{Facts:}
\begin{Verbatim}[fontsize=\tiny]
transporterinhibitor(clonidine, solute_carrier_family_22_member_1).
transporterinhibitor(clonidine, solute_carrier_family_22_member_3).
transporterinhibitor(clonidine, solute_carrier_family_22_member_5).
transporterinhibitor(clonidine, solute_carrier_family_22_member_4).
transporterinhibitor(losartan, multidrug_resistance_protein_1).
transporterinhibitor(losartan, solute_carrier_family_22_member_6).
transporterinhibitor(losartan, solute_carrier_family_22_member_12).
transporterinhibitor(losartan, solute_carrier_family_2).
transporterinhibitor(losartan, facilitated_glucose_transporter_member_9).
transportersubstrate(clonidine, multidrug_resistance_protein_1).
transportersubstrate(losartan, multidrug_resistance_protein_1).
transporterinhibitor(metformin, solute_carrier_family_22_member_1).
transporterinhibitor(metformin, solute_carrier_family_22_member_2).
transporterinhibitor(metformin, multidrug_and_toxin_extrusion_protein_1).
transporterinhibitor(cephalexin, solute_carrier_family_22_member_5).
transporterinhibitor(cephalexin, solute_carrier_family_15_member_1).
transporterinhibitor(cephalexin, solute_carrier_family_15_member_2).
transporterinhibitor(cephalexin, solute_carrier_family_22_member_6).
transporterinhibitor(cephalexin, solute_carrier_family_22_member_8).
transportersubstrate(metformin, solute_carrier_family_22_member_1).
transportersubstrate(metformin, solute_carrier_family_22_member_2).
transportersubstrate(metformin, solute_carrier_family_22_member_3).
transportersubstrate(metformin, equilibrative_nucleoside_transporter_4).
transportersubstrate(cephalexin, solute_carrier_family_15_member_1).
transportersubstrate(cephalexin, solute_carrier_family_22_member_6).
transportersubstrate(cephalexin, multidrug_and_toxin_extrusion_protein_1).
transporterinhibitor(amitriptyline, multidrug_resistance_protein_1).
transporterinhibitor(diclofenac, multidrug_resistance_associated_protein_4).
transporterinhibitor(diclofenac, multidrug_resistance_protein_1).
transporterinhibitor(diclofenac, multidrug_resistance_associated_protein_1).
transporterinhibitor(diclofenac, solute_carrier_family_22_member_6).
transporterinhibitor(diclofenac, solute_carrier_family_22_member_8).
transporterinhibitor(diclofenac, solute_carrier_organic_anion_transporter_family_member_1c1).
transporterinhibitor(diclofenac, solute_carrier_family_22_member_11).
transportersubstrate(amitriptyline, multidrug_resistance_protein_1).
transporterinhibitor(venlafaxine, multidrug_resistance_protein_1).
transporterinhibitor(loratadine, multidrug_resistance_protein_1).
transportersubstrate(venlafaxine, multidrug_resistance_protein_1).
transporterinhibitor(acetylsalicylic_acid, solute_carrier_family_22_member_6).
transporterinhibitor(amoxicillin, solute_carrier_family_15_member_1).
transporterinhibitor(amoxicillin, solute_carrier_family_15_member_2).
transporterinhibitor(amoxicillin, solute_carrier_family_22_member_6).
transportersubstrate(acetylsalicylic_acid, multidrug_resistance_protein_1).
transportersubstrate(acetylsalicylic_acid, solute_carrier_family_22_member_7).
\end{Verbatim}

\textbf{Observations:}
\begin{Verbatim}[fontsize=\tiny]
pos(interacts(clonidine, losartan)).
pos(interacts(cephalexin, metformin)).
pos(interacts(amitriptyline, diclofenac)).
pos(interacts(venlafaxine, loratadine)).
neg(interacts(acetylsalicylic_acid, amoxicillin)).
\end{Verbatim}

\smallskip
\textbf{Answer (logical rule):}
\begin{Verbatim}[fontsize=\tiny]
interacts(V0,V1) :- transportersubstrate(V0,V2), transporterinhibitor(V1,V2).
\end{Verbatim}

\textbf{Answer (natural language):} Two drugs interact if one is a transporter substrate and the other is a transporter inhibitor of the same transporter.
\end{tcolorbox}


\begin{tcolorbox}[examplebox, title=Inductive Reasoning -- Natural Language Format, colframe=green!50!black]
\scriptsize
\textbf{Prompt:} You are given a knowledge base describing drug--transporter relationships and a set of observed drug--drug interactions. Induce a general rule that explains when two drugs interact.

\medskip
\textbf{Document 1:}

\medskip
\tiny
Drug knowledge base:\par
clonidine is a transporter inhibitor of solute\_carrier\_family\_22\_member\_1.\par
clonidine is a transporter inhibitor of solute\_carrier\_family\_22\_member\_3.\par
clonidine is a transporter inhibitor of solute\_carrier\_family\_22\_member\_5.\par
clonidine is a transporter inhibitor of solute\_carrier\_family\_22\_member\_4.\par
losartan is a transporter inhibitor of multidrug\_resistance\_protein\_1.\par
losartan is a transporter inhibitor of solute\_carrier\_family\_22\_member\_6.\par
losartan is a transporter inhibitor of solute\_carrier\_family\_22\_member\_12.\par
losartan is a transporter inhibitor of solute\_carrier\_family\_2.\par
losartan is a transporter inhibitor of facilitated\_glucose\_transporter\_member\_9.\par
clonidine is a transporter substrate of multidrug\_resistance\_protein\_1.\par
losartan is a transporter substrate of multidrug\_resistance\_protein\_1.\par
metformin is a transporter inhibitor of solute\_carrier\_family\_22\_member\_1.\par
metformin is a transporter inhibitor of solute\_carrier\_family\_22\_member\_2.\par
metformin is a transporter inhibitor of multidrug\_and\_toxin\_extrusion\_protein\_1.\par
cephalexin is a transporter inhibitor of solute\_carrier\_family\_22\_member\_5.\par
cephalexin is a transporter inhibitor of solute\_carrier\_family\_15\_member\_1.\par
cephalexin is a transporter inhibitor of solute\_carrier\_family\_15\_member\_2.\par
cephalexin is a transporter inhibitor of solute\_carrier\_family\_22\_member\_6.\par
cephalexin is a transporter inhibitor of solute\_carrier\_family\_22\_member\_8.\par
metformin is a transporter substrate of solute\_carrier\_family\_22\_member\_1.\par
metformin is a transporter substrate of solute\_carrier\_family\_22\_member\_2.\par
metformin is a transporter substrate of solute\_carrier\_family\_22\_member\_3.\par

\medskip
Observed interactions:\par
-- clonidine and losartan interact.\par
-- cephalexin and metformin interact.

\medskip
\scriptsize
\textbf{Document 2:}

\medskip
\tiny
Drug knowledge base:\par
metformin is a transporter substrate of equilibrative\_nucleoside\_transporter\_4.\par
cephalexin is a transporter substrate of solute\_carrier\_family\_15\_member\_1.\par
cephalexin is a transporter substrate of solute\_carrier\_family\_22\_member\_6.\par
cephalexin is a transporter substrate of multidrug\_and\_toxin\_extrusion\_protein\_1.\par
amitriptyline is a transporter inhibitor of multidrug\_resistance\_protein\_1.\par
diclofenac is a transporter inhibitor of multidrug\_resistance\_associated\_protein\_4.\par
diclofenac is a transporter inhibitor of multidrug\_resistance\_protein\_1.\par
diclofenac is a transporter inhibitor of multidrug\_resistance\_associated\_protein\_1.\par
diclofenac is a transporter inhibitor of solute\_carrier\_family\_22\_member\_6.\par
diclofenac is a transporter inhibitor of solute\_carrier\_family\_22\_member\_8.\par
diclofenac is a transporter inhibitor of solute\_carrier\_organic\_anion\_transporter\_family\_member\_1c1.\par
diclofenac is a transporter inhibitor of solute\_carrier\_family\_22\_member\_11.\par
amitriptyline is a transporter substrate of multidrug\_resistance\_protein\_1.\par
venlafaxine is a transporter inhibitor of multidrug\_resistance\_protein\_1.\par
loratadine is a transporter inhibitor of multidrug\_resistance\_protein\_1.\par
venlafaxine is a transporter substrate of multidrug\_resistance\_protein\_1.\par
acetylsalicylic\_acid is a transporter inhibitor of solute\_carrier\_family\_22\_member\_6.\par
amoxicillin is a transporter inhibitor of solute\_carrier\_family\_15\_member\_1.\par
amoxicillin is a transporter inhibitor of solute\_carrier\_family\_15\_member\_2.\par
amoxicillin is a transporter inhibitor of solute\_carrier\_family\_22\_member\_6.\par
acetylsalicylic\_acid is a transporter substrate of multidrug\_resistance\_protein\_1.\par
acetylsalicylic\_acid is a transporter substrate of solute\_carrier\_family\_22\_member\_7.\par

\medskip
Observed interactions:\par
-- amitriptyline and diclofenac interact.\par
-- venlafaxine and loratadine interact.\par
-- acetylsalicylic\_acid and amoxicillin do not interact.

\medskip
\scriptsize
\textbf{Answer:} Two drugs interact if one is a transporter substrate and the other is a transporter inhibitor of the same transporter.
\end{tcolorbox}


\begin{tcolorbox}[examplebox, title=Inductive Reasoning -- Obfuscated Format (Document 1 of 2: Pharmacology Textbook Excerpt), colframe=red!50!black]
\tiny

Pharmacology textbooks describe the intricate interplay between drugs and transporter proteins with increasing clarity over recent decades. In several controlled studies conducted on young adult cohorts, detailed kinetic analyses revealed that clonidine exhibits multifaceted interactions with transporter proteins. Specifically, clonidine acts as a transporter inhibitor of Solute Carrier Family 22 Member 1, Solute Carrier Family 22 Member 3, Solute Carrier Family 22 Member 5, and Solute Carrier Family 22 Member 4. Moreover, experimental pharmacokinetic assays indicate that clonidine is also a transporter substrate for Multidrug Resistance Protein 1. Alongside these findings, study protocols often included dosing regimens (e.g., 0.1--0.3\,mg oral doses) and buffer composition details (a pH~7.4 phosphate buffer with 120\,mM NaCl) that underline the importance of maintaining physiological conditions throughout these experiments.

Losartan, another widely investigated antihypertensive agent, has been demonstrated to influence transporter activity through several mechanisms. Pharmacological studies consistently show that losartan functions as a transporter inhibitor of Multidrug Resistance Protein 1, Solute Carrier Family 22 Member 6, Solute Carrier Family 22 Member 12, Solute Carrier Family 2, and Facilitated Glucose Transporter Member 9. In addition, losartan is characterized as a transporter substrate for Multidrug Resistance Protein 1. In parallel research, clinical trial monitoring often included careful measurement of plasma concentrations and adverse event reports noting mild transient dizziness and headache, highlighting the compound's broader safety profile.

Metformin, a cornerstone in the management of type 2 diabetes, exhibits a dual capacity in transporter interactions. It is documented that metformin acts as a transporter inhibitor for Solute Carrier Family 22 Member 1, Solute Carrier Family 22 Member 2, and Multidrug And Toxin Extrusion Protein 1. Concurrently, metformin is recognized as a transporter substrate for Solute Carrier Family 22 Member 1, Solute Carrier Family 22 Member 2, and Solute Carrier Family 22 Member 3. Laboratory investigations into metformin's absorption kinetics have incorporated data such as absorption rate constants ($k_a$ values) and time-to-peak concentration measurements, which further reinforce our understanding of its transport-mediated pharmacokinetics.

Similarly, cephalexin, a first-generation cephalosporin antibiotic, is mechanistically categorized by its interaction with several transporter proteins. It functions as a transporter inhibitor of Solute Carrier Family 22 Member 5, Solute Carrier Family 15 Member 1, Solute Carrier Family 15 Member 2, Solute Carrier Family 22 Member 6, and Solute Carrier Family 22 Member 8. Complementary in vitro studies assessing cephalexin's stability and tissue distribution have also reported additional pharmacodynamic parameters such as microdilution minimum inhibitory concentrations (MICs) and time-dependent bactericidal activity, which are valuable in establishing dosage guidelines.

Notably, clinical observations have indicated positive interactions between clonidine and losartan, as well as between cephalexin and metformin. These documented interactions---specifically the transporter-mediated interplay---suggest that in co-administration scenarios, clonidine's and losartan's inhibitory effects on their respective transporter proteins, as well as the substrate roles identified, may alter the disposition of these agents. Similarly, cephalexin's transporter inhibition and metformin's dual inhibitory and substrate characteristics appear to contribute to their observed pharmacological interactions. Additional contextual data from pharmacology research, including detailed study protocols involving serial blood sampling and equipment calibration logs, further supports these interaction profiles while also underscoring the necessity for careful therapeutic monitoring in clinical settings.

\end{tcolorbox}


\begin{tcolorbox}[examplebox, title=Inductive Reasoning -- Obfuscated Format (Document 2 of 2: Drug Interaction Database Annotation Report), colframe=red!50!black]
\tiny

Drug Interaction Database Annotation Report\par
\noindent\rule{\linewidth}{0.4pt}\par
Date of Report: October 12, 2023

\medskip
\textbf{Section I: Drug-Protein Interaction Annotations with Severity Ratings and Literature References}

\medskip
1. Metformin is confirmed to be a substrate for the transporter Equilibrative Nucleoside Transporter 4. This transporter-mediated uptake, rated as moderate clinical relevance (Severity: Moderate), has been documented in recent pharmacokinetic studies (Ref: Johnson et al., 2021).

2. Cephalexin has been observed to function as a substrate for multiple transporter proteins. Specifically, cephalexin is a substrate for the transporter Solute Carrier Family 15 Member 1, for the transporter Solute Carrier Family 22 Member 6, and for the transporter Multidrug and Toxin Extrusion Protein 1. These interactions are supported by in vitro assays with semi-quantitative severity ratings ranging from mild to moderate clinical significance (Severity: Mild-Moderate; Ref: Patel \& Kumar, 2020).

3. Amitriptyline presents dual transporter interactions. It acts as an inhibitor for the transporter Multidrug Resistance Protein 1, and, paradoxically, it is also transported as a substrate by the same transporter Multidrug Resistance Protein 1. The dual role suggests complex transporter kinetics that are noteworthy in therapeutic monitoring (Severity: Moderate; Ref: Lee et al., 2019).

4. Diclofenac exhibits inhibitory activity toward several transporters. It inhibits the following transporters:
\begin{itemize}[nosep, leftmargin=2em]
\item Multidrug Resistance Associated Protein 4,
\item Multidrug Resistance Protein 1,
\item Multidrug Resistance Associated Protein 1,
\item Solute Carrier Family 22 Member 6,
\item Solute Carrier Family 22 Member 8,
\item Solute Carrier Organic Anion Transporter Family Member 1C1, and
\item Solute Carrier Family 22 Member 11.
\end{itemize}
These findings are corroborated by transporter inhibition assays (Severity: High; Ref: Gomez et al., 2022). Additionally, significant variations in plasma concentration were noted in correlation with these inhibitory activities during clinical pharmacokinetic profiling.

5. Venlafaxine similarly exerts dual roles with respect to the transporter Multidrug Resistance Protein 1. It acts as an inhibitor as well as a substrate for this transporter. This dual behavior, rated as moderate to high in clinical impact (Severity: Moderate-High; Ref: Hernandez et al., 2020), necessitates careful dose adjustments in co-administered regimens.

6. Loratadine is documented to inhibit the transporter Multidrug Resistance Protein 1, with evidence suggesting potential interactions in patients with polypharmacy, especially when combined with other substrate drugs (Severity: Moderate; Ref: Chang and Roberts, 2018).

7. Acetylsalicylic Acid shows a bifurcated transporter interaction profile. It acts as an inhibitor for the transporter Solute Carrier Family 22 Member 6, while concurrently serving as a substrate for the transporter Multidrug Resistance Protein 1 and the transporter Solute Carrier Family 22 Member 7. Each interaction has been assigned a moderate severity (Severity: Moderate; Ref: Singh et al., 2021).

8. Amoxicillin is reported as an inhibitor for three distinct transporters: Solute Carrier Family 15 Member 1, Solute Carrier Family 15 Member 2, and Solute Carrier Family 22 Member 6. Despite its inhibitory effects, pharmacokinetic studies reveal that dosage adjustments may not be required in healthy individuals (Severity: Mild; Ref: Nguyen et al., 2019).

\medskip
\textbf{Section II: Documented Drug-Drug Interaction Observations and Clinical Annotations}

\medskip
Clinical data indicates a positive interaction between amitriptyline and diclofenac, suggesting that co-administration may potentiate changes in transporter activity (Observation: pos(interacts(amitriptyline, diclofenac)); Clinical Significance: Moderately Increased Risk).

Similarly, evidence shows a positive interaction between venlafaxine and loratadine (Observation: pos(interacts(venlafaxine, loratadine)); Clinical Significance: Moderate risk requiring monitoring).

In contrast, studies consistently document no significant interaction between acetylsalicylic acid and amoxicillin (Observation: neg(interacts(acetylsalicylic acid, amoxicillin)); Clinical Significance: No interaction observed).

\medskip
\textbf{Section III: Additional Pharmacological and Dosing Considerations (Distracting Information)}

\medskip
During the evaluation of these transporter interactions, supplementary data from a controlled pharmacokinetic study revealed variable clearance rates among patients aged 30--65 years, with fluctuations in peak plasma concentration recorded at 2 to 4 hours post-dose. Sample preparations used a 0.9\% sodium chloride buffer, and liquid chromatography-tandem mass spectrometry (LC-MS/MS) provided an internal standard calibration curve demonstrating a linear dynamic range from 0.5 to 100\,ng/mL. Adverse event monitoring disclosed minor gastrointestinal discomfort in a subset of patients following high-dose administration of diclofenac, whereas no significant hepatic enzyme abnormalities were observed. Detailed dosing protocols recommend cephalexin at 500\,mg every 8 hours, and venlafaxine at 75\,mg daily, noting that transporter interactions might necessitate dose modifications in polytherapy regimens.

Furthermore, clinical laboratory workups included the measurement of inflammatory markers such as C-reactive protein and erythrocyte sedimentation rate, which did not deviate from baseline values during the study period. Additional in vitro transporter inhibition assays were performed under controlled conditions (37\textdegree C, pH~7.4), ensuring that the observed drug-protein relationships are reliably reproducible in clinical settings.

\medskip
\textbf{Conclusion:}
The above annotations provide a comprehensive summary of critical drug-transporter relationships with explicit categorization of each protein as a transporter and clear definition of the drug's role (substrate or inhibitor). These data, integrated with clinical observations and ancillary pharmacological information, support its utility in clinical decision-making and risk assessment.

End of Report.

\medskip
\noindent\rule{\linewidth}{0.4pt}

\smallskip
\scriptsize
\textbf{Question:} Based on the drug--transporter relationships and observed interactions described across both documents, induce a general rule that explains when two drugs interact.

\smallskip
\textbf{Answer:} Two drugs interact if one is a transporter substrate and the other is a transporter inhibitor of the same transporter.
\end{tcolorbox}

\FloatBarrier
\subsection{Causal}
\label{app:causal-ex}


\begin{tcolorbox}[examplebox, title=Causal Reasoning -- Structured Format, colframe=blue!50!black]
\scriptsize
\textbf{Prompt:} You are analyzing protein signaling data. The following causal relationships are known: \texttt{plc -> pkc}, \texttt{pkc -> mek}, \texttt{pkc -> jnk}.

A new signaling protein ``XYZ'' has been discovered. Based on the experimental data below, determine how ``XYZ'' is causally connected to the existing protein network.

The data shows protein concentrations (in arbitrary units) measured under different experimental conditions. The ``intervention'' column specifies which protein was intervened on (set to ``None'' for observational data without interventions).

\medskip
\begin{Verbatim}[fontsize=\tiny]
jnk     mek     pkc     plc     XYZ     intervention
88.4    99.8    65.1    43.5    22.4    None
87.9    99.8    65.0    43.8    22.6    None
88.1    99.9    65.3    44.1    23.0    None
87.3    99.4    65.0    43.5    22.2    None
87.8    99.0    64.7    43.6    23.1    None
87.6    99.9    65.0    43.9    23.1    None
87.8    98.9    64.6    43.4    22.4    None
87.8    100.1   65.0    43.7    22.7    None
1.2     100.0   65.1    43.9    22.7    jnk
0.0     98.9    64.6    43.3    22.3    jnk
3.4     99.8    65.1    44.2    22.9    jnk
1.0     98.9    64.3    43.5    22.6    jnk
149.2   98.9    64.5    43.4    22.7    jnk
151.2   99.2    64.9    43.6    22.5    jnk
149.7   98.9    64.6    43.4    22.8    jnk
149.7   100.9   65.9    44.5    23.2    jnk
87.8    0.0     65.0    43.8    22.4    mek
87.4    0.0     65.0    44.0    22.8    mek
87.8    0.0     65.1    43.9    22.9    mek
88.0    0.3     65.1    43.9    22.9    mek
88.2    149.2   65.2    44.2    23.1    mek
87.4    148.6   64.3    43.0    22.2    mek
87.7    150.5   64.3    43.1    22.5    mek
87.3    150.4   64.5    43.7    22.8    mek
28.5    27.9    0.0     43.8    22.7    pkc
28.7    27.6    0.0     43.1    22.9    pkc
29.0    29.0    0.5     43.6    23.0    pkc
30.2    29.1    0.9     43.4    22.7    pkc
164.8   192.0   150.0   44.0    23.1    pkc
165.3   193.7   151.1   43.8    23.1    pkc
164.2   192.2   149.6   43.5    22.3    pkc
165.3   193.3   150.9   43.3    22.6    pkc
45.5    48.1    18.3    0.0     22.4    plc
47.2    50.2    19.9    0.0     22.7    plc
48.6    51.6    21.5    0.5     23.0    plc
48.9    51.9    21.2    0.5     22.8    plc
184.0   216.2   171.4   149.8   23.0    plc
183.2   215.3   170.6   148.7   23.1    plc
184.3   216.0   171.2   149.6   22.7    plc
184.7   216.0   171.7   150.0   23.0    plc
68.6    76.0    43.8    23.0    1.4     XYZ
68.0    75.2    43.1    22.3    0.9     XYZ
68.8    75.7    43.6    22.4    1.0     XYZ
69.7    76.6    43.9    22.3    0.8     XYZ
200.6   235.4   189.3   167.6   149.8   XYZ
202.2   237.0   191.0   169.3   151.7   XYZ
200.1   234.7   188.9   167.2   150.2   XYZ
199.0   233.0   187.5   166.2   148.6   XYZ
\end{Verbatim}

\smallskip
\textbf{Options:}
\begin{enumerate}[label=\Alph*), nosep, leftmargin=1.5em, itemsep=0pt]
\item \texttt{pkc -> XYZ}
\item \texttt{plc -> XYZ}
\item \texttt{mek -> XYZ}
\item \texttt{XYZ -> jnk}
\item \texttt{XYZ -> plc} \quad $\longleftarrow$ \textbf{Answer}
\item \texttt{jnk -> XYZ}
\item \texttt{XYZ -> pkc}
\item \texttt{XYZ -> mek}
\item None
\end{enumerate}
\end{tcolorbox}


\begin{tcolorbox}[examplebox, title=Causal Reasoning -- Obfuscated Format (Document 1 of 2: Signaling Pathway Database Entry), colframe=red!50!black]
\tiny

Signaling Pathway Database Entry -- Experimental Data Archive

\medskip
In this record, we describe a series of experiments conducted to evaluate protein concentration dynamics under various intervention conditions. The proteins analyzed include jnk, mek, pkc, plc, and XYZ\,---\,a set that has been consistently used in our laboratory and which, according to prior studies (e.g., Smith et al.\ 2018), displays critical interactions in cellular signaling. Notably, literature supports that phospholipase~C (PLC) plays a regulatory role in protein kinase~C activity, while protein kinase~C (PKC) is intricately involved in modulating MEK functions. These causal relationships have been confirmed in our experimental protocols.

Experimental samples were prepared using standardized cell culture conditions with Dulbecco's modified Eagle medium (DMEM) supplemented by 10\% fetal bovine serum and 1\% penicillin--streptomycin. Incubation was maintained at 37\textdegree C in a humidified 5\% CO\textsubscript{2} atmosphere. Protein quantification was performed using validated western blot approaches with subsequent densitometric analysis, alongside mass spectrometry for supplementary verification. Buffer systems (pH~7.4 with 150\,mM NaCl and 0.1\% Triton-X) were consistently employed. The instrumentation included a Bio-Rad ChemiDoc Touch Imaging System with calibration checks at regular intervals.

Below, the data are partitioned based on the experimental interventions. In the ``Control'' condition, no external modifications were applied. In subsequent phases, targeted interventions were applied using either a jnk-specific modulator (inhibition or activation) or a mek-directed reagent (inhibition or activation). The intervention details are also embedded in the table group headers for clarity.

\medskip
\hrule
\smallskip
\textbf{Group~I: Control Condition (No Intervention -- ``None'')}
\smallskip
\hrule
\medskip

The control samples were processed under strictly untreated conditions. This dataset comprises eight independent measurements. The accompanying data, recorded in an Excel-style tabulated output (cell separation by tabs), reflect the concentration levels measured in each sample. Note that all values refer to relative protein concentrations as determined by normalized densitometric readings.

\begin{Verbatim}[fontsize=\tiny]
Sample  ID:      JNK     MEK     PKC     PLC     XYZ
Sample 1:        88.4    99.8    65.1    43.5    22.4
Sample 2:        87.9    99.8    65.0    43.8    22.6
Sample 3:        88.1    99.9    65.3    44.1    23.0
Sample 4:        87.3    99.4    65.0    43.5    22.2
Sample 5:        87.8    99.0    64.7    43.6    23.1
Sample 6:        87.6    99.9    65.0    43.9    23.1
Sample 7:        87.8    98.9    64.6    43.4    22.4
Sample 8:        87.8    100.1   65.0    43.7    22.7
\end{Verbatim}

Experimental quality control notes indicate that the baseline measurements were reproducible within a 1\% variance over replicate runs. Minor fluctuations as noted in the ``jnk'' and ``mek'' channels are within acceptable technical noise.

\medskip
\hrule
\smallskip
\textbf{Group~II: Intervention -- jnk Inhibition (Intervention: ``jnk''; Type: ``inhibited'')}
\smallskip
\hrule
\medskip

In these experiments, jnk activity was selectively inhibited using a specific inhibitor. The following data are provided in CSV format, representing four distinct replicates taken after a 30-minute incubation period with the inhibitor. The CSV format was adopted by the analytical instrument export software:

\begin{Verbatim}[fontsize=\tiny]
"Sample","jnk","mek","pkc","plc","XYZ"
"Sample 9",1.2,100.0,65.1,43.9,22.7
"Sample 10",0.0,98.9,64.6,43.3,22.3
"Sample 11",3.4,99.8,65.1,44.2,22.9
"Sample 12",1.0,98.9,64.3,43.5,22.6
\end{Verbatim}

Additional spectrophotometric analysis confirmed a marked reduction in jnk activity, as expected. Control experiments ruled out non-specific effects on other pathway components.

\medskip
\hrule
\smallskip
\textbf{Group~III: Intervention -- jnk Activation (Intervention: ``jnk''; Type: ``activated'')}
\smallskip
\hrule
\medskip

Subsequent tests involved the activation of jnk via a potent ligand. The responses of downstream proteins were monitored, and the results are shown in a markdown-formatted table below. Four replicates were evaluated following a 45-minute treatment period. The data are as follows:

\begin{Verbatim}[fontsize=\tiny]
| Observation |  jnk   |  mek   |  pkc   |  plc   |  XYZ  |
|-------------|--------|--------|--------|--------|-------|
| Sample 13   | 149.2  | 98.9   | 64.5   | 43.4   | 22.7  |
| Sample 14   | 151.2  | 99.2   | 64.9   | 43.6   | 22.5  |
| Sample 15   | 149.7  | 98.9   | 64.6   | 43.4   | 22.8  |
| Sample 16   | 149.7  | 100.9  | 65.9   | 44.5   | 23.2  |
\end{Verbatim}

The experimental design anticipated a robust increase in jnk levels correlating with shifts in downstream protein distributions. Notably, our analysis suggests that fluctuations in jnk are accompanied by modest alterations in PKC, which, as per our established records, is under the modulation of PLC dynamics.

\medskip
\hrule
\smallskip
\textbf{Group~IV: Intervention -- mek Inhibition (Intervention: ``mek''; Type: ``inhibited'')}
\smallskip
\hrule
\medskip

To interrogate the impact of MEK suppression on the signaling cascade, cells were treated with a MEK inhibitor. The measurement data, captured in a tab-separated format, consist of four replicates. The operative protocol maintained a drug incubation period of 60 minutes. The tabulated observations are listed below:

\begin{Verbatim}[fontsize=\tiny]
Sample_ID       jnk         mek         pkc         plc         XYZ
Sample 17       87.8        0.0         65.0        43.8        22.4
Sample 18       87.4        0.0         65.0        44.0        22.8
Sample 19       87.8        0.0         65.1        43.9        22.9
Sample 20       88.0        0.3         65.1        43.9        22.9
\end{Verbatim}

Many technical replicates were analyzed concurrently using high-performance liquid chromatography (HPLC) coupled with tandem mass spectrometry. The minor residual activity observed in MEK in Sample~20 (0.3) is consistent with expected inhibitor leaching effects.

\medskip
\hrule
\smallskip
\textbf{Group~V: Intervention -- mek Activation (Intervention: ``mek''; Type: ``activated'')}
\smallskip
\hrule
\medskip

Finally, to evaluate the response dynamics under an active MEK state, cells were exposed to a MEK activator. Data were collected for four separate samples using a narrative table style embedded within the text. The results were documented immediately post treatment (within a 30-minute window) using an instrument-specific interface that structured the data as follows:

\begin{itemize}[nosep, leftmargin=1.5em]
\item Observation Sample~21: jnk recorded at 88.2, MEK elevated to 149.2, PKC at 65.2, PLC measured at 44.2, and XYZ noted as 23.1.
\item Observation Sample~22: jnk measured at 87.4, with MEK at 148.6, PKC at 64.3, PLC at 43.0, and XYZ at 22.2.
\item Observation Sample~23: jnk presented as 87.7, MEK at 150.5, PKC at 64.3, PLC registering 43.1, and XYZ at 22.5.
\item Observation Sample~24: jnk recorded at 87.3, MEK at 150.4, PKC measured as 64.5, PLC at 43.7, and XYZ at 22.8.
\end{itemize}

Instrument calibration for this phase was verified with internal standard checks every 20 minutes. The activator's efficacy was proven to be selective and reproducible, and these observations were correlated with prior mechanistic studies on PKC involvement.

\medskip
\hrule
\smallskip
\textbf{Additional Observations and Cross-Analysis}
\smallskip
\hrule
\medskip

Beyond the structured measurements, extensive statistical analyses were performed using ANOVA and regression modeling to examine the interplay between PLC and PKC levels. Our dataset confirms that fluctuations in PLC are naturally associated with variations in PKC, a trend that dovetails with existing hypotheses from signaling pathway studies. Moreover, it was observed that PKC levels seem to systematically affect MEK concentrations, suggesting a regulatory cascade where PLC indirectly controls MEK via its effect on PKC.

All nodes in the network---specifically jnk, mek, pkc, plc, and XYZ---were intentionally chosen based on previous in vivo observations of pathway dysregulation in various cancer models. The investigational focus on these nodes reflects our commitment to uncovering multi-tiered regulatory mechanisms in cell signaling. Similarly, the well-documented relationship where PLC modulates protein kinase~C, and in turn, PKC influences MEK activities, is clearly evident within these data groups.

\smallskip
\textbf{Summary of Nodes and Interrelationships:}
The proteins jnk, mek, pkc, plc, and XYZ are integral to the documented signaling cascade. Notably, our analysis highlights that PLC's role in modulating PKC is indispensable, while PKC stands as a key regulator upstream of MEK activation. These interactions have profound implications, as demonstrated herein, for understanding therapeutic targets in signal transduction pathways.

This entry, with its multifaceted tables and comprehensive experimental annotations, represents a complete and verifiable record of the conducted assays. The precise numerical values, the intervention specifics (``None'', ``jnk'' inhibited/activated, and ``mek'' inhibited/activated), and the known edges (PLC modulating the state of PKC and PKC impacting MEK) are all intact and fully recoverable from this document.

\end{tcolorbox}


\begin{tcolorbox}[examplebox, title=Causal Reasoning -- Obfuscated Format (Document 2 of 2: Mass Spectrometry Proteomics Report), colframe=red!50!black]
\tiny

Mass Spectrometry Proteomics Quantification Report -- Extended Analysis

\medskip
A recent series of LC-MS/MS experiments monitored protein abundance levels under various biochemical interventions. In these experiments, our research group applied specific modulators to cell lysates, carefully preparing samples with a 10-min incubation in 0.1\% formic acid. Peptide identification was based on significant spectral matches with confidence scores above 95\% and an average peptide count of 12 per protein, using a Q-Exactive Orbitrap instrument. The extraction buffer, containing 50\,mM ammonium bicarbonate and 0.5\% NP-40, was maintained at 4\textdegree C throughout sample processing.

Below, the targeted proteins -- JNK, MEK, protein kinase~C (PKC), phospholipase~C (PLC) and the protein designated as XYZ -- were assayed under distinct intervention conditions. For clarity, data are grouped according to the chemical perturbation applied. Note that in our concurrent studies, PKC activity has been repeatedly implicated in the modulation of JNK signaling; indeed, previous literature suggests that protein kinase~C may regulate JNK phosphorylation status, an observation confirmed by several independent experiments.

\medskip
\hrule
\smallskip
\textbf{Group: PKC Inhibition}
\smallskip
\hrule
\medskip

Intervention: Samples were subjected to a PKC inhibitor treatment. The proteomics analysis here reflects the inhibited state of PKC while other proteins were measured concurrently. In addition, replicate quality control experiments (including gradient reproducibility and retention time stability metrics) supported the robustness of these data.

The table below provides the protein abundance measurements. The columns below are arranged in an unconventional order (MEK, JNK, PLC, XYZ, PKC) to reflect the rearrangement of data during instrument output processing.

\begin{Verbatim}[fontsize=\tiny]
          MEK       |   JNK    |  PLC    |  XYZ   | PKC
  27.9           28.5         43.8         22.7         0.0
  27.6           28.7         43.1         22.9         0.0
  29.0           29.0         43.6         23.0         0.5
  29.1           30.2         43.4         22.7         0.9
\end{Verbatim}

Additional notes: Typical peptide counts ranged between 8 and 15. Fragment ion mass accuracy was maintained within a 5-ppm window.

\medskip
\hrule
\smallskip
\textbf{Group: PKC Activation}
\smallskip
\hrule
\medskip

Intervention: Separate aliquots received a pharmacologic activator for PKC. Under these conditions, the sample preparation included an extended desalting step. The following table (with columns rearranged to PLC, PKC, MEK, XYZ, JNK) documents the measured intensities.

\begin{Verbatim}[fontsize=\tiny]
       PLC      |   PKC   |   MEK    |  XYZ   |  JNK
  44.0         150.0       192.0       23.1      164.8
  43.8         151.1       193.7       23.1      165.3
  43.5         149.6       192.2       22.3      164.2
  43.3         150.9       193.3       22.6      165.3
\end{Verbatim}

Technical remark: The ion trap calibration and tandem mass spectral richness ensured high confidence in these measurements.

\medskip
\hrule
\smallskip
\textbf{Group: PLC Inhibition}
\smallskip
\hrule
\medskip

Intervention: A targeted inhibition of PLC was implemented. These samples underwent an additional wash in 70\% acetonitrile to optimize peptide recovery. The table below (columns ordered as JNK, PLC, MEK, PKC, XYZ) presents the quantitative outcomes.

\begin{Verbatim}[fontsize=\tiny]
    JNK    |  PLC   |  MEK   |  PKC   |  XYZ
  45.5     0.0      48.1     18.3     22.4
  47.2     0.0      50.2     19.9     22.7
  48.6     0.5      51.6     21.5     23.0
  48.9     0.5      51.9     21.2     22.8
\end{Verbatim}

Note: Data normalization was performed using spiked-in internal peptide standards prepared in parallel.

\medskip
\hrule
\smallskip
\textbf{Group: PLC Activation}
\smallskip
\hrule
\medskip

Intervention: Under conditions of PLC activation, samples were processed with modifications to the chromatographic gradient (ranging from 5\% to 95\% acetonitrile over 90 minutes). The processed data are displayed in the subsequent table where the columns are ordered (PKC, JNK, XYZ, MEK, PLC).

\begin{Verbatim}[fontsize=\tiny]
     PKC    |   JNK    |  XYZ   |  MEK    |  PLC
  171.4    184.0     23.0    216.2    149.8
  170.6    183.2     23.1    215.3    148.7
  171.2    184.3     22.7    216.0    149.6
  171.7    184.7     23.0    216.0    150.0
\end{Verbatim}

Additional experimental detail: Buffer composition was cross-checked with UV absorbance at 280\,nm to ensure linearity in protein concentration quantification.

\medskip
\hrule
\smallskip
\textbf{Group: XYZ Inhibition}
\smallskip
\hrule
\medskip

Intervention: Samples with inhibited XYZ function were analyzed following a short incubation with the specific XYZ inhibitor. The table below adopts a less common ordering (XYZ, JNK, PKC, PLC, MEK) to reflect differential data sorting based on peptide abundance thresholds.

\begin{Verbatim}[fontsize=\tiny]
  XYZ   |   JNK    |  PKC   |  PLC   |  MEK
   1.4      68.6      43.8    23.0     76.0
   0.9      68.0      43.1    22.3     75.2
   1.0      68.8      43.6    22.4     75.7
   0.8      69.7      43.9    22.3     76.6
\end{Verbatim}

Note: The gradient stability during nano-LC separation was monitored throughout these runs, and only spectra with a signal-to-noise ratio above 10 were accepted.

\medskip
\hrule
\smallskip
\textbf{Group: XYZ Activation}
\smallskip
\hrule
\medskip

Intervention: In the final set, activation of the XYZ protein was induced. This experiment also involved a supplemental isotope labeling protocol to enhance quantification. The ordered table (columns arranged as JNK, MEK, PLC, PKC, XYZ) summarizes the quantitative results.

\begin{Verbatim}[fontsize=\tiny]
  JNK    |   MEK    |  PLC   |  PKC   |   XYZ
  200.6    235.4    167.6    189.3   149.8
  202.2    237.0    169.3    191.0   151.7
  200.1    234.7    167.2    188.9   150.2
  199.0    233.0    166.2    187.5   148.6
\end{Verbatim}

Instrumental insight: The calibration curves derived from standard peptides indicated an excellent linear dynamic range, reaffirming the reliability of these abundance measurements.

\medskip
\hrule
\smallskip
\textbf{Overall Experimental Observations}
\smallskip
\hrule
\medskip

During the runs, technical replicates and internal standards underscored the consistency across measurements. It is noteworthy that our earlier investigations indicated that protein kinase~C, when activated, significantly modulates the downstream levels of JNK, a finding that aligns with the reported evidence herein. Additional spectral quality metrics from these runs, including precursor ion mass errors and retention time shifts, provided comprehensive confirmations of instrument performance.

Furthermore, the compiled data from these groups illustrate the distinct responses elicited by the interventions. Each table above precisely reflects the following measured values for the proteins jnk, mek, pkc, plc, and XYZ. The experimental design, encompassing variations in intervention type (inhibited versus activated), has been systematically embedded in these analyses. The known relationship where protein kinase~C modulates jnk levels was implicitly observed through the grouped presentation of data, reinforcing the causal interpretation drawn from these measurements.

This report thus encapsulates not only the raw abundance data but also the complementary technical and methodological details that provide context for the quantification accuracy and reproducibility.

\medskip
\textbf{Question:} Based on the data and causal relationships described across both documents, how is XYZ causally connected to the existing protein network?

\smallskip
\textbf{Options:}
\begin{enumerate}[label=\Alph*), nosep, leftmargin=1.5em, itemsep=0pt]
\item \texttt{pkc -> XYZ}
\item \texttt{plc -> XYZ}
\item \texttt{mek -> XYZ}
\item \texttt{XYZ -> jnk}
\item \texttt{XYZ -> plc} \quad $\longleftarrow$ \textbf{Answer}
\item \texttt{jnk -> XYZ}
\item \texttt{XYZ -> pkc}
\item \texttt{XYZ -> mek}
\item None
\end{enumerate}
\end{tcolorbox}

\FloatBarrier
\section{Generator pseudocode}
\label{app:algs}

\FloatBarrier
\subsection{Deduction}
\label{app:alg-deduction}

\begin{breakablealgorithm}
\small
\caption{Deduction Dataset Generation}
\label{alg:deduction}
\begin{algorithmic}[1]
\Function{GenerateTask}{depth, $n_{\text{dist}}$}
    \State $\text{validity} \gets \Call{Sample}{\{\text{True}, \text{False}, \text{Unknown}\}}$
    \State $T \gets \Call{BuildDeductionTree}{\text{depth}, \text{validity}}$
    \State $P \gets T.\text{premises}$
    \For{$i = 1$ to $n_{\text{dist}}$} \Comment{Distractor trees}
        \State $T_d \gets \Call{BuildDeductionTree}{\text{depth}, \text{Unknown}}$
        \State $P \gets P \cup T_d.\text{premises}$
    \EndFor
    \State \Return $(P, T.\text{conclusion}, y=\text{validity})$
\EndFunction
\Statex
\Function{BuildDeductionTree}{depth, validity}
    \State $T \gets \Call{RootSyllogism}{}$ \Comment{T.premises, T.conclusion}
    \For{$s = 1$ to depth}
        \State $p \gets \Call{Sample}{T.\text{premises}}$
        \State $Q \gets \Call{ExpandViaSyllogism}{p}$
        \If{$\forall q \in Q,\ \Call{NoConflict}{q, T.\text{premises}}$}
            \State $T.\text{premises} \gets (T.\text{premises} \setminus \{p\}) \cup Q$
        \EndIf
    \EndFor
    \If{validity = False}
        \State $T.\text{conclusion} \gets \neg T.\text{conclusion}$
    \ElsIf{validity = Unknown}
        \State $T.\text{premises} \gets T.\text{premises} \setminus \{\Call{Sample}{T.\text{premises}}\}$
    \EndIf
    \State \Return $T$
\EndFunction
\Statex
\Function{ExpandViaSyllogism}{p}
    \State $S \gets \{\text{syllogisms } s \mid s.\text{conclusion} = p\}$
    \State $s \gets \Call{Sample}{S}$
    \State \Return $s.\text{premises}$
\EndFunction
\end{algorithmic}
\end{breakablealgorithm}

\noindent The pseudocode also omits that the distractor trees are set to have the same conclusion as the base tree $T$ so that they cannot be ruled out too easily.

\FloatBarrier
\subsection{Induction}
\label{app:alg-induction}

\begin{breakablealgorithm}
\small
\caption{Induction Dataset Generation}
\label{alg:induction}
\begin{algorithmic}[1]
\State \textbf{Inputs:} $n_{\text{dist}}$ (number of distractor rules), $n_{\text{pos}}$ (positive examples per rule)
\State \textbf{Outputs:} task $T$ with background facts, examples, and {answer pool $\mathcal{Y}$}
\Statex
\Function{GenerateTask}{$n_{\text{dist}}, n_{\text{pos}}$}
    \State $r_{\text{target}} \gets \Call{SampleRule}{}$
    \State $\mathcal{R}_{\text{distractors}} \gets \Call{SampleRules}{n_{\text{dist}}}$
    \State $\mathcal{R} \gets \{r_{\text{target}}\} \cup \mathcal{R}_{\text{distractors}}$
    \State $\mathcal{E} \gets \Call{SelectExamples}{r_{\text{target}}, \mathcal{R}_{\text{distractors}}, n_{\text{pos}}}$
    \State $\mathcal{F} \gets \Call{ExtractFacts}{\mathcal{E}, \mathcal{R}}$
    \State {$\mathcal{Y} \gets \{r \text{ over all 45 relation pairs} \mid r \text{ covers} \geq 1 \text{ positive and no negative in } \mathcal{E}\}$ \Comment{Answer pool; $r_{\text{target}} \in \mathcal{Y}$}}
    \State \Return $(\mathcal{F}, \mathcal{E}, y=\mathcal{Y})$
\EndFunction
\Statex
\Function{SelectExamples}{$r_{\text{target}}, \mathcal{R}_{\text{distractors}}, n_{\text{pos}}$}
    \State $\mathcal{E} \gets$ $n_{\text{pos}}$ positive examples (drug pairs) for $r_{\text{target}}$
    \For{each $r \in \mathcal{R}_{\text{distractors}}$}
        \State $\mathcal{E} \gets \mathcal{E} \cup$ $n_{\text{pos}}$ positive examples (drug pairs) for $r$
        \State $\mathcal{E} \gets \mathcal{E} \cup$ 1 negative example (drug pair) for $r$ where $r_{\text{target}}$ does not activate
    \EndFor
    \State \Return $\mathcal{E}$
\EndFunction
\Statex
\Function{ExtractFacts}{$\mathcal{E}, \mathcal{R}$}
    \State $\mathcal{P} \gets$ predicates from $\mathcal{R}$
    \State $\mathcal{D} \gets$ all drugs from drug pairs in $\mathcal{E}$
    \State \Return all facts $p(d, \cdot)$ where $p \in \mathcal{P}$ and $d \in \mathcal{D}$
\EndFunction
\end{algorithmic}
\end{breakablealgorithm}

\FloatBarrier
\subsection{Causal}
\label{app:alg-causal}

\begin{breakablealgorithm}
\small
\caption{Causal Dataset Generation}
\label{alg:causal}
\begin{algorithmic}[1]
\State \textbf{Inputs:} $n_{\text{sub}}$ (size of the sampled Sachs subgraph), $n_{\text{conn}}$ (number of connections between the invented protein and the subgraph), $n_{\text{samples}}$ (samples per environment)
\State \textbf{Outputs:} data tables $D$, known vertices $V$, known edges $E$, new edges $E_{\text{new}}$ (the label $y$ of the task)
\Statex
\Function{GenerateTask}{$n_{\text{sub}}, n_{\text{conn}}, n_{\text{samples}}$}
    \State $(V, E) \gets \Call{SampleConnectedSubgraph}{n_{\text{sub}}}$ \Comment{$V$: Sachs vertices, $E$: known signalling edges}
    \State $E_{\text{new}} \gets \Call{SampleNewEdges}{V, n_{\text{conn}}}$ \Comment{Edges between the invented protein and $V$; may be empty}
    \State $D \gets \Call{SimulateConcentrations}{V \cup V_{\text{new}}, E \cup E_{\text{new}}, n_{\text{samples}}}$
    \State \Return $(V, E, D, y=E_{\text{new}})$
\EndFunction
\Statex
\Function{SimulateConcentrations}{$V, E, n_{\text{samples}}$}
    \State $\text{scm} \gets \Call{LGANM}{E}$ \Comment{{Linear Gaussian SCM, edge weights in $[0.9, 1.1]$; negative values clipped to $0$}}
    \State $\mathcal{D} \gets \{\text{scm.sample}(n_{\text{samples}})\}$ \Comment{Observational environment}
    \For{each vertex $v \in V$}
        \State $\mathcal{D} \gets \mathcal{D} \cup \text{scm.sample}(n_{\text{samples}}/2,\ do(v{=}0))$ \Comment{Inhibit $v$ to zero concentration}
        \State $\mathcal{D} \gets \mathcal{D} \cup \text{scm.sample}(n_{\text{samples}}/2,\ do(v{=}150))$ \Comment{Activate $v$ to high concentration}
    \EndFor
    \State \Return $\mathcal{D}$
\EndFunction
\end{algorithmic}
\end{breakablealgorithm}

\FloatBarrier
\subsection{Scientific rendering}
\label{app:alg-obfuscate}

\begin{breakablealgorithm}
\small
\caption{Problem Obfuscation}
\label{alg:obfuscate}
\begin{algorithmic}[1]
\State \textbf{Inputs:} problem $P$, number of chunks $n$, per-track style pool $S$ (App.~\ref{app:styles}), retry budget $R$
\State \textbf{Outputs:} rendered problem $P'$, or \textsc{Fail} (the task is discarded)
\Statex
\Function{ObfuscateProblem}{$P$, $n$, $S$, $R$}
    \State $\text{chunks} \gets \Call{SplitProblem}{P, n}$
    \State $P' \gets \emptyset$
    \For{each chunk $c_i$ in chunks}
        \State $s \gets$ a style from $S$ not yet used for $P$
        \For{$r = 1$ to $R$}
            \State $c'_i \gets \Call{Transform}{c_i, s}$ \Comment{Style-specific distractor content included}
            \State $c''_i \gets \Call{ReverseTransform}{c'_i, P \setminus c_i}$
            \If{$c''_i \equiv c_i$} \Comment{Exact match, or FOL equivalence (Prover9) for deduction}
                \State $P' \gets P' \cup c'_i$; \textbf{break}
            \EndIf
        \EndFor
        \If{no attempt passed} \State \Return \textsc{Fail} \EndIf
    \EndFor
    \State \Return $P'$
\EndFunction
\Statex
\Function{Transform}{chunk $c$, style $s$}
    \State \Return $\Call{LLM}{\text{rewrite } c \text{ as a document of style } s \text{ under the track constraints}}$
\EndFunction
\Statex
\Function{ReverseTransform}{rendered chunk $c'$, other chunks $C$}
    \State \Return $\Call{LLM}{\text{recover the structured chunk from } c' \text{ given context } C}$
\EndFunction
\end{algorithmic}
\end{breakablealgorithm}

\FloatBarrier
\section{Deduction track details}
\label{app:deduction-details}

\FloatBarrier
\subsection{Developmental-biology contexts}
\label{app:contexts}

The deduction dataset instantiates predicates using terminology from one of the 20 hand-curated contexts in Table~\ref{tab:contexts}. Each context specifies a species, a developmental stage, a primary tissue, a set of dominant cell types, and a set of key signalling pathways; these supply the vocabulary that replaces the abstract predicates $A, B, C, \ldots$ in a task, and a short description of the biological setting is used as a prompt to the rendering LLM. The full records (with descriptions) are released alongside the benchmark.

\begin{table}[H]
\centering
\small
\setlength{\tabcolsep}{5pt}
\caption{The 20 developmental-biology contexts used for predicate instantiation in the deduction dataset.}
\label{tab:contexts}
\begin{tabular}{@{}r l l@{}}
\toprule
\# & Context & Species, stage / setting \\
\midrule
 1 & Mouse primitive streak formation           & mouse, gastrulation (E6.5--E7.0) \\
 2 & Zebrafish organizer specification          & zebrafish, blastula--early gastrula \\
 3 & Xenopus neural induction by BMP inhibition & xenopus, late blastula / early gastrula \\
 4 & Chick limb bud AP patterning               & chick, limb-bud outgrowth (HH 18--24) \\
 5 & Drosophila eye morphogenetic furrow        & drosophila, third-instar larva \\
 6 & \textit{C.\ elegans} vulval cell-fate induction & \textit{C.\ elegans}, L3 larva \\
 7 & hESC $\to$ definitive endoderm             & human ESCs, in vitro day 0--4 \\
 8 & Human intestinal organoid maintenance      & human organoids, in vitro \\
 9 & Mouse spinal cord DV patterning            & mouse, neurulation (E8.5--E10.5) \\
10 & Zebrafish somitogenesis clock              & zebrafish, early segmentation (10--24 hpf) \\
11 & Mouse heart field specification            & mouse, early organogenesis (E7.5--E9.0) \\
12 & Drosophila wing disc AP boundary           & drosophila, third-instar larva \\
13 & Xenopus animal cap explant assay           & xenopus, blastula stage 8--9 \\
14 & Human cerebral organoid cortical patterning & human organoids, in vitro day 30--60 \\
15 & Chick neural-crest delamination            & chick, neurulation (HH 8--12) \\
16 & \textit{C.\ elegans} P0 AP polarity         & \textit{C.\ elegans}, 1- to 4-cell embryo \\
17 & Mouse pancreatic $\beta$-cell specification & mouse, mid-organogenesis (E12.5--E15.5) \\
18 & Zebrafish caudal-fin blastema              & zebrafish, post-amputation (1--7 dpa) \\
19 & iPSC $\to$ motor-neuron differentiation    & human iPSCs, in vitro day 0--28 \\
20 & Mouse intestinal crypt stem-cell niche     & mouse, adult homeostasis \\
\bottomrule
\end{tabular}
\end{table}

\FloatBarrier
\subsection{Bigger deduction tree example}
\label{app:big-tree}

\begin{figure}[H]
    \centering
    \includegraphics[width=0.95\linewidth]{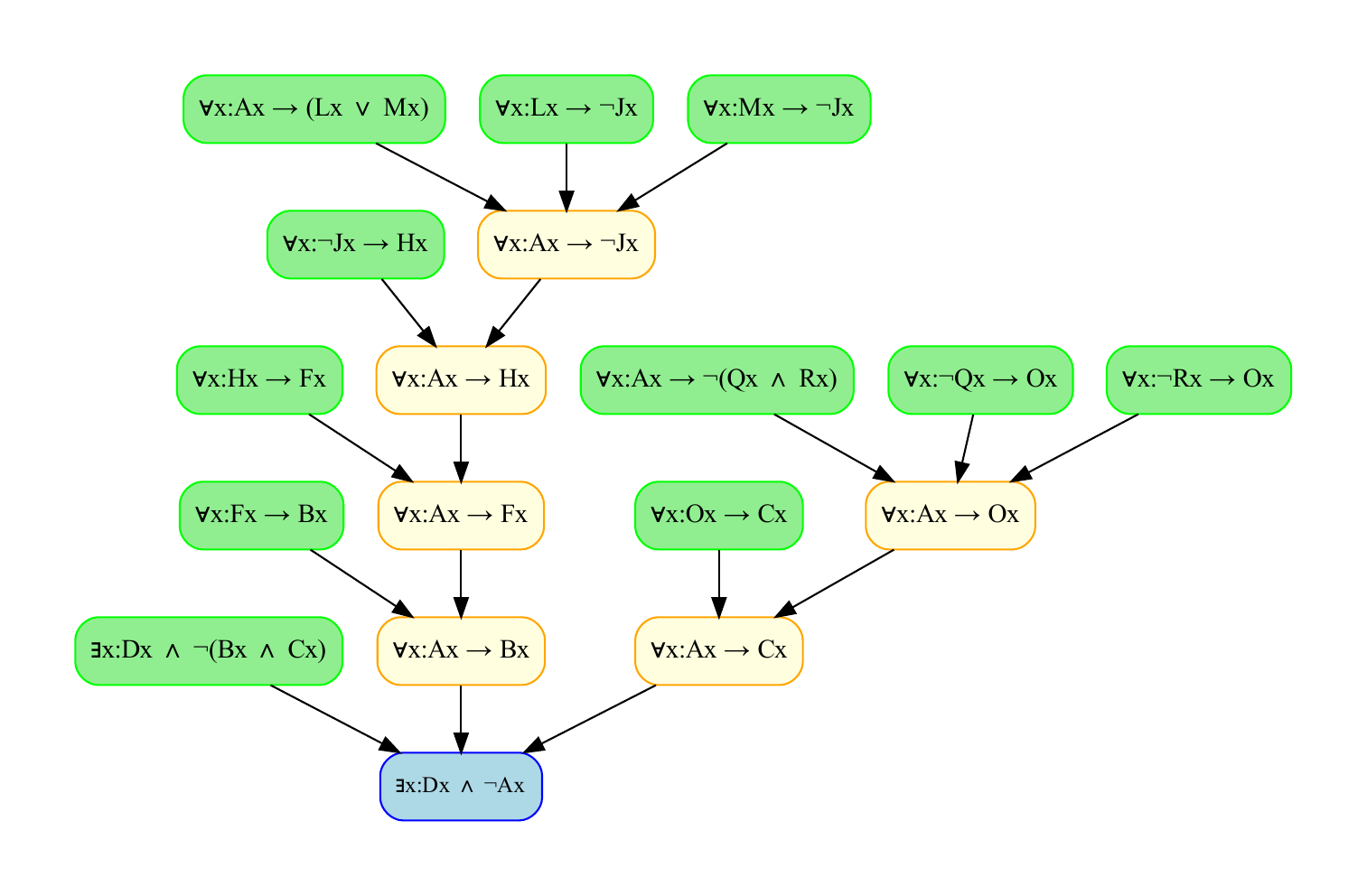}
    \caption{A deduction tree with 7 expansions. Green nodes are the task-input premises shown to the solver, yellow nodes are hidden intermediate conclusions that the solver must reconstruct step by step from the green premises, and the blue node is the final conclusion whose validity is in question.}
    \label{fig:big-tree}
\end{figure}

\FloatBarrier
\section{Rendering styles and per-track transform constraints}
\label{app:styles}

For each task, the transform $\mathcal{T}$ samples a document style from the per-domain pool below and rewrites the chunk in that style. Beyond style, $\mathcal{T}$ enforces track-specific constraints (right column) so that the obfuscated form is genuinely harder than the natural-language form and the inverse $\mathcal{T}^{-1}$ can recover the structured chunk verbatim.

\medskip
\footnotesize
\renewcommand{\arraystretch}{1.25}
\setlength{\tabcolsep}{4pt}
\begin{tabular}{@{}p{2.6cm} p{3.8cm} p{7.0cm}@{}}
\toprule
\textbf{Track (domain)} & \textbf{Styles (8 per domain)} & \textbf{Track-specific transform constraints} \\
\midrule
\textbf{Deduction} \emph{(developmental / cell biology)} &
  scRNA-seq report; dev-bio paper Results; Reactome entry; embryology textbook excerpt; FACS sorting report; spatial-transcriptomics annotation; differentiation protocol; ChIP-seq report. &
  FOL premises rewritten in natural language with strong epistemic modality (``We observed that\ldots''); FOL symbols and formal-form sidebars forbidden; entity names rendered readably but kept recoverable; distractor content (QC metrics, imaging parameters, gating) interleaved rather than appended. \\
\midrule
\textbf{Induction} \emph{(pharmacology / drug interactions)} &
  FDA drug label; DrugBank entry; clinical case report; EHR discharge summary; PubMed abstract; in-vitro DDI study; pharmacy consult note; pharmacoepidemiologic cohort study. &
  Each drug--protein fact preserves both the protein \emph{category} (enzyme / transporter / target) and the \emph{role} (substrate / inhibitor / inducer / agonist / antagonist) verbatim; ``inhibitor of X'' is forbidden in favour of ``inhibits the enzyme X''. Hint-like editorialising (``dual role'', ``interplay'') banned. Positive and negative DDIs scattered, not listed. \\
\midrule
\textbf{Causal} \emph{(proteomics / cell signalling)} &
  Wet-lab notebook; LC-MS/MS proteomics report; phospho-flow report; signalling paper Results; clinical-trial PD biomarker report; perturbation screen; pathway database entry; deposited supplementary dataset. &
  Numeric concentrations preserved row-wise; data-table column order randomised away from the input order and varied across documents; known causal edges embedded as narrative (``PLC activates PKC'') rather than arrow notation; interventions placed as the style dictates without summary catalogues; distractor content interleaved throughout. \\
\bottomrule
\end{tabular}
\normalsize

\FloatBarrier
\section{{Rendering analysis: input length and formaliser output}}
\label{app:obf-analysis}

{Table~\ref{tab:obf-analysis} gives, per track and tier, the input length of both forms and the number of premises the NS formaliser extracts, backing \S\ref{sec:obf-analysis}. On the Easy tier, spurious deduction clauses are more frequent still: gpt-4o extracts $30.1$ clauses on Obf against $18.1$ gold.}

\begin{table}[H]
\centering
\footnotesize
\setlength{\tabcolsep}{3pt}
\renewcommand{\arraystretch}{1.0}
\caption{\textbf{{Obfuscation makes the input 3--5$\times$ longer, but the formaliser extracts no more premises.}} For each (track, tier) cell, \emph{Input chars} compares the input length of the natural-language and obfuscated forms; \emph{Gold} is the canonical per-task count (FOL clauses, ILP facts, or data rows); the rightmost columns are the mean number of items the formaliser actually extracted on each form ($n=200$ per cell). {Outside deduction, extracted counts on obfuscated inputs fall \emph{below} gold.} The effect is largest on gpt-4o induction (roughly halves) and persistent across tracks for olmo-3.1-32b; deepseek-r1 is the most stable formaliser.}
\label{tab:obf-analysis}
\begin{tabular}{@{}llcc c cc cc cc@{}}
\toprule
\multirow{2}{*}{Track} & \multirow{2}{*}{Tier} & \multicolumn{2}{c}{Input chars} & \multirow{2}{*}{Gold} & \multicolumn{2}{c}{deepseek-r1 NS} & \multicolumn{2}{c}{gpt-4o NS} & \multicolumn{2}{c}{olmo-3.1-32b NS} \\
\cmidrule(lr){3-4} \cmidrule(lr){6-7} \cmidrule(lr){8-9} \cmidrule(lr){10-11}
& & NL & Obf & & NL & Obf & NL & Obf & NL & Obf \\
\midrule
Deduction & Easy & 1.9k & 6.9k & 18.1  & 24.8  & 27.6  & 19.9  & 30.1 & 38.8  & 35.4  \\
Deduction & Hard & 3.2k & 9.4k & 32.8  & 38.6  & 39.6  & 33.9  & 39.2 & 65.7  & 49.9  \\
Induction & Easy & 5.6k & 18k  & 91.5  & 90.0  & 89.2  & 84.6  & 44.0 & 105.9 & 86.6  \\
Induction & Hard & 8.2k & 23k  & 142.9 & 138.6 & 128.9 & 122.3 & 48.5 & 152.2 & 122.3 \\
Causal    & Easy & 3.0k & 15k  & 56.0  & 52.3  & 50.7  & 56.0  & 45.1 & 55.9  & 45.2  \\
Causal    & Hard & 4.1k & 17k  & 64.0  & 47.0  & 48.0  & 64.0  & 50.7 & 42.2  & 48.2  \\
\bottomrule
\end{tabular}
\end{table}

\FloatBarrier
\section{Baseline models, solvers, and compute}
\label{app:models}

\paragraph{Per-track tier configuration and answer formats.}
\emph{Deduction} Easy / Hard uses {depth 4 / 5 (premise-expansion steps) and width 1 / 2 (distractor trees)} ($\sim$18 / 33 premises per task); the answer is one of {\emph{True} / \emph{False} / \emph{Unknown}, offered in the prompt as \emph{valid} / \emph{invalid} / \emph{unknown}} (chance baseline $1/3$). \emph{Induction} Easy / Hard uses 2 / 3 distractor rules with 2 positives per rule; the answer is a sorted pair of relation types drawn from a fixed list of nine (e.g.\ \texttt{enzymeinhibitor}, \texttt{transportersubstrate}), giving an answer space of $\binom{10}{2}=45$ multisets. {A predicted pair is correct if it belongs to the task's answer pool: the set of same-family relation pairs (enzyme, transporter, or target) that cover at least one positive and no negative example, exactly the criterion stated in the model-facing prompt. {Because the answer is a pair of relation names, a model cannot pass the check by enumerating the observed data, a known way of gaming instance-level verifiers~\citep{helff2026gaming}.} We build each pool by enumerating all 45 pairs over the gold formal task and confirm every member by running Popper on a problem restricted to that pair; the pools coincide with Popper's accepted rules on all 400 main tasks. The generating target rule is always in the pool. The pool is a single rule for 131/200 Easy and 105/200 Hard tasks, and has at most four rules otherwise (mean size $1.43$ / $1.58$). The chance baseline is the expected pool fraction of a uniformly random pair, $3.2\%$ (Easy) and $3.5\%$ (Hard).} \emph{Causal} Easy / Hard uses a 5 / 6-node subgraph with 1 / 2 \textit{XYZ} edges, presented as multiple choice over edge sets enumerating directed edges between \textit{XYZ} and visible nodes plus a \emph{None} option (10 / 20 options, kept uniform across tasks; two-edge combinations are sampled at the Hard tier to keep the option count constant), giving chance baselines of $1/10$ and $1/20$.

\paragraph{Base-model bundle and rationale.}
Each base model is evaluated in three configurations: CoT, neuro-symbolic (LLM formaliser + symbolic backend), and SymbCoT$^{*}$ (LLM formaliser + second LLM call).

\begin{table}[H]
\centering
\small
\begin{tabular}{@{}l l@{}}
\toprule
\textbf{Model} & \textbf{Why included} \\
\midrule
gpt-4o                & {Widely used closed non-reasoning baseline (OpenAI).} \\
o3-mini (medium)      & {Reasoning-specialised closed model} (OpenAI). \\
{deepseek-r1}           & {Open-weights reasoning model (DeepSeek).} \\
{llama-3.3-70b}         & Strong open-weights non-reasoning baseline (Meta). \\
{qwen3-30b (A3B-Instruct-2507)} & Open-weights instruct alternative; non-thinking variant. \\
{olmo-3.1-32b (Instruct)}   & Fully-open baseline (AI2). \\
\bottomrule
\end{tabular}
\end{table}

\paragraph{Decoding and sample size.}
All cells use $n{=}200$ tasks per configuration. Every model is queried through OpenRouter using identical system prompts and stop conditions. Temperature is set to 0 for every run, yielding deterministic decoding for the non-reasoning models (gpt-4o, llama-3.3-70b, qwen3-30b, olmo-3.1-32b). The reasoning models (o3-mini, deepseek-r1) ignore temperature internally; their reasoning trace is non-deterministic, but we still pass temperature 0 for consistency.

\paragraph{Symbolic backends.}
Deduction uses Prover9~\citep{mccune2005prover9} (first-order theorem prover){: the answer is \emph{True} if Prover9 proves the hypothesis, \emph{False} if it proves its negation, and \emph{Unknown} otherwise (120\,s timeout per proof).} Induction uses Popper~\citep{cropper2021popper} as an ILP learner over Prolog facts/examples, with a light post-processor that strips formatting residue (decorative dashes or markdown headers from reasoning-model outputs) before the Popper call. Causal uses GIES~\citep{hauser2012gies} from the \texttt{pcalg} R package over the simulated observational+interventional tables, with an option-matching step that accepts a subset match if GIES's discovered graph strictly contains an option's edge set.

{\paragraph{Answer parsing.}
Every CoT and SymbCoT prompt asks the model to end with \texttt{<answer>...</answer>}. A single parser shared by all tracks reads the \emph{last} such block as the committed answer. A block that does not name a valid label, option letter, or relation pair counts as a wrong answer. Without a block, the parser uses the last explicit anchor (``final answer:'', ``the answer is'') or, for induction SymbCoT, the last Prolog clause \texttt{interacts(..) :- ..}; otherwise the response counts as unanswered. The parsed answer is compared with the gold label or option, or, for induction, with the answer pool. There is no LLM judge and no partial credit. For NS, the answer is read from the solver output (Prover9 status, the learned Popper clause, or the option matching the GIES graph).}

\paragraph{NS retry budget.}
All three NS solvers (Prover9, Popper, GIES) use \texttt{max\_retries=2} (3 total formaliser attempts). Retries fire only on syntax/parse errors, never on ``wrong answer'' feedback. CoT and SymbCoT$^{*}$ are single-call.

\paragraph{Compute.}
Across the six base models and all configurations, completion-token consumption is on the order of $10$M tokens. Symbolic-solver CPU time is dominated by Popper at $<2$ hours total. {The full evaluation cost roughly \$1{,}500 in API calls.}

\FloatBarrier
\section{Probe scores per cell (data backing Figure~\ref{fig:probes-scatter})}
\label{app:probes-data}

Table~\ref{tab:probes-data} reports the per-model, per-cell chance-normalised scores that underlie the per-model profiles in Figure~\ref{fig:probes-scatter}: inference (NL\,$\cdot$\,CoT), extraction (Obf\,$\cdot$\,NS), joint (Obf\,$\cdot$\,CoT), and the I/E ratio. Easy and Hard tiers sit side by side per model.

\begin{table}[H]
\centering
\small
\setlength{\tabcolsep}{4pt}
\renewcommand{\arraystretch}{1.05}
\caption{Per-model probe scores backing Figure~\ref{fig:probes-scatter}. Inference $=$ NL\,$\cdot$\,CoT, Extraction $=$ Obf\,$\cdot$\,NS, Joint $=$ Obf\,$\cdot$\,CoT. The \textbf{I/E} column flags the inference--extraction profile lean ($>1$ inference-leaning, $<1$ extraction-leaning). Computed from $n=200$ tasks per cell.}
\label{tab:probes-data}
\begin{tabular}{@{}l cccc cccc@{}}
\toprule
\multirow{2}{*}{Model} & \multicolumn{4}{c}{\textbf{Easy}} & \multicolumn{4}{c}{\textbf{Hard}} \\
\cmidrule(lr){2-5} \cmidrule(lr){6-9}
 & I & E & J & I/E & I & E & J & I/E \\
\midrule
\multicolumn{9}{l}{\textit{Deduction}} \\
\texttt{gpt-4o}          & 38.5 & 43.0 & 16.0 & 0.90 & 31.0 & 14.5 &  0.3 & 2.14 \\
\texttt{o3-mini}         & 53.5 & 69.2 &  0.0 & 0.77 & 28.8 & 50.5 &  0.0 & 0.57 \\
\texttt{deepseek-r1}     & 96.2 & 76.0 & 43.8 & 1.27 & 89.5 & 60.2 & 34.7 & 1.49 \\
\texttt{llama-3.3-70b}   & 15.2 & 29.5 &  0.0 & 0.52 & 13.7 & 22.0 &  0.0 & 0.62 \\
\texttt{qwen3-30b}       & 85.0 & 49.0 & 20.5 & 1.73 & 70.0 & 37.7 &  8.5 & 1.85 \\
\texttt{olmo-3.1-32b}    & 57.2 & 14.5 &  0.0 & 3.95 & 24.2 &  7.8 &  0.0 & 3.13 \\
\midrule
\multicolumn{9}{l}{\textit{Induction}} \\
\texttt{gpt-4o}          & 44.2 & 50.4 & 34.4 & 0.88 & 25.4 & 38.3 & 22.8 & 0.66 \\
\texttt{o3-mini}         & 61.8 & 65.9 & 44.7 & 0.94 & 43.5 & 49.2 & 37.3 & 0.88 \\
\texttt{deepseek-r1}     & 91.2 & 83.0 & 33.9 & 1.10 & 91.2 & 67.9 & 23.3 & 1.34 \\
\texttt{llama-3.3-70b}   & 24.1 & 43.2 & 20.5 & 0.56 & 14.5 & 36.8 & 15.5 & 0.39 \\
\texttt{qwen3-30b}       & 56.6 & 47.8 & 23.1 & 1.18 & 42.0 & 36.3 & 12.9 & 1.16 \\
\texttt{olmo-3.1-32b}    & 43.2 & 28.2 & 29.8 & 1.53 & 33.7 & 14.5 & 20.7 & 2.32 \\
\midrule
\multicolumn{9}{l}{\textit{Causal}} \\
\texttt{gpt-4o}          & 45.0 & 83.3 & 40.0 & 0.54 & 43.2 & 77.4 & 29.5 & 0.56 \\
\texttt{o3-mini}         & 97.8 & 96.1 & 86.7 & 1.02 & 84.7 & 96.8 & 68.9 & 0.88 \\
\texttt{deepseek-r1}     & 99.4 & 88.9 & 100.0 & 1.12 & 87.9 & 73.7 & 84.7 & 1.19 \\
\texttt{llama-3.3-70b}   & 33.3 & 83.9 & 22.2 & 0.40 & 29.5 & 79.5 & 18.9 & 0.37 \\
\texttt{qwen3-30b}       & 72.8 & 73.3 & 72.2 & 0.99 & 41.6 & 68.9 & 46.8 & 0.60 \\
\texttt{olmo-3.1-32b}    &  0.0 & 35.0 & 11.7 & 0.00 &  0.0 & 20.0 &  6.3 & 0.00 \\
\bottomrule
\end{tabular}
\end{table}

{
\FloatBarrier
\section{Chain-of-thought trace analysis: protocol and failure catalogue}
\label{app:cot-traces}

\FloatBarrier
\subsection{Protocol}

\paragraph{Sample.} For gpt-4o and deepseek-r1, we took the first 30 failed CoT traces (in evaluation order) of the Hard\,$\cdot$\,Obf cell of each track. Under answer-pool scoring, 3 of the originally sampled deepseek-r1 induction traces are correct; we replaced them with the next three failures.

\paragraph{Labelling.} For each trace, an LLM judge received the trace, the rendered problem, and the gold formal task, and proposed a label (R or E), a sub-category, and the decisive line of the trace; the authors checked every label. The decisive test for deduction and causal discovery is whether the error would survive on the clean gold premises. If the trace leans on content that is not a gold premise (for example narrative wording introduced by the rendering), or misses or misreads a gold premise, the failure is extraction. If it reasons invalidly over correctly read content, it is reasoning. In causal discovery, confusing mediated and direct effects counts as reasoning in either direction, even when the rendering makes it more likely (or the model then abstains with \emph{None}), because the trace reads the data correctly. Induction failures are usually silent: the trace reads drug roles correctly and substitutes a familiar rule, so the trace alone cannot show whether the rendering caused the failure. There we use the matched clean NL task as a control: a failure that disappears on the NL version is extraction, one that persists is reasoning.

Below, each category is listed with its count (gpt-4o / deepseek-r1, out of 30; ``--'' when the category was not used for that model) and a decisive line quoted verbatim from a failed trace. Gold and predicted labels are given as gold$\to$predicted.

\FloatBarrier
\subsection{Deduction: logical fallacies and extraction failures}

\paragraph{Reasoning (18 / 11).}
\begin{itemize}[leftmargin=*, itemsep=2pt]
  \item \textbf{Consistency taken as proof} (8 / --). The model shows that the hypothesis is \emph{consistent} with the premises and concludes that it is \emph{entailed}. Unknown$\to$True: \emph{``no contradiction arises from assuming $\neg$LIN3\_signal(vpc\_cell\_a), the hypothesis is consistent with the premises''}.
  \item \textbf{Non-entailment taken as refutation} (2 / 1). Failing to derive the hypothesis is read as deriving its negation. Unknown$\to$False: \emph{``The hypothesis is not valid because there is no evidence that all FGF8 signaling cells must exhibit BMP4 signaling.''}
  \item \textbf{Ignored disjunct} (2 / --). A disjunctive premise ($A \to B \lor C$) is used as if it named one branch. Unknown$\to$True: \emph{``it is logical to conclude that some neuroepithelial cells will indeed become ventral neuron progenitors.''}
  \item \textbf{Unlicensed existential import} (2 / 2). Existence is inferred from universal premises alone. False$\to$True: \emph{``there must exist at least one primitive streak cell that is not a member of definitive endoderm, supporting the existential quantifier''}.
  \item \textbf{Invented mutual exclusivity} (1 / 2). Two predicates are assumed to exclude each other with no premise saying so, typically from biological background. Unknown$\to$True: \emph{``BMP signaling is exclusively associated with dorsal identities\ldots{} mutually exclusive based on spatial and functional segregation''}.
  \item \textbf{Reversed conditional} (1 / --). From $A \land B \to C$, the model infers $C \to A$. \emph{``since a maturation factor is only activated in a differentiated state, any cell with a maturation factor must already be in a differentiated state''}.
  \item \textbf{Closed-world assumption} (1 / --). A fact not stated is taken to be false. \emph{``There is no explicit mention of cell\_a expressing a competence booster signal''}.
  \item \textbf{Other} (1 / 6), e.g.\ generalising from an example to a universal.
\end{itemize}

\paragraph{Extraction (12 / 19).}
\begin{itemize}[leftmargin=*, itemsep=2pt]
  \item \textbf{Missing premise} (8 / 1). A gold premise needed for the proof is never recovered, so the deciding chain never appears. True$\to$False: \emph{``they could have Dpp expression instead''} (the premise eliminating Dpp was not extracted).
  \item \textbf{Retreat to \emph{Unknown}} (2 / 7). The model loses the chain across documents and abstains. False$\to$Unknown: \emph{``we cannot definitively conclude whether Cell A expresses the blastema progenitor signal or not.''}
  \item \textbf{Distractor wording absorbed} (2 / 6). Narrative wording added by the rendering is treated as a premise. The rendering wrote that one state \emph{presages} another; Unknown$\to$False: \emph{``The term `presages' implies a temporal sequence\ldots{} a cell cannot be in both states simultaneously.''}
  \item \textbf{Merged predicates} (-- / 5). Two distinct predicates with similar descriptions are identified. Unknown$\to$True: \emph{``we assume $\neg$pluripotency\_marker(x) $\to$ $\neg$stemness\_marker(x) and vice versa, as they represent the same concept''}.
\end{itemize}

\FloatBarrier
\subsection{Causal discovery: direct versus mediated effects}

\paragraph{Reasoning (17 / 29).}
\begin{itemize}[leftmargin=*, itemsep=2pt]
  \item \textbf{Mediated effect read as a direct edge} (14 / 21). An intervention effect that travels through the known network is attributed to a direct edge with \textit{XYZ}, including preferring a distant ancestor over the proximal parent. \emph{``Changes in XYZ levels have a significant effect on akt, suggesting that XYZ is upstream of akt''} (akt is reached through the known network).
  \item \textbf{Reversed direction} (2 / --). The trace identifies the correct dependence, then selects the option with the opposite edge: \emph{``JNK appears to influence XYZ, and XYZ seems to influence ERK. This matches option R: XYZ -> erk, XYZ -> jnk''}.
  \item \textbf{Direct edge dismissed, then \emph{None}} (-- / 7). The mirror error: a true direct edge is judged to be mediated, no remaining option fits, and the model selects \emph{None}: \emph{``XYZ$\to$mek is not direct (mek is affected via pkc$\to$mek\ldots)''}.
  \item \textbf{Other} (1 / 1), e.g.\ asserting an edge from a flat, correctly read intervention.
\end{itemize}

\paragraph{Extraction (13 / 0).}
\begin{itemize}[leftmargin=*, itemsep=2pt]
  \item \textbf{Misread data} (10 / --). Large intervention effects are reported as absent, often ending in the \emph{None} option: \emph{``there are no significant changes in the levels of other proteins, suggesting that XYZ is not upstream of any measured proteins''}, whereas intervening on \textit{XYZ} raises mek and pkc by more than 40\%.
  \item \textbf{Missed intervention} (3 / --). An intervention block in another document is never analysed: \emph{``There is no strong evidence from the data that PLC, PKC, PKA, or AKT directly influence XYZ''}, whereas the plc intervention nearly doubles \textit{XYZ}.
\end{itemize}

\FloatBarrier
\subsection{Induction: familiar rules over the data}

The R/E split for induction comes from the NL control (gpt-4o 25 R / 5 E; deepseek-r1 4 R / 26 E). At the level of the trace, the failures fall into three patterns shared by both models.
\begin{itemize}[leftmargin=*, itemsep=2pt]
  \item \textbf{Familiar rule over the data.} The model picks a textbook drug-interaction mechanism and does not test it against the observations: \emph{``One drug being a substrate and the other an inhibitor of the same transporter.''} deepseek-r1 sometimes cites the gold rule and rejects it: \emph{``Enzyme-related pairs (e.g., enzymeinducer and enzymeinhibitor) \ldots{} are less directly tied to interactions in the data''}.
  \item \textbf{Negative example not excluded.} A supplied non-interacting pair that fires the chosen rule is explained away: \emph{``Both inhibit CYP3A4 and CYP2C9, but no significant interaction was observed\ldots{} this is an exception that does not contradict the overall pattern''}.
  \item \textbf{Merged rules.} Two partial rules are combined into a cross-family pair that no rule supports: \emph{``interactions occur when both drugs act as enzyme inhibitors or when both act as target antagonists''}, answered as (enzymeinhibitor, targetantagonist).
\end{itemize}
In the sharpest cases, the model completes the familiar rule by asserting a drug relation that appears neither in the data nor in the documents: \emph{``Aripiprazole (enzymeinhibitor of CYP2D6) and risperidone (targetsubstrate of CYP2D6)''}, although risperidone is only an enzyme inhibitor in the task.

\FloatBarrier
\section{Frontier-model check}
\label{app:frontier}

To test whether the evaluated tiers remain informative for current frontier models, we ran GPT-5.4 (via OpenRouter, high reasoning effort, output cap of 30k tokens, 10k for causal) with direct CoT on 50 tasks of each Hard\,$\cdot$\,Obf cell, parsed and scored exactly as in the main evaluation (Appendix~\ref{app:models}). All responses finished normally.

\begin{table}[H]
\centering
\small
\caption{GPT-5.4 (direct CoT, high reasoning effort) on 50 tasks per setting.}
\label{tab:frontier}
\begin{tabular}{@{}l c@{}}
\toprule
Setting & Correct \\
\midrule
Deduction, Hard\,$\cdot$\,Obf & 49/50 \\
Induction, Hard\,$\cdot$\,Obf & 25/50 \\
Induction, Hard\,$\cdot$\,NL (same tasks) & 50/50 \\
Causal, Hard\,$\cdot$\,Obf & 50/50 \\
Causal, full Sachs graph, open answer (NL) & 30/50 \\
\bottomrule
\end{tabular}
\end{table}

\paragraph{The Hard deduction and causal tiers are nearly saturated; induction is not.} GPT-5.4 solves 49/50 deduction and 50/50 causal tasks, but only 25/50 induction tasks, although it solves all 50 in their clean NL form. The gap is not explained by answer ambiguity: on the 25 tasks with a single valid rule, it scores 25/25 on NL and 8/25 on the rendered form. In 24 of its 25 rendered failures it picks one of two familiar rules (a transporter inhibitor with a transporter substrate, or shared enzyme inhibition), and in all 25 the chosen rule fires on a supplied non-interacting pair that the trace does not check. On the clean form, every answer excludes all negative pairs. As with deepseek-r1 in \S\ref{sec:cot-traces}, the failures come from the rendered evidence, not from the induction step itself.

\paragraph{The generator can produce harder tiers.} To probe the causal track beyond the evaluated tiers, we generated tasks on the complete 11-protein, 19-edge Sachs graph with \textit{XYZ} attached by five directed edges, and asked for the exact edge set as an open answer, without disclosing the number of edges. GPT-5.4 recovers the exact edge set in 30/50 tasks, showing that the same generator can move the causal track out of the saturated regime. Because this setting is in clean NL and also replaces multiple choice with an open answer, part of the drop may come from the answer format.
}

\FloatBarrier
\section{Extended related work}
\label{app:related-extended}

This appendix expands the main-body discussion. {We organise prior work into the three clusters of the main text (controlled reasoning benchmarks, scientific benchmarks, and reasoning robustness to presentation), plus neuro-symbolic and other reasoning methods, and close with} a feature-by-feature comparison (Table~\ref{tab:related}).

\paragraph{{Controlled reasoning benchmarks.}}
These benchmarks expose mechanistic ground truth and parametric difficulty but on templated common-sense surface. ProofWriter~\citep{tafjord2021proofwriter} and ProntoQA~\citep{saparov2023prontoqa} test deductive theorem-proving over templated rule bases; EntailmentBank~\citep{dalvi2021entailmentbank} provides annotated multi-step entailment trees; AR-LSAT~\citep{zhong2022arlsat}, FOLIO~\citep{han2024folio}, and LogiQA-2~\citep{liu2023logiqa2} use prose surface but cover narrow logical genres; ZebraLogic~\citep{lin2025zebralogic} and JustLogic~\citep{chen2025justlogic} stress constraint satisfaction and propositional inference at scale. {bAbI~\citep{weston2015babi} and AbductionRules~\citep{young2022abductionrules} probe elementary and abductive skills on templated text, KOR-Bench~\citep{ma2025korbench} tests the application of newly given rules, SLR~\citep{helff2025slr} synthesises inductive rule-learning tasks with a difficulty curriculum, and FineLogic~\citep{zhou2025finelogic} evaluates the soundness of individual deductive steps.} {CLadder}~\citep{jin2023cladder} and Corr2Cause~\citep{jin2024corr2cause} extend the same template-and-difficulty controls to causal reasoning, SIRBench~\citep{lin2025sirbench} to inductive rule learning, and {SylloBio-NLI}~\citep{wysocka2025syllobio} to biomedical syllogisms. Of these, JustLogic and {SylloBio-NLI} are the closest deductive analogues to our deduction track: JustLogic builds chained propositional-inference puzzles much as we chain syllogisms, but on a generic prose surface and without an obfuscation axis; {SylloBio-NLI} instantiates syllogisms with biomedical entities much as our deduction track fills tree predicates with developmental-biology pathway data, but renders into a single natural-language genre and exposes no parametric obfuscation knob. {Few of these use scientific vocabulary, none poses multi-document reading, and the surface is, by construction, deliberately clean for extraction.}

\paragraph{{Scientific benchmarks.}}
These benchmarks pair real scientific text with human-authored answers. SciFact~\citep{wadden2020scifact} and SciTab~\citep{lu2023scitab} provide claim verification against literature; QASPER~\citep{dasigi2021qasper} and PeerQA~\citep{baumgartner2025peerqa} target paper-grounded answer-span QA; ScienceQA~\citep{lu2022scienceqa} and SciEval~\citep{sun2024scieval} aggregate exam-style multiple-choice questions {(a separate ScienceQA dataset~\citep{saikh2022scienceqa} targets question answering over scholarly articles)}; M3SciQA~\citep{li2024m3sciqa} extends QA to multi-document, multi-modal scientific question answering. {GPQA~\citep{rein2024gpqa} and SciQ~\citep{welbl2017sciq} add expert-written and crowdsourced science multiple-choice questions.} Surface realism is high, but ground truth reduces to a human conclusion, so the reasoning steps remain implicit and difficulty cannot be parametrically controlled.

A separate line of work places benchmarks in scientific settings with auditable ground truth, but on one paradigm at a time or on open-ended workflows. CauSciBench~\citep{acharya2026causcibench} evaluates end-to-end causal inference, and LLM-SRBench~\citep{shojaee2025llmsrbench} targets scientific equation discovery; both have controlled scientific tasks but cover one reasoning style. SciAgent~\citep{ma2024sciagent} broadens evaluation to tool-augmented scientific problem solving. DiscoveryBench~\citep{majumder2024discoverybench} and the AI Scientist~\citep{lu2024aiscientist} formalise open-ended data-driven discovery, valuable for measuring discovery workflows but at the opposite end of the controlled-vs-open spectrum from a benchmark of reasoning-style coverage.

{\paragraph{Reasoning robustness to presentation.} A separate line of work keeps the reasoning problem fixed and varies how it is presented. MuSR~\citep{sprague2024musr} renders reasoning trees into long narratives, and SATBench~\citep{wei2025satbench} turns SAT formulas into story puzzles checked for consistency, much as our round-trip check validates each rendering. CLUTRR~\citep{sinha2019clutrr} and FLD~\citep{morishita2023fld} generate reasoning problems with distractor facts, BABILong~\citep{kuratov2024babilong} hides bAbI facts in long filler text, and Same Task, More Tokens~\citep{levy2024sametask} and NoLiMa~\citep{modarressi2025nolima} show that longer input and lexical mismatch degrade reasoning and retrieval well below context limits. GSM-IC~\citep{shi2023gsmic}, GSM-Symbolic~\citep{mirzadeh2025gsmsymbolic}, and GSM-$\infty$~\citep{zhou2025gsminfty} show that irrelevant context and surface variation break mathematical reasoning, GSM-$\infty$ with separate knobs for complexity and noise, and even the order of premises changes deductive accuracy~\citep{chen2024premise}. In biomedicine, MedCalc-Bench~\citep{khandekar2024medcalc} requires extracting values from patient notes before an exact computation, and RABBITS~\citep{gallifant2024rabbits} shows that swapping brand and generic drug names alone lowers accuracy. These works use synthetic or single-genre perturbations, mostly on math or toy logic; SciR renders the premises into several realistic scientific genres across three reasoning paradigms and shows that such rendering also degrades the formalisation step of neuro-symbolic pipelines.}

\paragraph{{Neuro-symbolic and other reasoning methods.}}
A growing line of method and benchmark papers studies a single reasoning paradigm in isolation. Deductive: LogicLM~\citep{pan2023logiclm}, LINC~\citep{olausson2023linc}, FLARE~\citep{arakelyan2025flare}, QuaSAR~\citep{ranaldi2025quasar}, adaptive autoformalisation~\citep{xu2026adaptive}, and SymbCoT~\citep{xu2024symbcot}. Inductive: hypothesis search~\citep{wang2024hypothesis} and {hypothesis refinement}~\citep{qiu2024phenomenal}. Causal: {LLM causal reasoning}~\citep{kiciman2024causal} and discovery verification~\citep{he2026doverifier}. Each paper covers one piece of the picture; none places the three paradigms on a shared scientifically-styled surface with parametric control on both inference complexity and premise obfuscation.

\paragraph{Feature-by-feature comparison.}
Table~\ref{tab:related} positions SciR against a representative subset of the above. The distinctive combination is multi-document scientific-genre surface with parametric control on both axes; individual columns are matched by prior work, but the combination is not.

\begin{table}[H]
\centering
\scriptsize
\setlength{\tabcolsep}{3pt}
\renewcommand{\arraystretch}{1.05}
\caption{SciR vs.\ representative prior benchmarks. \checkmark{} = supported, $\sim$ = partial, -- = not supported. Columns marked $\circlearrowright$ are tunable axes (the benchmark exposes a parametric knob along that dimension). Surface labels: \emph{templ.} = templated; \emph{prose} = generic natural-language prose; \emph{real-sci.} = real scientific text; \emph{sci.-shaped} = scientifically-styled but synthetic.}
\label{tab:related}
\resizebox{\textwidth}{!}{%
\begin{tabular}{@{}l l l c c c c c@{}}
\toprule
Benchmark & Reasoning family & Surface
  & \rotatebox{60}{Verifiable GT}
  & \rotatebox{60}{Inf.\ complexity\,$\circlearrowright$}
  & \rotatebox{60}{Obfuscation\,$\circlearrowright$}
  & \rotatebox{60}{Multi-doc}
  & \rotatebox{60}{Domain-tuned obf.} \\
\midrule
ProofWriter~\citep{tafjord2021proofwriter}        & deduction (theorem-pr.)         & templ.       & \checkmark & \checkmark & --         & --         & --         \\
ProntoQA~\citep{saparov2023prontoqa}              & deduction (theorem-pr.)         & templ.       & \checkmark & \checkmark & --         & --         & --         \\
FOLIO~\citep{han2024folio}                        & deduction (FOL, hand-written)   & prose        & \checkmark & --         & --         & --         & --         \\
JustLogic~\citep{chen2025justlogic}               & deduction (propositional)       & templ.       & \checkmark & \checkmark & --         & --         & --         \\
ZebraLogic~\citep{lin2025zebralogic}              & deduction (constraint-sat)      & templ.       & \checkmark & \checkmark & --         & --         & --         \\
LogiQA-2~\citep{liu2023logiqa2}                   & deduction (LSAT-style)          & prose        & $\sim$     & --         & --         & --         & --         \\
{CLadder}~\citep{jin2023cladder}                    & causal (Pearl-rung)             & prose        & \checkmark & $\sim$     & --         & --         & --         \\
Corr2Cause~\citep{jin2024corr2cause}              & causal (DAG discovery)          & prose        & \checkmark & $\sim$     & --         & --         & --         \\
SIRBench~\citep{lin2025sirbench}                  & induction (rule from ex.)       & sci.-shaped  & \checkmark & $\sim$     & --         & --         & --         \\
{SylloBio-NLI}~\citep{wysocka2025syllobio}              & deduction (biomedical)          & sci.-shaped  & \checkmark & $\sim$     & --         & --         & --         \\
SciFact~\citep{wadden2020scifact}                 & sci.\ QA (claim verif.)         & real-sci.    & --         & --         & --         & $\sim$     & --         \\
QASPER~\citep{dasigi2021qasper}                   & sci.\ QA (answer-span)          & real-sci.    & --         & --         & --         & --         & --         \\
DiscoveryBench~\citep{majumder2024discoverybench} & open-ended discovery            & real-sci.    & $\sim$     & --         & --         & $\sim$     & --         \\
HotpotQA~\citep{yang2018hotpotqa}                 & multi-hop factoid QA            & prose        & $\sim$     & --         & --         & \checkmark & --         \\
{GPQA~\citep{rein2024gpqa}} & {sci.\ QA (expert MC)} & {real-sci.} & {$\sim$} & {--} & {--} & {--} & {--} \\
{SciQ~\citep{welbl2017sciq}} & {sci.\ QA (crowdsourced MC)} & {real-sci.} & {$\sim$} & {--} & {--} & {--} & {--} \\
{KOR-Bench~\citep{ma2025korbench}} & {rule application (given rules)} & {prose} & {\checkmark} & {--} & {--} & {--} & {--} \\
{bAbI~\citep{weston2015babi}} & {toy QA skills} & {templ.} & {\checkmark} & {--} & {--} & {--} & {--} \\
{AbductionRules~\citep{young2022abductionrules}} & {abduction} & {templ.} & {\checkmark} & {$\sim$} & {--} & {--} & {--} \\
{SLR~\citep{helff2025slr}} & {induction (rule synthesis)} & {templ.} & {\checkmark} & {\checkmark} & {--} & {--} & {--} \\
{FineLogic~\citep{zhou2025finelogic}} & {deduction (stepwise eval.)} & {prose} & {\checkmark} & {--} & {--} & {--} & {--} \\
{MuSR~\citep{sprague2024musr}} & {multistep soft reasoning} & {prose (narratives)} & {\checkmark} & {$\sim$} & {--} & {--} & {--} \\
{BABILong~\citep{kuratov2024babilong}} & {long-context QA (bAbI)} & {prose (filler text)} & {\checkmark} & {$\sim$} & {\checkmark} & {--} & {--} \\
{GSM-IC~\citep{shi2023gsmic}} & {math word problems} & {prose} & {\checkmark} & {--} & {$\sim$} & {--} & {--} \\
{GSM-$\infty$~\citep{zhou2025gsminfty}} & {math word problems} & {templ.} & {\checkmark} & {\checkmark} & {\checkmark} & {--} & {--} \\
\midrule
\textbf{SciR (ours)} & deduction\,$+$\,induction\,$+$\,causal-discovery & sci.-shaped (multi-genre) & \checkmark & \checkmark & \checkmark & \checkmark & \checkmark \\
\bottomrule
\end{tabular}%
}
\end{table}

\end{document}